%% file: main_no_dataset.tex
\documentclass[11pt, a4paper, logo, copyright]{googlecloud}

\pdftrailerid{redacted}

\makeatletter
\renewcommand\bibentry[1]{\nocitep{#1}{\frenchspacing\@nameuse{BR@r@#1\@extra@b@citeb}}}
\makeatother

\usepackage{kantlipsum, lipsum}
\usepackage{dsfont}

\input{commands}

\usepackage[authoryear, sort&compress, round]{natbib}

\usepackage[utf8]{inputenc} %
\usepackage{lmodern}
\usepackage[T1]{fontenc}    %
\usepackage{hyperref}       %
\hypersetup{
  colorlinks,
  citecolor=blue,
  linkcolor=red,
  urlcolor=blue}
\usepackage{url}            %
\usepackage{booktabs}       %
\usepackage{amsfonts}       %
\usepackage{nicefrac}       %
\usepackage{microtype}      %
\usepackage{wrapfig} %
\usepackage{graphicx} %
\usepackage{soul} %
\usepackage{enumitem}
\usepackage{amssymb,amsmath}
\usepackage{tikz, adjustbox}
\usepackage[most]{tcolorbox}
\usepackage{subcaption}
\usepackage{multirow}
\usepackage{arydshln}
\usepackage{float}
\usepackage[capitalize,noabbrev,nameinlink]{cleveref}
\usepackage{fancyvrb}

\usepackage{tcolorbox}
\input{math_commands}

 \usepackage{amsthm}
\usepackage{algorithm}
\usepackage{algpseudocode}
\usepackage{amsmath}
\usepackage{xcolor} 
\definecolor{darkgreen}{rgb}{0.0, 0.5, 0.0}
\usepackage{enumitem}
\setlist[itemize]{align=parleft,left=0pt..1em}
\usepackage{multirow}
\usepackage{media9} 
\algrenewcommand{\algorithmiccomment}[1]{\textcolor{gray}{\hfill$\triangleright$ #1}}
\usepackage{booktabs}
\usepackage{graphicx}
\usepackage{booktabs}
\usepackage{subcaption}  
\usepackage{wrapfig}
\usepackage{arydshln}
\usepackage{tabularx}
\usepackage{caption}
\usepackage[most]{tcolorbox}
\usepackage{courier} 
\tcbuselibrary{skins}

\theoremstyle{definition}

\usepackage{caption}
\usepackage{soul}

\input{math_commands.tex}
\usepackage{hyperref}

\tcbuselibrary{listingsutf8}
\usepackage{listings}
\usepackage{xcolor}
\UseRawInputEncoding
\definecolor{lavender}{RGB}{242, 242, 255}  
\everymath{\displaystyle} 

\usepackage{graphicx}       
\usepackage{subcaption}     
\usepackage{caption}        
\usepackage{ragged2e}       
\usepackage{tikz}           
\usepackage{amsmath}        
\crefname{algorithm}{Alg.}{Algs.}
\Crefname{algorithm}{Alg.}{Algs.}

\usepackage{ragged2e} 
\definecolor{lightgray}{gray}{0.95}
\definecolor{myblue}{rgb}{0.1,0.1,0.6}

\newcommand{\model}{\textsc{VISTA}}

\usepackage{pifont}

\usepackage{bbold}

\usepackage{listings}

\lstdefinestyle{mystyle}{
    backgroundcolor=\color{backcolour},   
    commentstyle=\color{codegreen},
    keywordstyle=\color{magenta},
    numberstyle=\tiny\color{codegray},
    stringstyle=\color{codepurple},
    basicstyle=\ttfamily\scriptsize,
    breakatwhitespace=false,         
    breaklines=true,                 
    captionpos=b,                    
    keepspaces=true,                 
    numbers=left,                    
    numbersep=5pt,                  
    showspaces=false,                
    showstringspaces=false,
    showtabs=false,                  
    tabsize=2,
    frame=none,
    aboveskip=1pt,
    belowskip=1pt,
}
\lstdefinestyle{plainins}{
    backgroundcolor=\color{white},   
    commentstyle=\color{codegreen},
    keywordstyle=\color{magenta},
    numberstyle=\tiny\color{codegray},
    stringstyle=\color{codepurple},
    basicstyle=\ttfamily\scriptsize,
    breakatwhitespace=false,         
    breaklines=true,                 
    captionpos=b,                    
    keepspaces=true,                 
    numbers=none,                    
    numbersep=5pt,                  
    showspaces=false,                
    showstringspaces=false,
    showtabs=false,                  
    tabsize=2,
    aboveskip=0pt,
    belowskip=0pt,
    frame=single
}
\lstdefinestyle{plainexam}{
    backgroundcolor=\color[HTML]{FFFCF3},   
    commentstyle=\color{codegreen},
    keywordstyle=\color{magenta},
    numberstyle=\tiny\color{codegray},
    stringstyle=\color{codepurple},
    basicstyle=\ttfamily\scriptsize,
    breakatwhitespace=false,         
    breaklines=true,                 
    captionpos=b,                    
    keepspaces=true,                 
    numbers=none,                    
    numbersep=5pt,                  
    showspaces=false,                
    showstringspaces=false,
    showtabs=false,                  
    tabsize=2,
    aboveskip=0pt,
    belowskip=0pt
}

\title{\model{}: A Test-Time Self-Improving Video Generation Agent}

\correspondingauthor{xuanlong.do@u.nus.edu, xingchenw@google.com, soarik@google.com}

\renewcommand{\today}{}

\author[1 2 *]{Do Xuan Long}
\author[1]{Xingchen Wan}
\author[1]{Hootan Nakhost}
\author[1]{Chen-Yu Lee}
\author[1]{Tomas Pfister}
\author[1]{Sercan \"O. Ar\i k}

\affil[1]{Google}
\affil[2]{National University of Singapore}

\begin{abstract}
Despite rapid advances in text-to-video synthesis, generated video quality remains critically dependent on precise user prompts. Existing test-time optimization methods, successful in other domains, struggle with the multi-faceted nature of video. In this work, we introduce \model{}, a novel multi-agent system that autonomously improves video generation through refining prompts in an iterative loop. \model{} first decomposes a user's idea into a structured temporal plan. After generation, the best video is identified through a robust pairwise tournament. This winning video is then critiqued by a trio of specialized agents focusing on visual, audio, and contextual fidelity. Finally, a reasoning agent synthesizes this feedback to introspectively rewrite and enhance the prompt for the next generation cycle. Experiments on single- and multi-scene video generation scenarios show that while prior methods yield inconsistent gains, \model{} consistently improves video quality and alignment with user intent, achieving up to 60\% pairwise win rate against state-of-the-art baselines. Human evaluators concur, preferring \model{}'s outputs in 66.4\% of comparisons.
\end{abstract}

\begin{document}

\maketitle

\begin{table}[h]
\centering
\begin{tabularx}{\textwidth}{@{}X@{}}

\begin{tabular}{@{}c@{}c@{}c@{\hspace{6mm}}c@{}c@{}c@{}}
\multicolumn{3}{c}{\textbf{Direct Prompting (DP)}} & \multicolumn{3}{c}{\textbf{\model{} (Ours)}} \\
\includegraphics[width=0.16\linewidth]{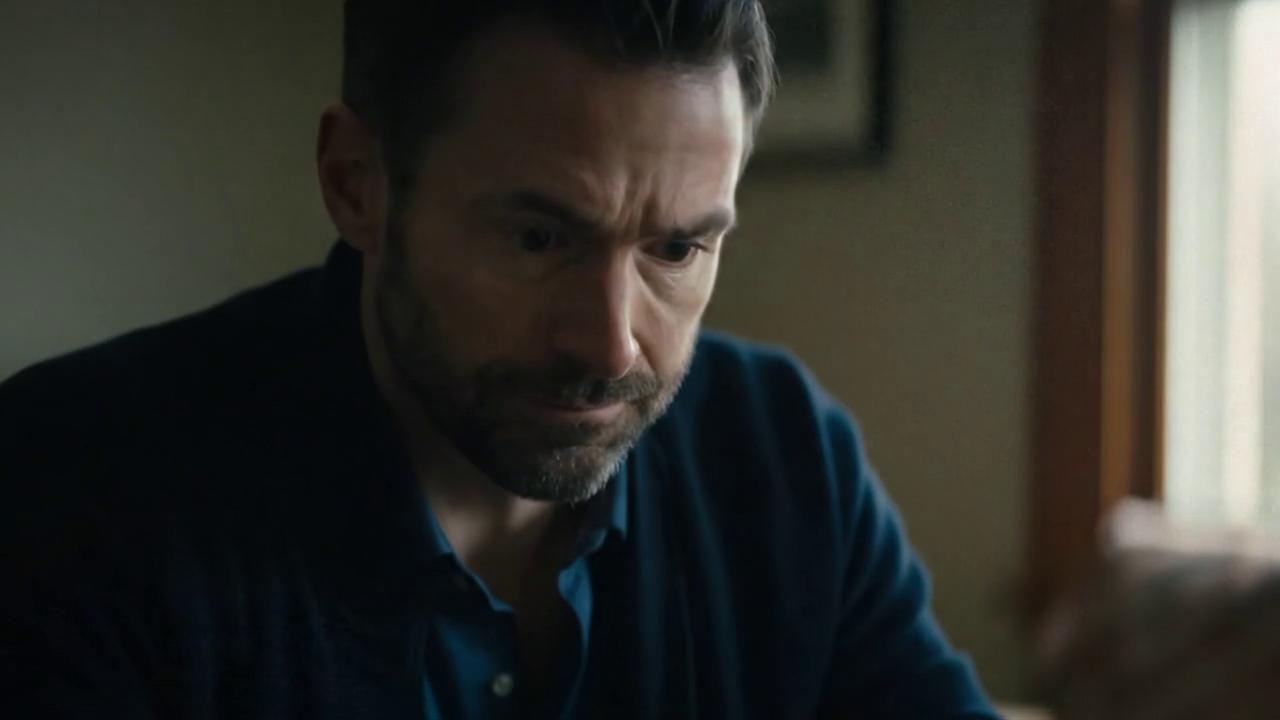} &
\includegraphics[width=0.16\linewidth]{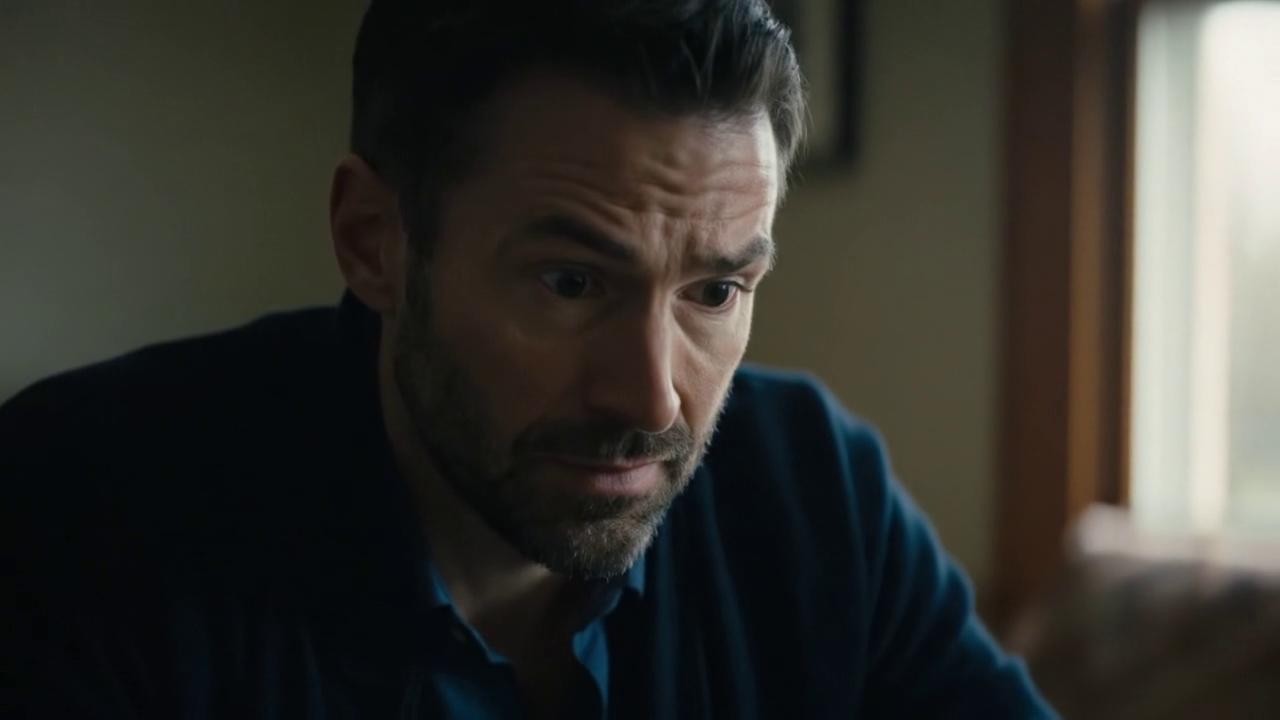} &
\includegraphics[width=0.16\linewidth]{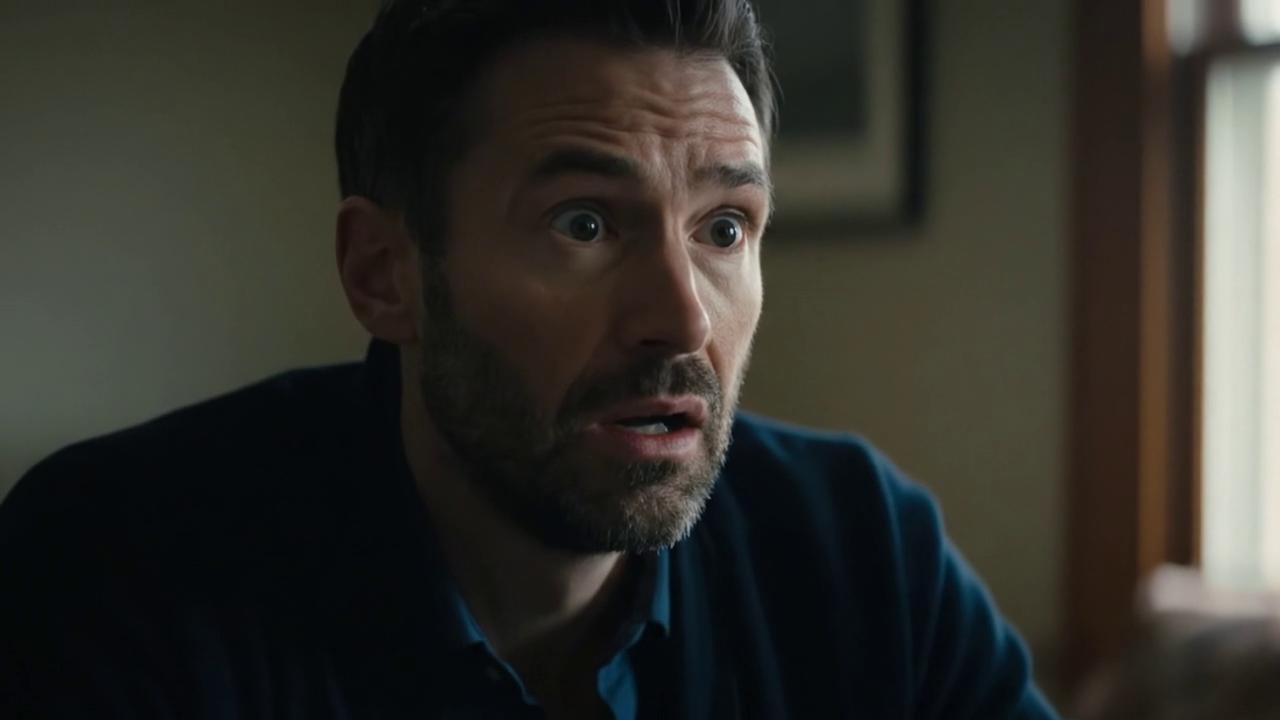} &

\includegraphics[width=0.16\linewidth]{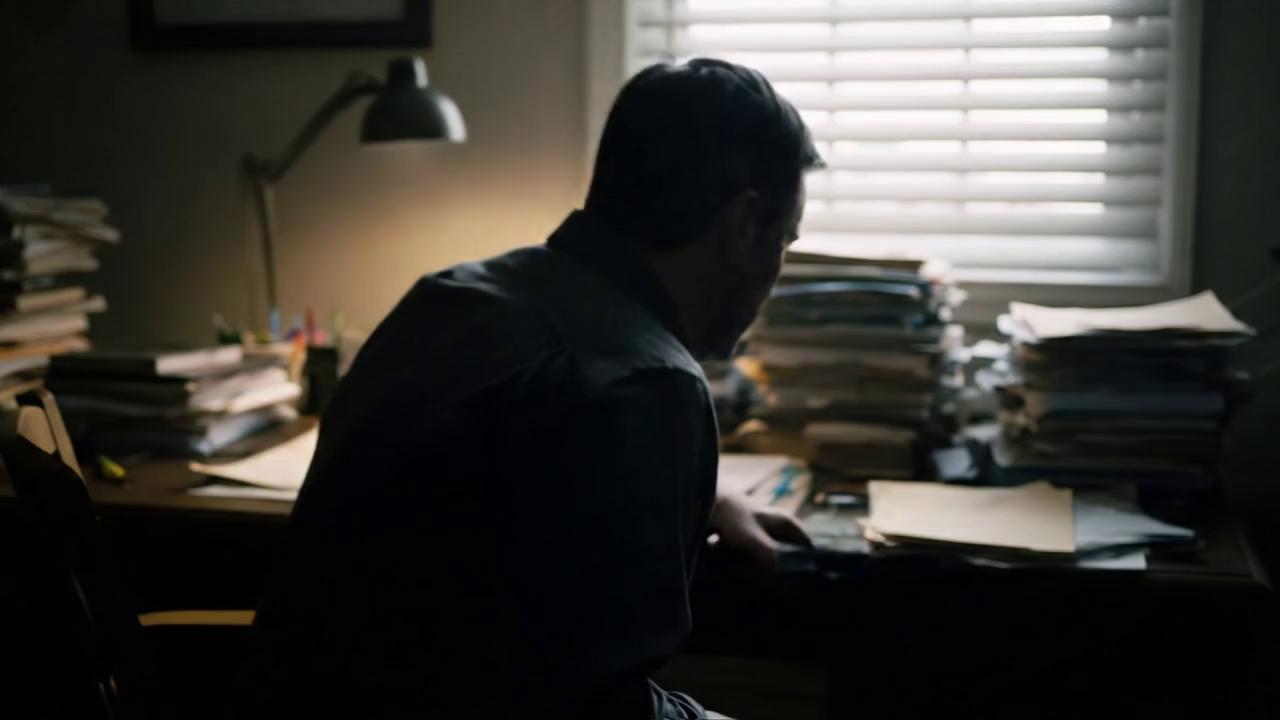} &
\includegraphics[width=0.16\linewidth]{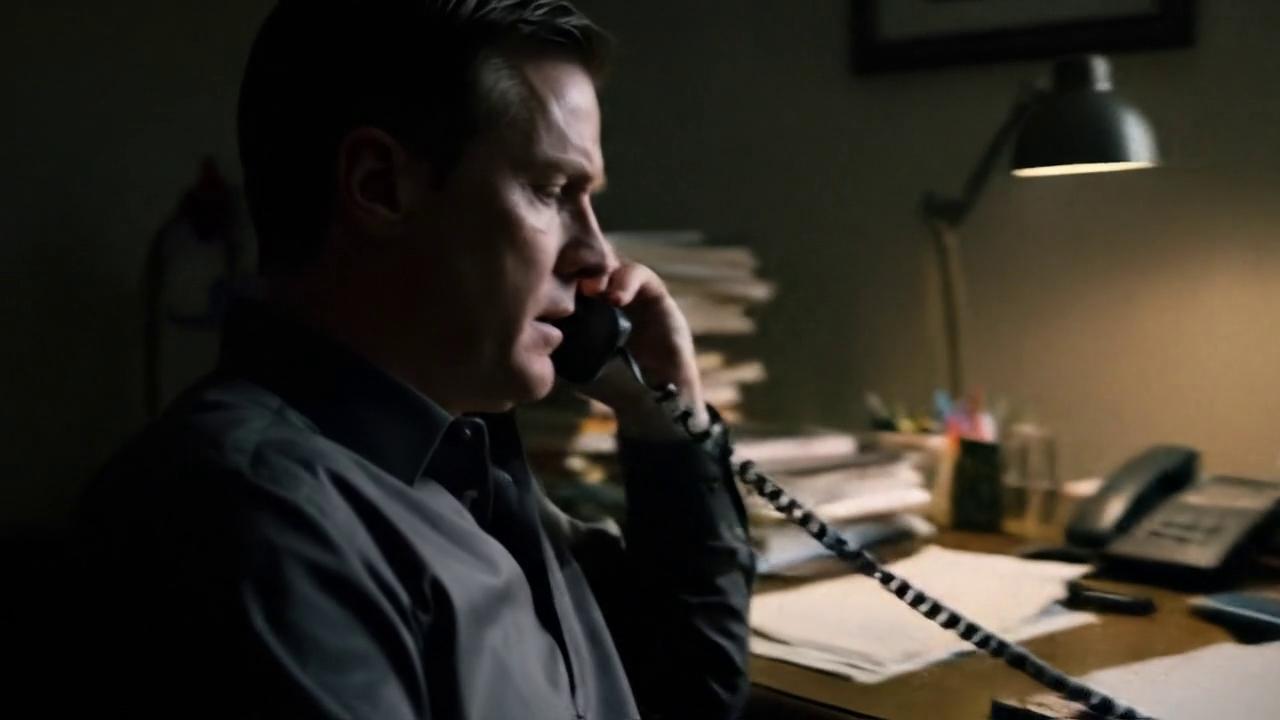} & \includegraphics[width=0.16\linewidth]{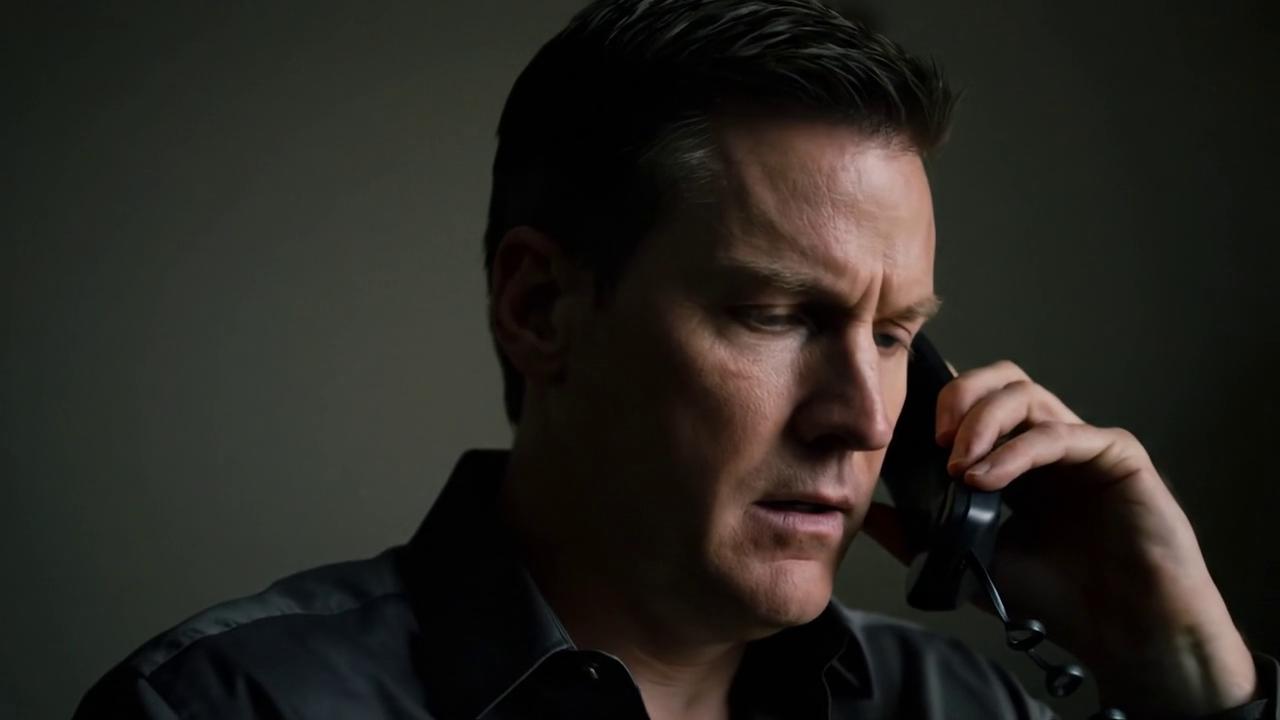} \\
\end{tabular} \\
\vspace{-5mm}
\footnotesize\emph{Single-scene \citep{polyak2025moviegencastmedia}: The person's forehead creased with worry as he listened to bad news.}
\\[2mm]

\begin{tabular}{@{}c@{}c@{}c@{\hspace{6mm}}c@{}c@{}c@{}}
\includegraphics[width=0.16\linewidth]{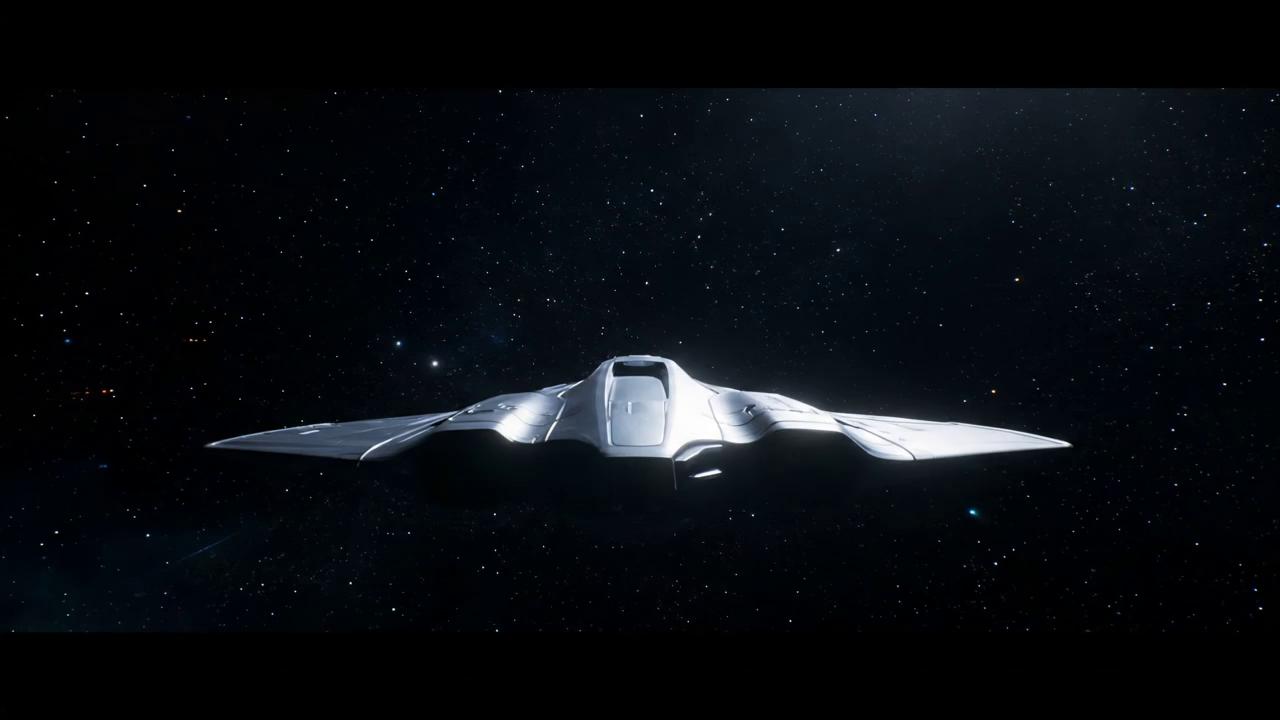} &
\includegraphics[width=0.16\linewidth]{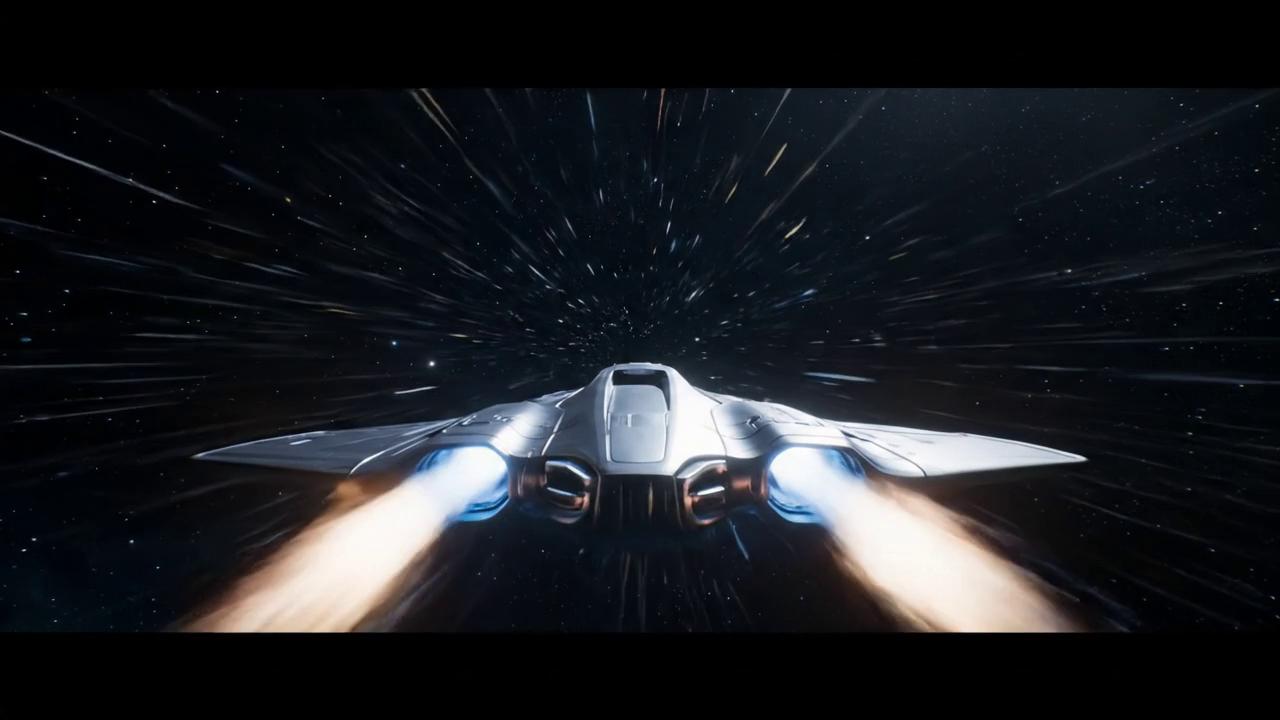} &
\includegraphics[width=0.16\linewidth]{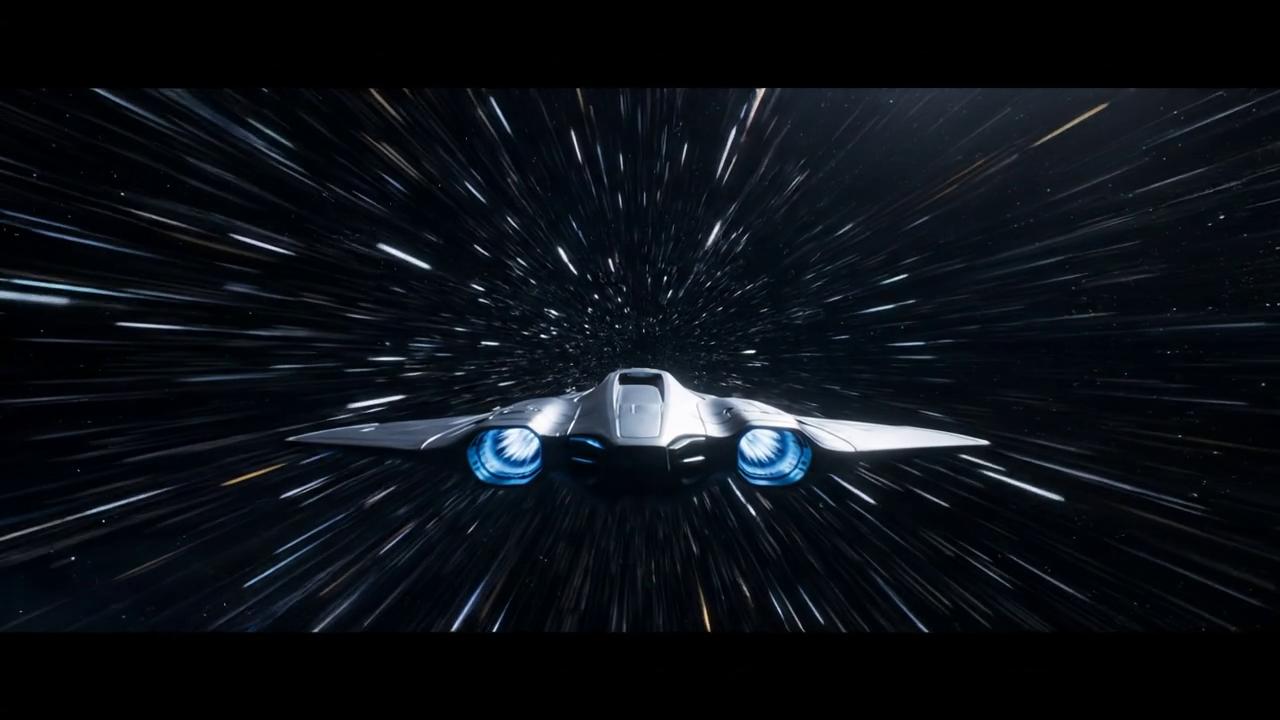} &

\includegraphics[width=0.16\linewidth]{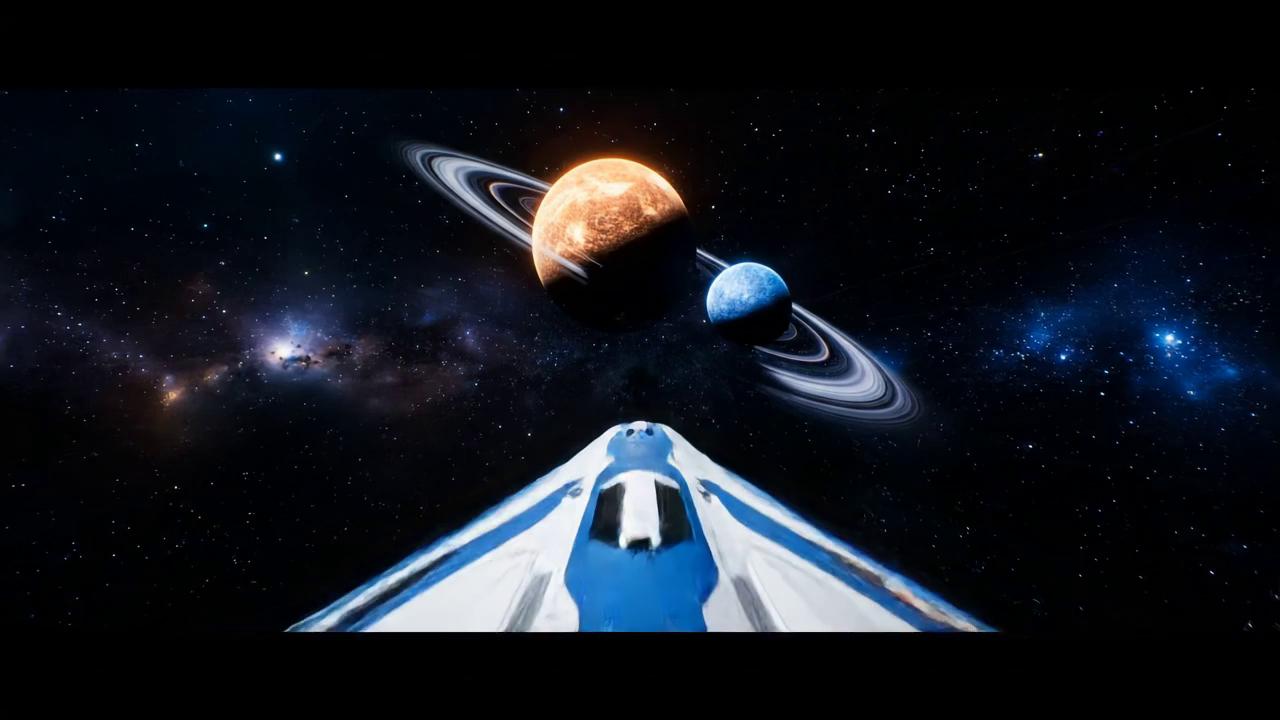} &
\includegraphics[width=0.16\linewidth]{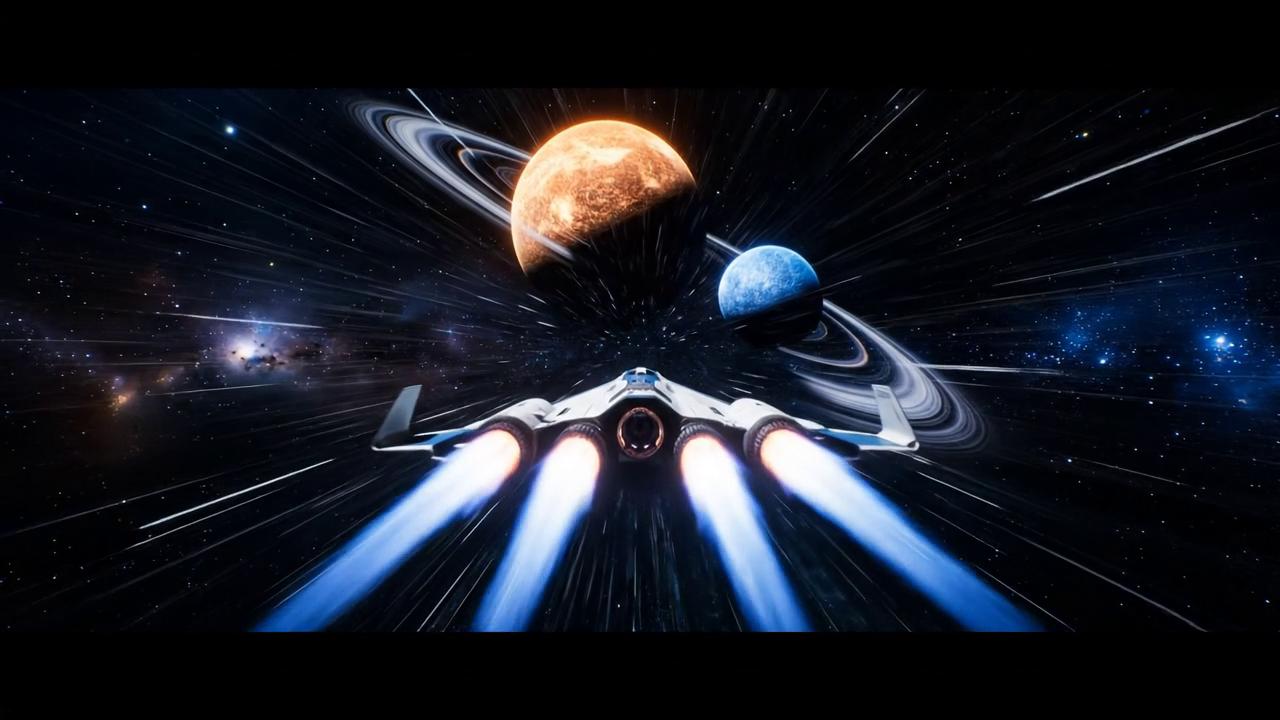} &
\includegraphics[width=0.16\linewidth]{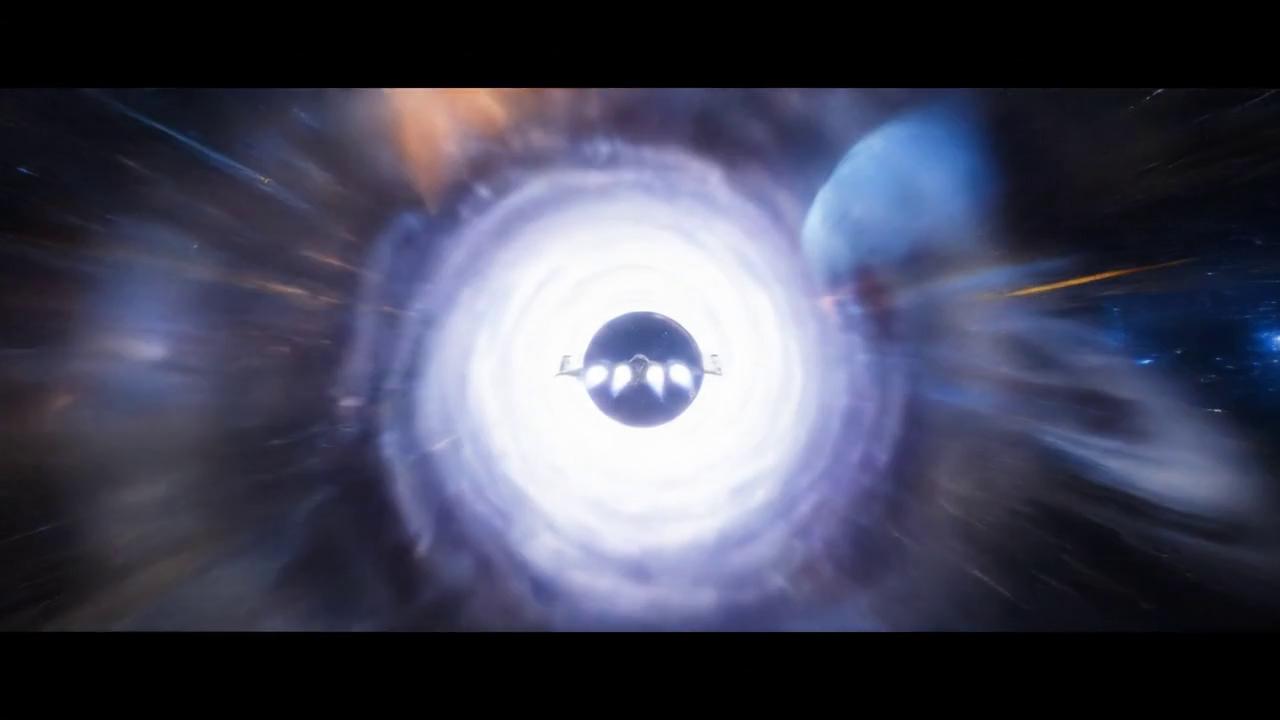} \\
\end{tabular} \\
\vspace{-5mm}
\footnotesize\emph{Single-scene \citep{polyak2025moviegencastmedia}: A spaceship entering hyperdrive, stars streaking past as it accelerates.}
\\[2mm]

\begin{tabular}{@{}c@{}c@{}c@{\hspace{6mm}}c@{}c@{}c@{}}
\includegraphics[width=0.16\linewidth]{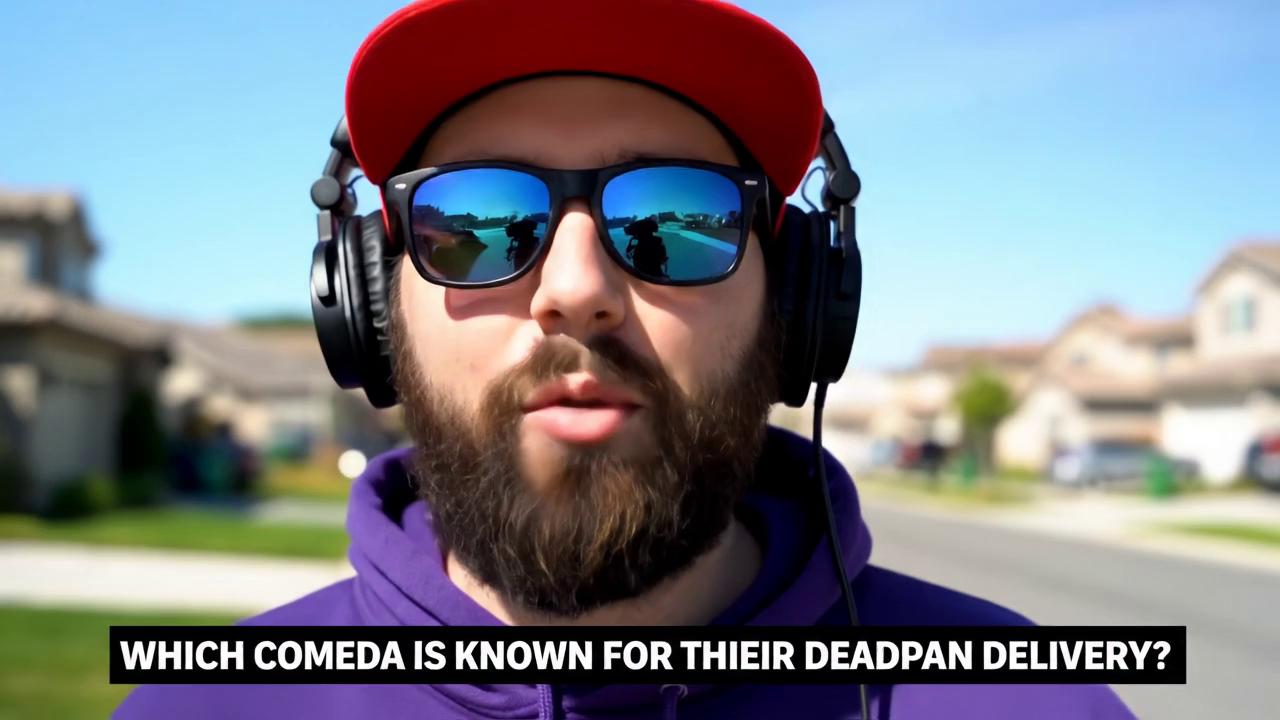} &
\includegraphics[width=0.16\linewidth]{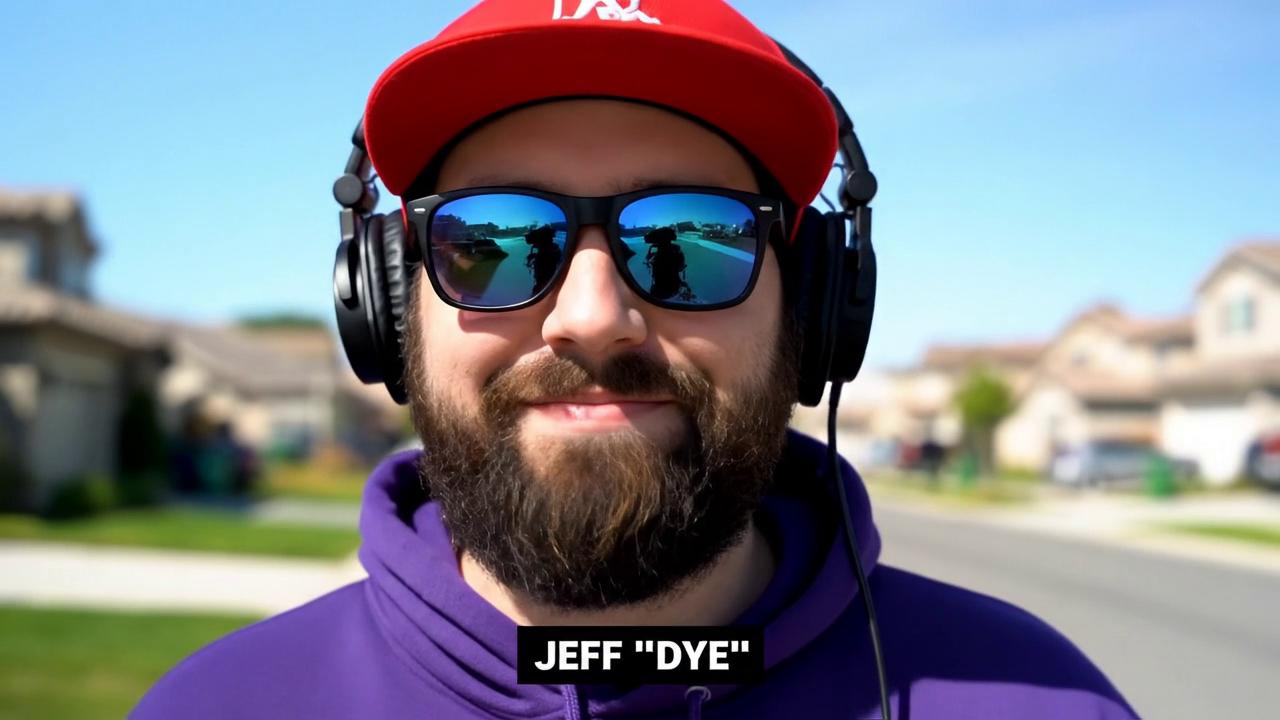} &
\includegraphics[width=0.16\linewidth]{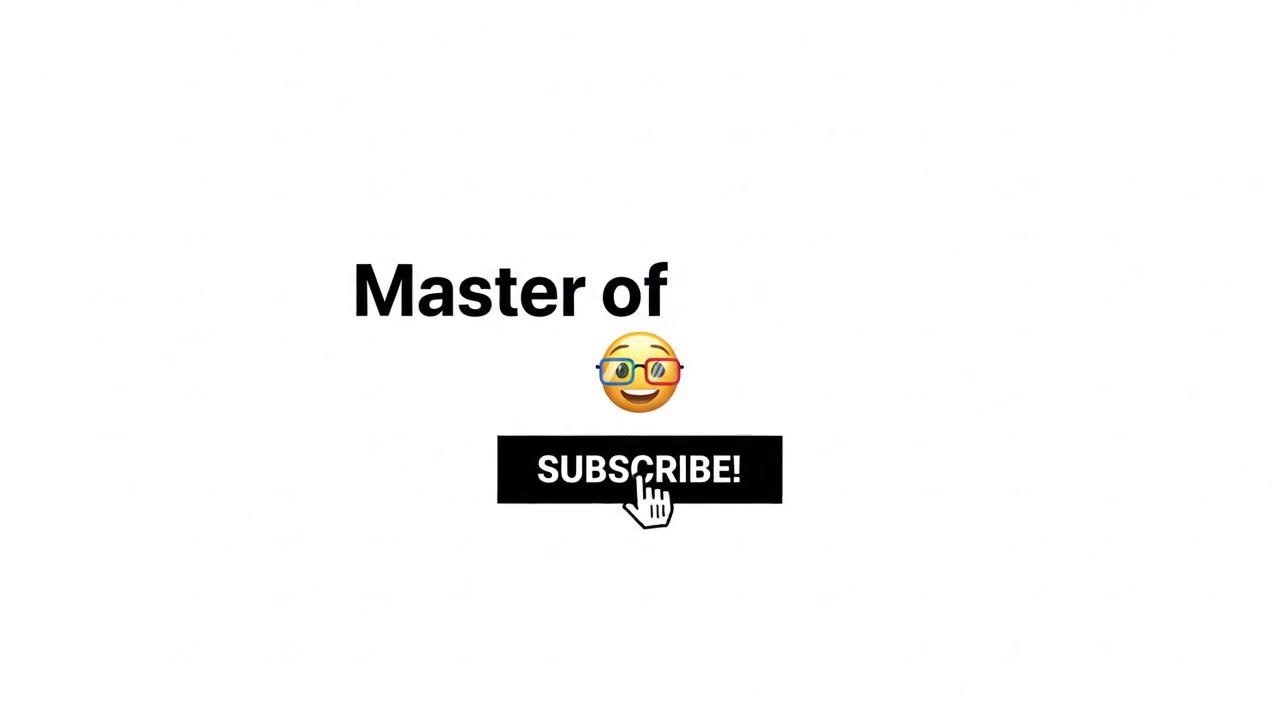} &

\includegraphics[width=0.16\linewidth]{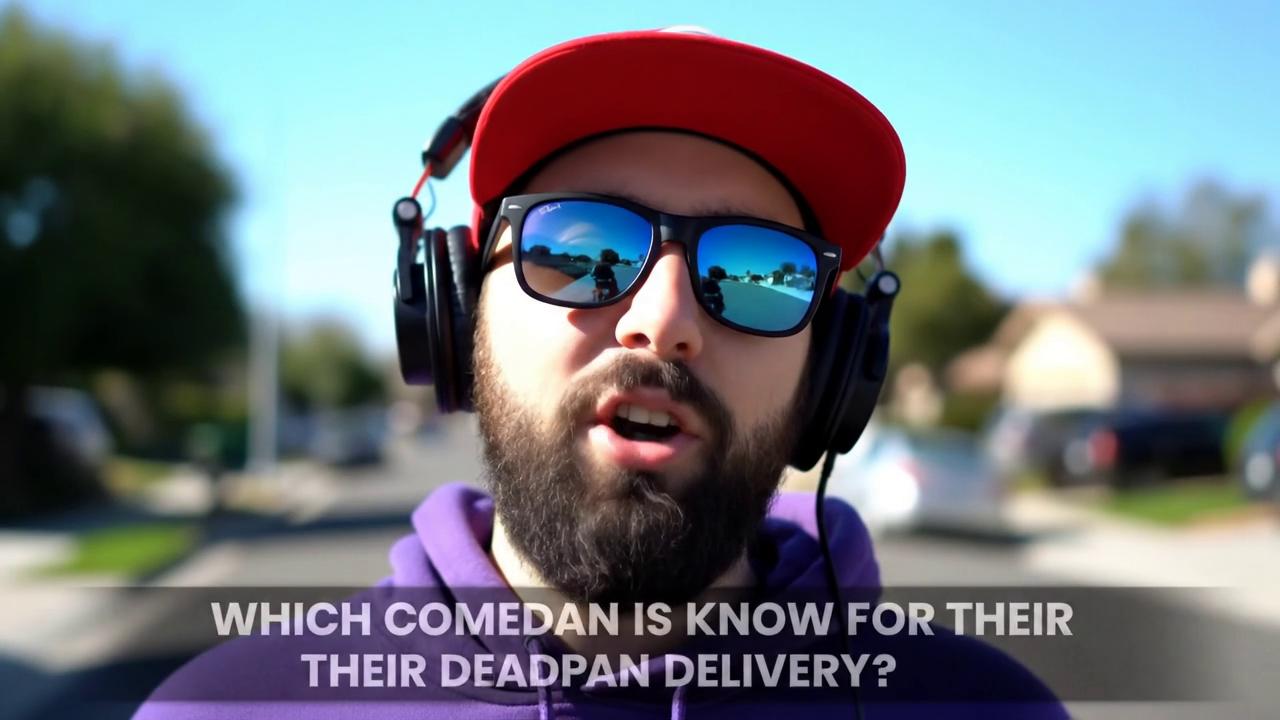} &
\includegraphics[width=0.16\linewidth]{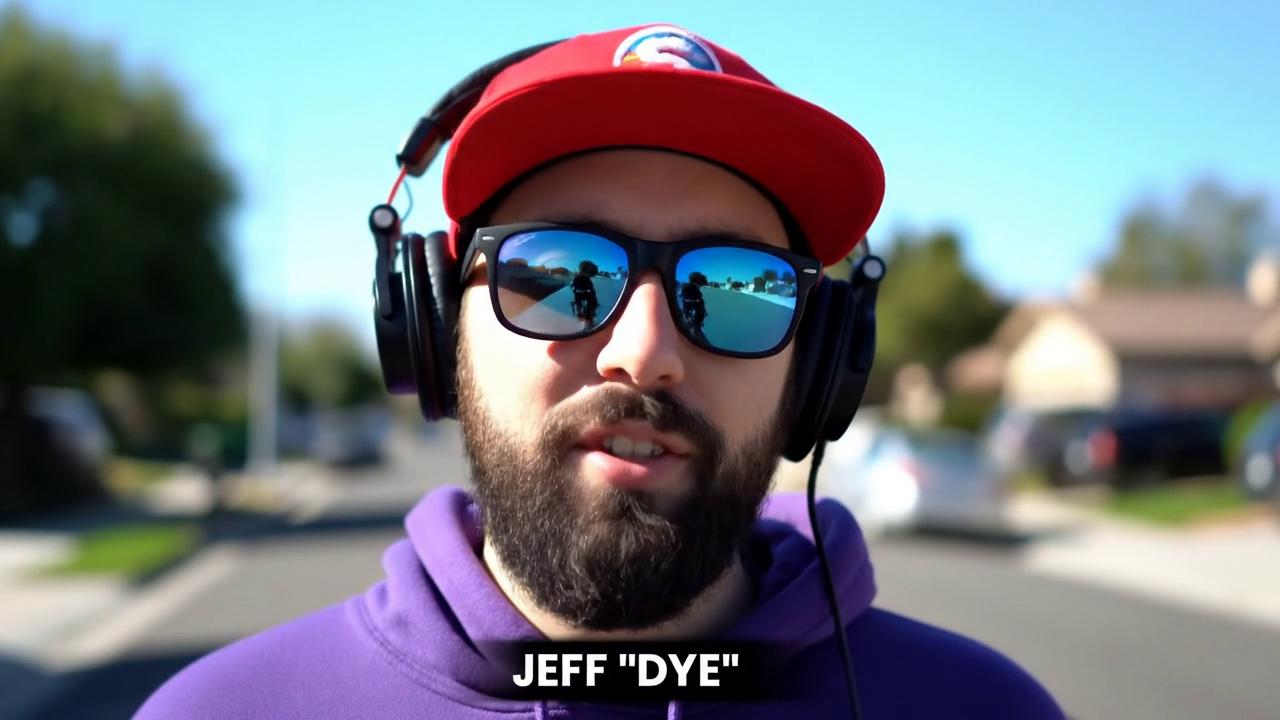} &
\includegraphics[width=0.16\linewidth]{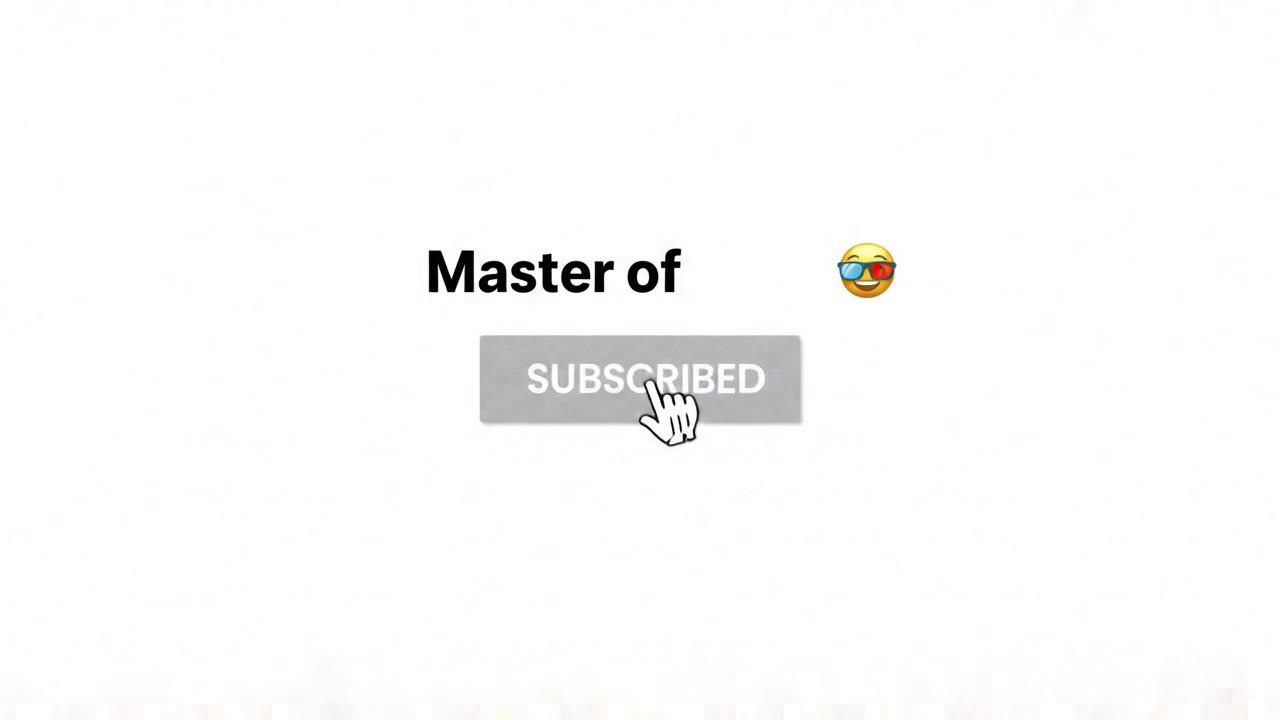} \\
\end{tabular} \\
\vspace{-5mm}
\footnotesize\emph{Multi-scene-Interviews: The video features a man outdoors, asking a trivia question about a comedian known\dots}
\\[2mm]

\begin{tabular}{@{}c@{}c@{}c@{\hspace{6mm}}c@{}c@{}c@{}}    
\includegraphics[width=0.16\linewidth]{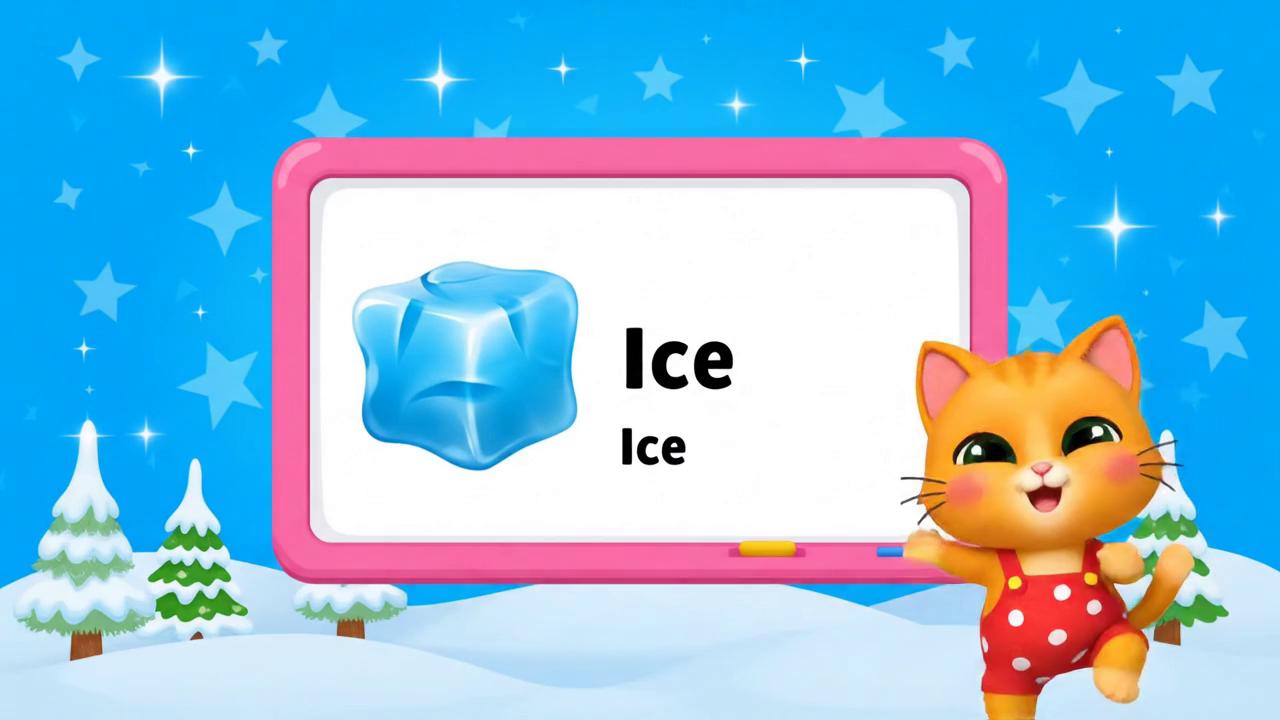} &
\includegraphics[width=0.16\linewidth]{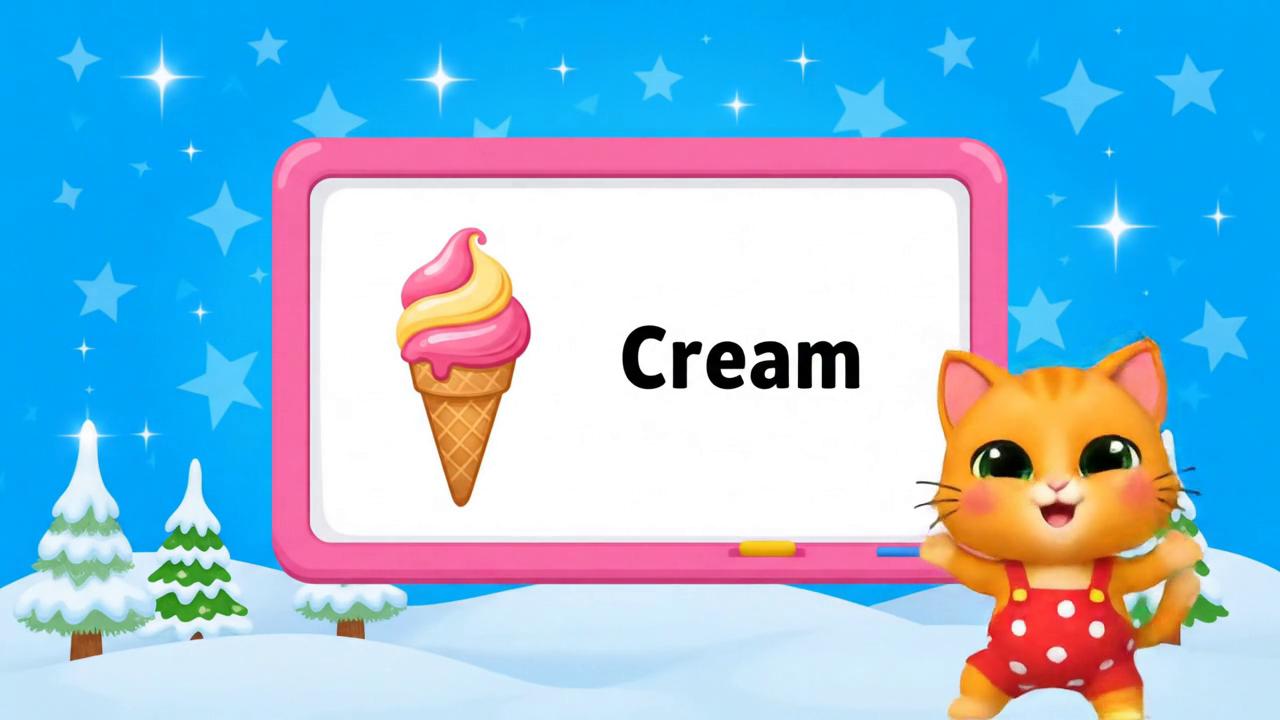} &
\includegraphics[width=0.16\linewidth]{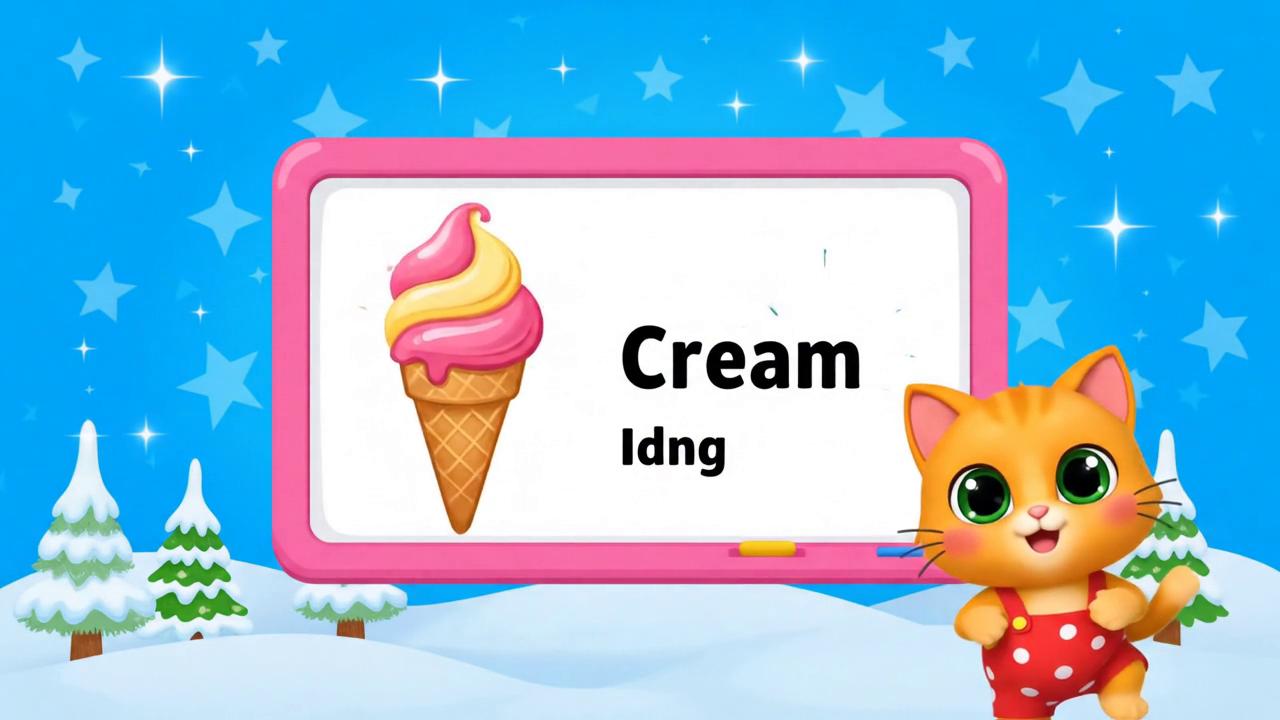} & 
\includegraphics[width=0.16\linewidth]{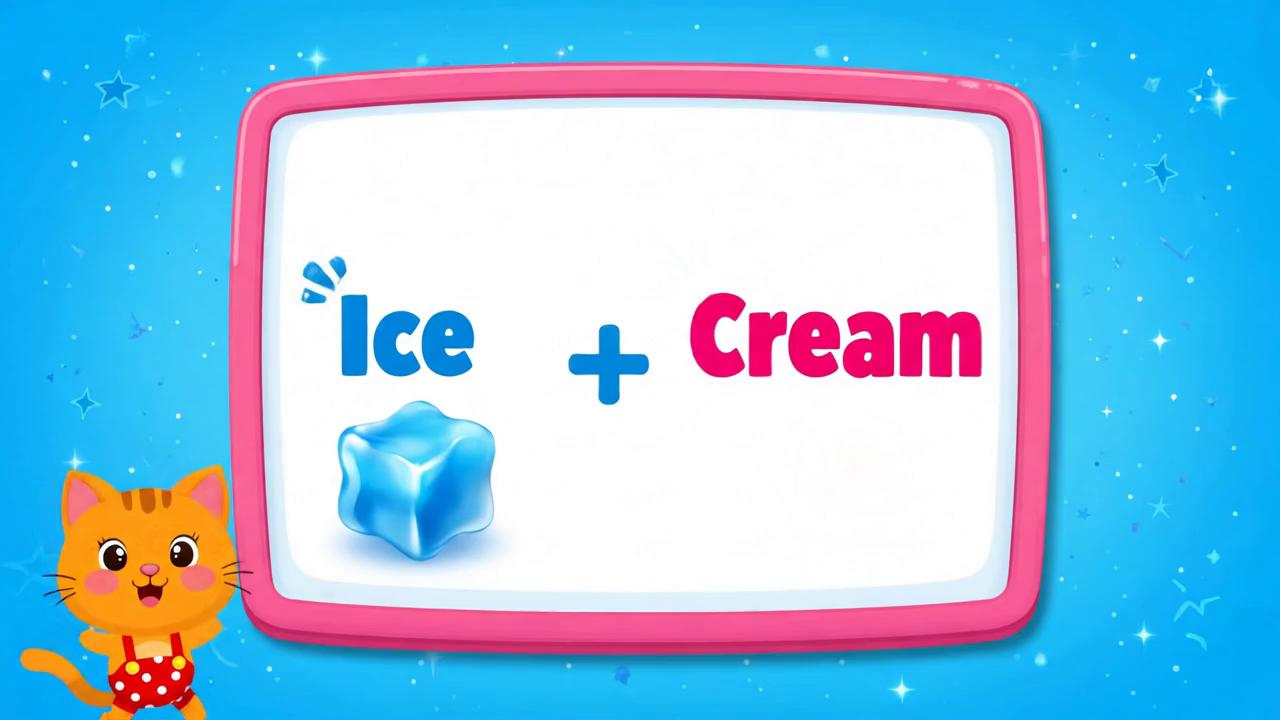} &
\includegraphics[width=0.16\linewidth]{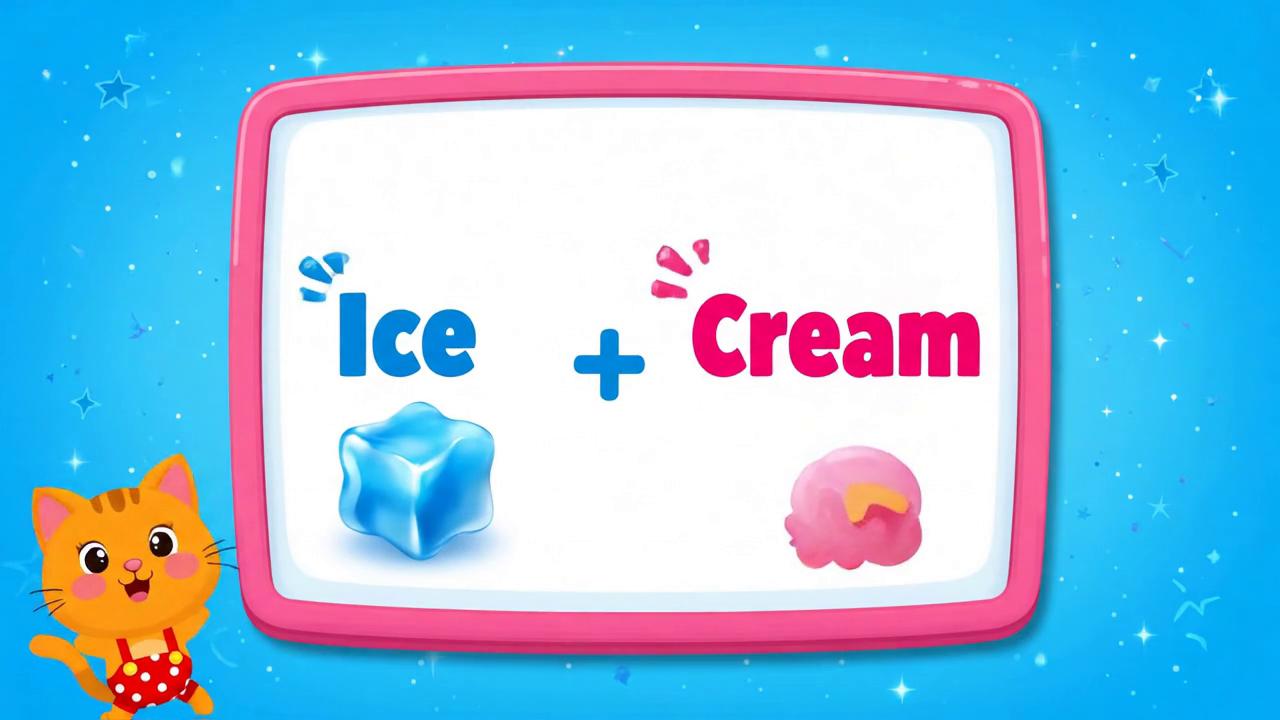} &
\includegraphics[width=0.16\linewidth]{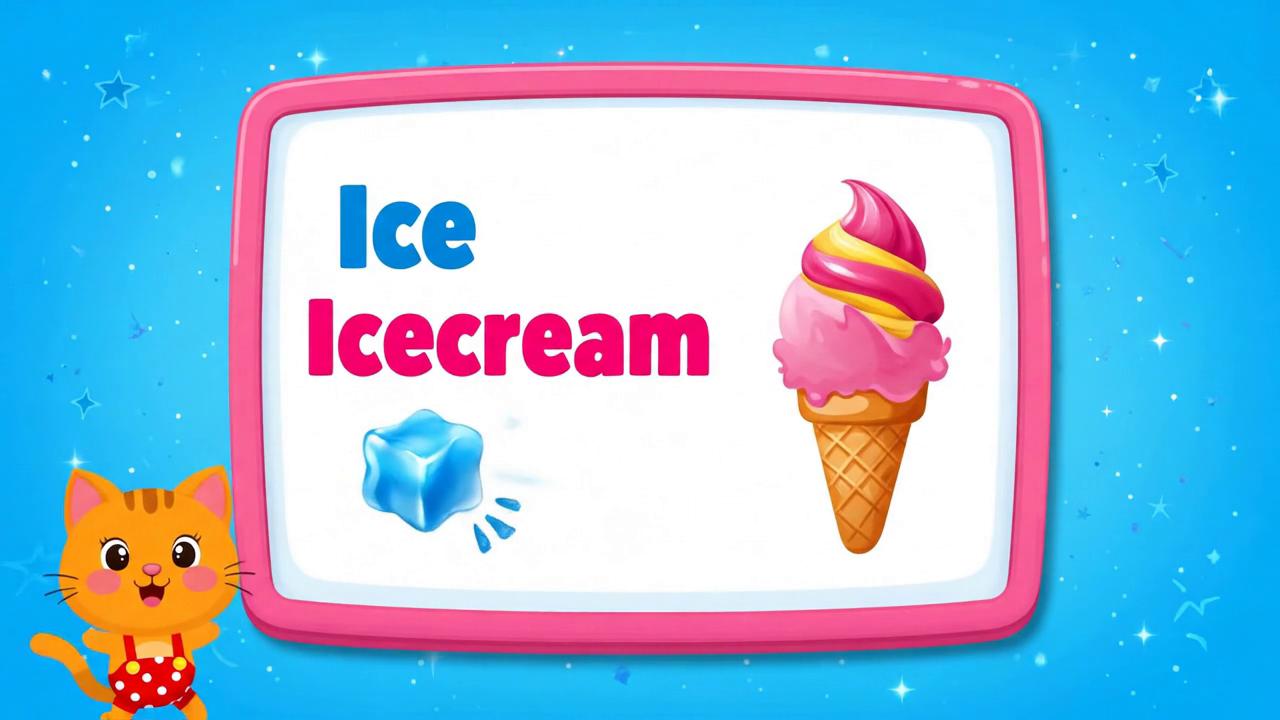} \\
\end{tabular} \\
\vspace{-5mm}
\footnotesize\emph{Multi-scene-Animation: The video is an educational animation designed for young children\dots}
\end{tabularx}
\vspace{-3mm}
\caption{Example videos generated by \model{}, showing improvements in visual fidelity, camera focus, smooth transitions, compelling storylines, and sounds. Optimized prompts and more examples are in our project page.}\label{fig:teaser}
\end{table}

\centerline{\small{\textbf{Project page:} \url{https://g-vista.github.io/}}}

\section{Introduction}
Text-to-video (T2V) generation has seen significant progress, with models like Veo 3 \citep{deepmind2025veo3} demonstrating a remarkable ability to generate coherent, high-quality video and audio from text. This progress positions T2V as a powerful tool for applications in creative storytelling, education, and content creation \citep{liu2024sora}. However, broader deployment is limited by several persistent challenges. Models often struggle with precise alignment with user goals, consistent adherence to physical laws and commonsense \citep{bansal2024videophy}, and a high sensitivity to the exact phrasing of input prompts. This sensitivity, in particular, forces users into a laborious trial-and-error cycle, requiring them to repeatedly tweak phrasing and filter outputs to achieve a desirable result.

Meanwhile, test-time optimization frameworks have shown promise in automatically improving generation quality and preference alignment for both text and image generation \citep{madaan2023self,hao2023optimizing,pryzant-etal-2023-automatic,snell2024scaling,muennighoff2025s1}. However, extending them to autonomous video generation presents substantially challenges. Unlike text or image data, videos unfold across multiple scenes, modalities, and high-level contextual meaning, making evaluation and optimization significantly more complex. Existing efforts often target only specific video properties, such as objects \citep{gao2025devil}, harmless-accurate-helpful traits \citep{cheng2025vpo}, or visual-reward fine-tuning \citep{ji2024prompt,soni2024videoagent,dalal2025one}. Yet, to the best of our knowledge, no studies have unified visual, audio, and contextual quality in a single optimization framework, despite these together being key to user satisfaction. 

We introduce \model{} (\underline{V}ideo \underline{I}terative \underline{S}elf-improvemen\underline{T} \underline{A}gent), a novel multi-agent framework that self-improves video-audio generation at test time. Inspired by how humans evaluate videos and refine prompts, \model{} jointly optimizes three key aspects of videos: Visual, Audio, and Context, through collaborative agents. This process is guided by a comprehensive and configurable suite of evaluation metrics tailored to each, and is driven by four key components: \textbf{(i) Structured Video Prompt Planning} (\Cref{subsec:initialization}), which transforms user input into temporally grounded, multi-scene, and multi-aspect descriptions; \textbf{(ii) Pairwise Tournament Selection} (\Cref{subsec:initialization}), a probing-driven algorithm to identify the best video-prompt candidates; \textbf{(iii) Multi-Dimensional Multi-Agent Critiques} (\Cref{subsec:selfimprovement}), a triadic agent system that provides nuanced critiques motivated by the {Jury Decision Process} \citep{klevorick1979model}; and a \textbf{(iv) Deep Thinking Prompting Agent}, which performs human-like introspective, structured reasoning to revise prompts in a targeted way. We rigorously evaluate \model{} on a widely used single-scene benchmark and an internal multi-scene benchmark with rich instructions. Experiments show that \model{} substantially outperforms prior test-time optimization methods, improving state-of-the-art (SOTA) T2V models like Veo 3 by up to 60\% under our metrics, further validated by human evaluations. Our work enables more reliable, user-aligned text-to-video generation and paves the way for broader video synthesis applications. In summary, this paper makes the following contributions:
\begin{itemize}
    \item We propose \model{}, a novel multi-agent framework that emulates human-like prompt refinement to improve T2V generation. To the best of our knowledge, \model{} is the first to jointly improve the visual, audio, and context dimensions of videos. See \Cref{fig:teaser} for its exemplary generated videos.
    \item We develop \model{}'s components and meticulously design their configurable evaluation metrics that enable fully autonomous, model-driven video evaluation and refinement.
    \item We conduct extensive experiments supported by in-depth analysis and human studies showing that \model{} consistently outperforms existing baselines and improves user preferences.  
\end{itemize}


\section{\model{}} \label{sec:method}

\begin{figure}[htbp]
    \centering
\includegraphics[width=1\linewidth]{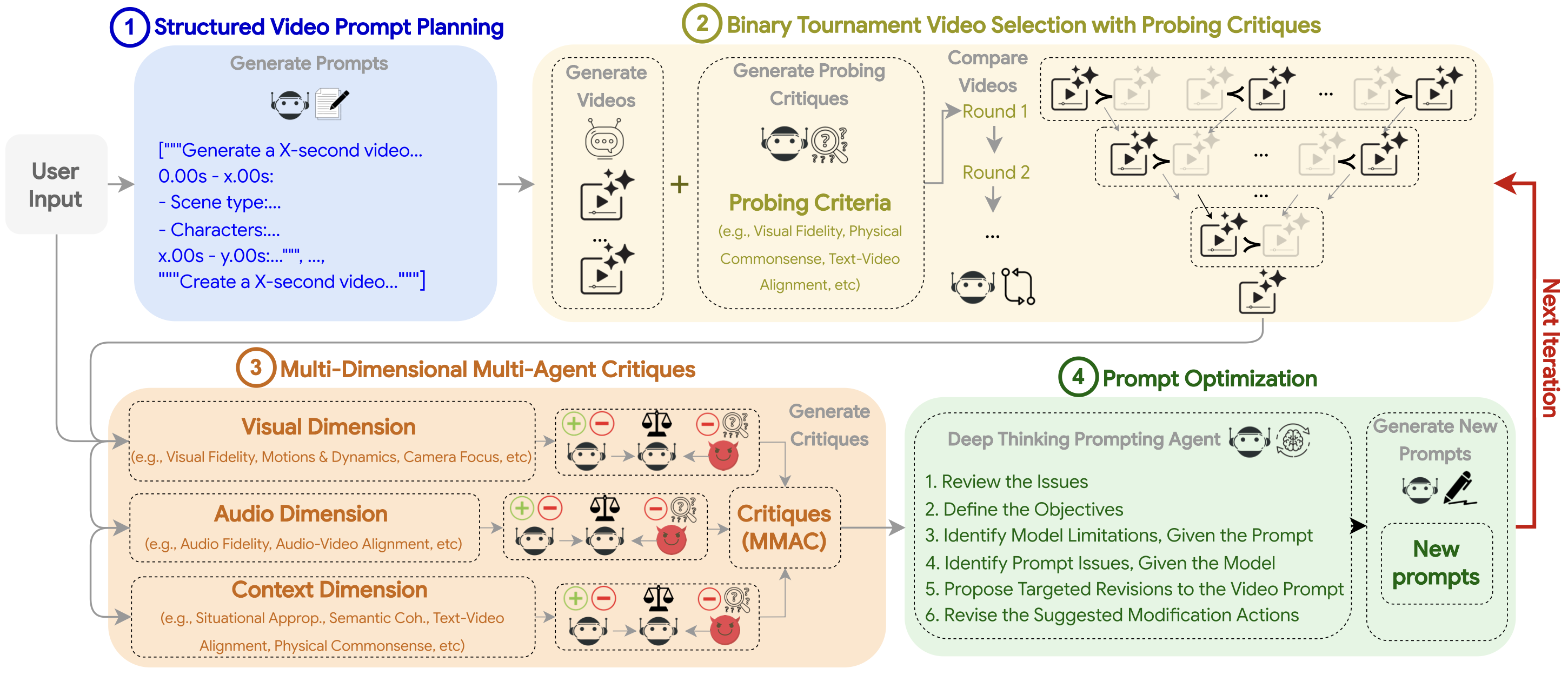} 
\vspace{-10mm}
\caption{The workflow of our proposed multi-agent framework, \model{}. \includegraphics[width=0.65cm]{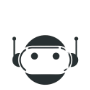}: MLLM Agent; \includegraphics[width=0.4cm]{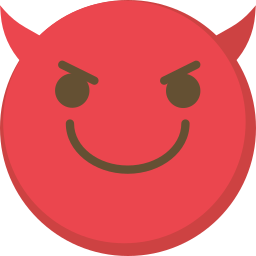} : Adversarial MLLM Agent; \includegraphics[width=0.45cm]{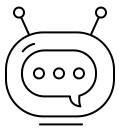} : Video Generation Agent.}
    \label{fig:method-overview}
\end{figure}

\model{} (\Cref{fig:method-overview} and \Cref{alg:mavpo}) is a modular, configurable framework for optimizing text-to-video generation. Given a user video prompt $P$, it produces an optimized video $V^*$ and its refined prompt $P^*$ through two phases: \textbf{(i) Initialization} and \textbf{(ii) Self-Improvement}, inspired by the human video optimization process via prompting. During (i), the prompt is parsed and planned into variants to generate candidate videos (\textbf{Step 1}), after which the best video-prompt pair is selected (\textbf{Step 2}). In (ii), the system generates multi-dimensional, multi-agent critiques (\textbf{Step 3}), refines the prompt (\textbf{Step 4}), produces new videos, and reselects the champion pair (\textbf{Step 2}). This phase continues until a stopping criterion is met or the maximum number of iterations is reached.


\subsection{Initialization Phase} \label{subsec:initialization}

\paragraph{Step 1. Structured  Video Prompt Planning (\Cref{alg:mavpo}-L1).}
The user video generation prompt $P$ is first parsed into $m$ \textbf{timed sequences of scenes} as prompt candidates where each candidate is defined as $P_i :
= \text{[}S_{i,1}, S_{i,2}, \dots\text{]}$. By default, each scene configuration consists of nine properties spanning context, visual, and audio dimensions:
\textbf{ (1) Duration:} The length (in seconds) of the scene; \textbf{Scene Type:} The type of the scene (e.g., action, montage); \textbf{(3) Characters:} Any key figures or entities on screen, whether human, animal, or object, that drive the scene (e.g., a dog, a flower); \textbf{(4) Actions:} Any deliberate and meaningful movements or behaviors that the characters above perform (e.g., the flower is blooming); \textbf{(5) Dialogues:} Any spoken lines, voiceovers, or on-screen text that drive the scene; \textbf{(6) Visual Environment:} Any environmental backdrop or setting that establishes the scene's world (e.g., a tranquil canopy of boundless sky); \textbf{(7) Camera:} Any cinematographic technique, including framing, movement, or angle, that shapes the scene (e.g., a close-up capturing subtle emotion); \textbf{(8) Sounds:} Any auditory elements, such as soundtrack, voiceovers, or ambient noises, that enrich the scene (e.g., a crashing waves in the background); \textbf{(9) Moods:} Any prevailing tone or atmosphere that powers the scene (e.g., serene).

\model{} implements a configurable interface over the above properties and planning constraints on scenes (see below). This introduces two main novelties over prior work \citep{Huang_2024_CVPR,google_vertex_video_prompt_guide,polyak2025moviegencastmedia}: \textbf{(i) temporal scene-level decomposition}, where user prompts are structured as semantically coherent segments to support reasoning over complex content, and \textbf{(ii) fine-grained multi-modal video prompting}, enabling automated multi-dimensional critiques and self-improvement. A multi-modal LLM (MLLM) is used to obtain $S_{i,j}$ and, when needed, infer missing scene properties. To preserve fidelity to the user's intent and avoid scenario drift, the system by default enforces planning constraints on \emph{Realism, Relevancy, and Creativity}, though they are configurable and optional: videos are grounded in real-world physics unless the prompt specifies otherwise (e.g., animated or fantastical); only elements explicitly stated or implied are included, avoiding unnecessary invention; ambient sounds and effects are encouraged when beneficial; and excessive scene transitions are discouraged for simple or short prompts. {See \Cref{appdx:prompt-step-1} for defaults.} Finally, we retain a residual $P$ in the sampled set $\mathcal{P}$ to allow models that do not benefit from decomposition.

\begin{algorithm}[t!]
\caption{\model{}: Configurable Self-Improving Video Generation Agent}
\label{alg:mavpo}
\begin{algorithmic}[1]
\Require User prompt $P$, a text-to-video model $\text{T2V()}$, a multi-modal $\text{MLLM}()$, \#iteration $T$, optional early-stop $m$, optional configurable criteria for video selection {$\mathcal{M}^S_{user}$} and critiques {$\mathcal{M}^C_{user}$}
\Ensure Optimized video output $V^*$ with its prompt $P^*$
\Statex \emph{\textbf{[Initialization Phase]}}
\State $\text{Plan and generate video prompts } \mathcal{P} := \{P_1, \dots, P_m, {P}\} \gets \text{\Call{PromptPlanner}{$P$}}$
\State $\text{Generate video candidates } \mathcal{V} := \{V_1, \dots, V_n\}, \text{ where } \{V_i, \dots\} \gets \text{\Call{T2V}{$P_j$}}$
\State $\text{Select best video-prompt pair } \text{({$V^*, P^*$})} \gets \text{\Call{PairwiseSelect}{$\mathcal{V}, \mathcal{P}, P, {\mathcal{M}^S_{user}}$}}$
\Statex \emph{\textbf{[Self-Improvement Phase]}}
\For{$t = 1$ to $T$}
    \State Generate multi-dimensional multi-agent critiques $\mathcal{F}^t \gets \text{\Call{MMAC}{{$V^*$}, {$P^*$}, $P$, ${\mathcal{M}^C_{user}}$}}$
    \State Optimize prompts $\mathcal{P}^t := \{P^t_1,\dots,P^t_m, {P^*}\} \gets \text{\Call{PromptOptimizer}{{$P^*$}, $P$, $\mathcal{F}^t$}}$ 
    \State $\text{Generate video candidates } \mathcal{V}^t := \{V^t_1, \dots, V^t_n, {V^*}\}, \text{ where } \{V^t_i, \dots\} \gets \text{\Call{T2V}{$P^t_j$}}$
    \State $\text{Update best video-prompt pair } \text{({$V^*, P^*$})} \gets \text{\Call{PairwiseSelect}{$\mathcal{V}^t, \mathcal{P}^t, P, {\mathcal{M}^S_{user}}$}}$
    \If{$\text{({$V^*, P^*$})}$ does not change after $m$ iterations}
        \State \textbf{break}
    \EndIf
\EndFor
\State \Return $\text{({$V^*, P^*$})}$
\end{algorithmic}
\end{algorithm}



\paragraph{Step 2. Binary Tournament Video Selection with Probing Critiques (\Cref{alg:mavpo}-L3, L10).} 
After obtaining candidate video-prompt pairs, this step selects the pair ($V^*$, $P^*$) with the highest video quality for self-improvement. Conventional video evaluation methods often rely on complex metric-based systems \citep{Huang_2024_CVPR}, which are computationally expensive in practice \citep{zhang2024evaluation}. Inspired by recent advances demonstrating the strong video understanding capabilities of MLLMs \citep{li2024mvbench, fu2025video}, we employ an MLLM-as-a-Judge to evaluate videos on customizable evaluation criteria  $\mathcal{M}^S_{user}$. However, scoring videos without ground-truth references is inherently subjective and unreliable. We adopt a video pairwise comparison strategy, which better aligns with human preferences in reinforcement learning and language tasks, while avoiding model-induced biases \citep{leerlaif, liu2024aligning}. As detailed in \Cref{alg:pairwise_tournament}, we iteratively reduce the set of candidate videos via \textbf{Binary Tournaments} \citep{miller1995genetic}. At each iteration, videos are grouped into pairs and compared bidirectionally via swapping; only the winning videos advance to the next round. We swap the videos to avoid token biases  \citep{zheng2024large}.

Comparing $\text{(}V_i, V_j\text{)}$, however, remains unreliable to align with human preferences. To improve this, we employ a criteria-based approach \citep{li2024llms} where the model judges videos across metrics $\mathcal{M}^S_{user}$. We meticulously design a default configuration where  \textbf{${\mathcal{M}^S_{user}} := \{$Visual Fidelity; Physical Commonsense; Text-Video Alignment; Audio-Video Alignment; Engagement}\} capturing the core aspects of video quality, and the video winning on more criteria is the winner. While this approach is more principled, we still observe that model often fails to provide sufficiently critical assessments. We attribute this to the dual burden placed on the model: analyzing videos and comparing them simultaneously. To mitigate this, we introduce a two-step decomposition: first, for each video, the model generates probing critiques ({$\mathcal{Q}$} in \Cref{alg:pairwise_tournament}-L1) on $\mathcal{M}^S_{user}$, then these critiques are used to support the comparisons. {Finally, we implement customizable penalty mechanisms for T2V failures during selection: by default, \model{} penalizes common failures in $\mathcal{M}^S_{user}$ unless the user prompt explicitly specifies otherwise.} The final scores of video candidates $V_i$ and $V_j$ are $s_i$ and $s_j$:

\[
s_i \gets \frac{1}{k} 
  {\text{\LARGE$\Sigma$}_{C\in \mathcal{M}^S_{user}}} 
  \Bigr(\delta\!\left(C, V_i, V_j\right)
  - \lambda \cdot \mathbb{1}\big(C, V_i\big)\Bigr), 
  \quad
  s_j \gets \frac{1}{k} 
  {\text{\LARGE$\Sigma$}_{C\in \mathcal{M}^S_{user}}} 
  \Bigl(1 - \delta\!\left(C, V_i, V_j\right)
  - \lambda \cdot \mathbb{1}\big(C, V_j\big)\Bigr)
\]

where $\delta \big(C, V_i, V_j \big) \in \{0, 0.5, 1\}$ represents the outcome of $V_i$ against $V_j$ on $C$, corresponding to \{Loss, Tie, Win\}, and $\mathbb{1}\big(C, V\big) \in \{0,1\}$ indicates whether $V$ violates $C$ to some extent. The term $\lambda$ is a penalty applied for violations. Noteworthily, $\mathbb{1}\big(C, V\big)$ can be customized to any preferred constraints, not necessarily restricted to $\mathcal{M}^C_{\mathrm{user}}$. {See \Cref{appdx:prompts-step-2} for our default prompts and metrics definitions.}

\begin{algorithm}[t!]
\caption{Pairwise Tournament Selection (\textsc{PairwiseSelect})}
\label{alg:pairwise_tournament}
\begin{algorithmic}[1]
\Require Video prompt $P$, list of videos $\mathcal{V} = \{V_1, \dots, V_n\}$ and their prompts $\mathcal{P} = \{P_1, \dots, P_n\}$, and optional configurable criteria  {$\mathcal{M}^S_{user}$}
\Ensure Best video $V^*$ and its prompt $P^*$
\State $\text{Generate probing critiques } \mathcal{Q} := \{Q_1, \dots, Q_n\} \gets \text{\Call{MLLM}{$\mathcal{V}, \mathcal{P}$, $\mathcal{M}^S_{user}$}}$
\While{$|\mathcal{V}| > 1$}
    \State Group $\mathcal{V}$ into pairs.
    \For{any pair ($V_i, V_j$)}
    \State $\left(V^f_{{win}}, V^f_{{lose}}\right) \gets \text{\Call{MLLM}{$V_i, Q_i, V_j, Q_j$, {$\mathcal{M}^S_{user}$}}}$ \Comment{Forward pairwise comparison.}
    \State $\left(V^s_{{win}}, V^s_{{lose}}\right) \gets \text{\Call{MLLM}{$V_j, Q_j, V_i, Q_i$, {$\mathcal{M}^S_{user}$}}}$ \Comment{Swapped pairwise comparison.}
    \State $\left(V_{win}, V_{lose}\right) \gets \left(V^f_{win}, V^f_{lose}\right)$ if $\left(V^f_{win}, V^f_{lose}\right) == \left(V^s_{win}, V^s_{lose}\right)$ else assign randomly
    \State $\mathcal{V}.\text{remove}\left(V_{{lose}}\right)$; $\mathcal{P}.\text{remove}\left(P_{{lose}}\right)$
    \EndFor
\EndWhile
\State \Return $\text{(}\mathcal{V}\text{[}0\text{]}, \mathcal{P}\text{[}0\text{]}\text{)}$.
\end{algorithmic}
\end{algorithm}

\subsection{Self-Improvement Phase} \label{subsec:selfimprovement}

\paragraph{Step 3. Multi-Dimensional Multi-Agent Critiques
(\Cref{alg:mavpo}-L5).} Given $\left(P, V^*, P^*\right)$, this step elicits targeted critiques to refine the prompt $P^*$. Video evaluation is difficult due to its multi-dimensional nature. We address this with a multi-agent critique framework decomposed into $\mathcal{D}=\{\textbf{Visual, Audio, Context}\}$, each assessed independently by a dedicated system. Evaluation criteria are configured through ${\mathcal{M}^C_{user}}$. While prior work \citep{zheng2024video,liu2024evalcrafter} proposed diverse metrics, recent SOTA models \citep{deepmind2025veo3,wan2025wan} already excel on most of them, suggesting they insufficiently differentiate between high-quality generations (see \Cref{subsec:automatic-eval}). Therefore, we carefully design a comprehensive default configuration for ${\mathcal{M}^C_{user}}$ that expose modality-specific failures even in SOTA T2V models. These are strategically selected and refined from \citep{gao2023high,liu2024evalcrafter,bansal2024videophy,cheng2025mmaudio}:



\begin{itemize}
\item \textbf{Visual}: Visual Fidelity, Motions and Dynamics, Temporal Consistency, Camera Focus, Visual Safety.
\item \textbf{Audio}: Audio Fidelity, Audio-Video Alignment, Audio Safety.
\item \textbf{Context}: Situational Appropriateness, Semantic Coherence, Text-Video Alignment, Physical Commonsense, Engagement, Video Format (Beginning, Ending, Transitions).
\end{itemize}

These include both human-centric criteria such as situational appropriateness and video format, and fine-grained video metrics like visual fidelity\footnote{This set of metrics is more comprehensive than in Step 2 because this step prioritizes in-depth and granular critiques.}. However, we observe that directly employing MLLM-as-a-Judge often yields shallow and unuseful critiques (even when being explicitly asked to be critical). This is because SOTA T2V models such as Veo 3 already produce high-quality outputs that are difficult to critique at surface level even by humans. To address this gap, we introduce \textbf{Multi-Dimensional Multi-Agent Critiques (\textsc{MMAC})}: inspired by the \textbf{Jury Decision Process} \citep{klevorick1979model}, for each evaluation dimension $D \in \mathcal{D}$, we construct a triadic court consisting of a \textbf{Normal Judge} that critically assesses and scores the video on $D$'s metrics in both good and bad faiths, an \textbf{Adversarial Judge} which generates probing questions, counterarguments, and scores to expose video flaws on $D$'s metrics, and a \textbf{Meta Judge} that consolidates the judges from both sides:

\begin{equation}
\begin{aligned}
\{C_D^+, S_D^+\} &\gets J_D^+\left(P, V^*, P^*\right) \quad \text{(Normal Judge)} \\
\{C_D^-, S_D^-\} &\gets J_D^-\left(P, V^*, P^*\right) \quad \text{(Adversarial Judge)} \\
\{C_D^*, S_D^*\} &\gets J_D^*\left(P, C_D^+, S_D^+, C_D^-, S_D^-\right) \quad \text{(Meta Judge)}
\end{aligned}
\end{equation}

The \textsc{MMAC} is $\mathcal{F} := \{C_D^*, S_D^* | D \in \mathcal{D}\}$ where $C_D$ and $S_D$ are metrics' critiques and scores (on a scale of 1-10 \citep{zheng2023judging}). {See \Cref{appdx:prompts-step-3} for default prompts with metrics' definitions}.

\paragraph{Step 4. Prompt Optimization (\Cref{alg:mavpo}-L6).} 
After obtaining $\mathcal{F}$, this step refines $P^*$ via a \textbf{Deep Thinking Prompting Agent (DTPA)}. Direct MLLM optimization often overcomplicates prompts and interprets critiques shallowly (see \Cref{subsec:analyses} for examples); DTPA instead performs a six-step, self-reflective reasoning to suggest prompt modifications in one chain-of-thought: (1) identifying video issues via metrics with low meta scores ($\leq 8$), (2) clarifying the expected outcome and success criteria, (3) evaluating the context sufficiency in the current prompt, (4) determining whether failures stem from model limitations or prompt, and (5) detecting potential conflicts or vagueness within the prompt itself. Based on this introspective analysis, the DTPA proposes a set of modification actions, which are then (6) reviewed and refined to ensure they fully address the failures identified in (1). See \Cref{subsec:optimized-prompts,appdx:deep-thinking-prompting-example} for \model{}'s optimized prompts and \Cref{appdx:prompts-step-4} for method's prompts. 

\begin{equation}
\mathcal{M} := \{M_1,\dots\} \gets \Call{DTPA}{\left(P, P^*, \mathcal{F}\right)} 
\quad \text{($M_i$ are suggested modifications)}
\end{equation}

These modifications are then used to sample improved prompts:
\begin{equation}
\mathcal{P} := \{P_1,\dots,P_n, P^*\} \gets \text{MLLM}\!\left(P, P^*, \mathcal{M}\right)
\end{equation}

\section{Related Work}

\subsection{Text-To-Video Synthesis and Optimization}
Recent years have seen major advances in T2V synthesis \citep{hong2023cogvideo,openai2024sora,wan2025wan,polyak2025moviegencastmedia,deepmind2025veo3}, with SOTA models such as Sora and Veo 3 receiving widespread attention--Veo 3 notably pioneering high-quality audio-video generation. Yet, current models remain highly prompt-sensitive. Existing prompt optimization and test-time self-evolving (agentic) methods, while being successful in other domains \citep{schulhoff2024prompt,long-etal-2025-makes,manas2024improving,wan2025maestro,gao2025survey,fang2025comprehensive}, are limited in video optimization, often requiring white-box access or fine-tuning. For example, VideoAgent \citep{soni2024videoagent} refines video plans by executing them online and using successful trajectories to fine-tune the generation model, while MotionPrompt \citep{nam2025optical} learns token embeddings to improve motion fidelity, and RAPO \citep{gao2025devil} rewrites and augments prompts but relies on target prompts during training. Closest to our work, VPO \citep{cheng2025vpo} optimizes for harmlessness, accuracy, and helpfulness, but not at test time. Meanwhile, LM-powered multi-agent systems such as Mora \citep{yuan2024mora} and FilmAgent \citep{xu2025filmagent} tackle tasks such as scriptwriting and cinematography, yet omit test-time optimization. \model{} combines these directions and is the first to explore black-box prompt optimization for video generation.

\subsection{Video and Audio Generation Evaluation}
Video generation is uniquely challenging to evaluate: there is no single, definitive ``ground truth'', and videos inherently span multiple dimensions that require complex and holistic reasoning to evaluate effectively. Conventional visual quality metrics, such as Inception Score (IS) \citep{salimans2016improved}, FID \citep{heusel2017gans}, and CLIP-Score \citep{hessel-etal-2021-clipscore}, typically  operate along a single dimension and do not to provide a comprehensive evaluation. Recent video benchmarks such as T2I-CompBench \citep{huang2023t2i}, VBench \citep{Huang_2024_CVPR}, and EvalCrafter \citep{liu2024evalcrafter} offer multi-dimensional evaluations tailored to specific model capabilities. However, they still heavily rely on conventional single-metric measures that are time-intensive, inefficient for autonomous improvement, and exclude audio evaluation. Audio-visual extensions like TAVGBench \citep{mao2024tavgbench} and ACVUBench \citep{yang2025audiocentricvideounderstandingbenchmark} illustrate complementary but rigid approaches: the former emphasizes embedding similarity without failure-focused reasoning, while the latter focuses on evaluating MLLM understanding capabilities. Closest to our multi-agent critiques approach are VideoScore \citep{he-etal-2024-videoscore} and Evaluation Agent \citep{zhang2024evaluation}, nevertheless, they are not failure-focused and overlook audio dimensions. \model{} emphasizes failure-sensitive visual and audio metrics even for SOTA T2V models, aiming to efficiently enhance AI-generated videos.

\section{Experiments}\label{sec:experiments}
\paragraph{Benchmarks.} 
We evaluate our approach and baselines on two benchmarks representing distinct scenarios: \textbf{single-scene} and \textbf{multi-scene} generation. For single-scene evaluation, we follow \citet{dalal2025one} to use {MovieGenVideo} \citep{polyak2025moviegencastmedia} benchmark via randomly selecting 100 prompts. For multi-scene evaluation, we use 161 prompts with at least two scenes from our internal dataset covering diverse topics and matching the T2V model's duration requirements.

\paragraph{Models and Baselines.} 
\model{} employs two core components: a multimodal large language model (MLLM) and a text-to-video generation model (T2V). For our experiments, we use \textbf{Gemini 2.5 Flash} (\texttt{Gemini-2.5-flash-preview-05-20}) \citep{google2025gemini2.5} as the MLLM and \textbf{Veo 3} (\texttt{Veo-3.0-generate-preview}) \citep{deepmind2025veo3} as the video generator because they are among the current state of the art. We also experiment with a less powerful T2V model (\textbf{Veo 2} \citep{veo2googledeepmind}) in \Cref{subsec:analyses}. Since there is almost no test-time prompt optimization or multi-agent system for video generation up-to-date, we compare \model{} with four baselines: \textbf{(1) Direct Prompting (DP)}, which directly uses the user prompt as input for video generation; \textbf{(2) Visual Self-Refine (VSR) \citep{madaan2023self}\footnote{https://github.com/madaan/self-refine/blob/main/colabs/Visual-Self-Refine-GPT4V.ipynb}}, which leverages the MLLM to iteratively evaluate the video generated by the T2V model and subsequently refine the prompt; \textbf{(3) Rewrite \citep{google_vertex_video_prompt_guide}}, which involves using the MLLM to rewrite the user prompt using the Vertex AI video generation prompt guidelines provided by Google; \textbf{(4) VPO \citep{cheng2025vpo}}, which expands the user prompts based on three core principles of harmlessness, accuracy, and helpfulness.

We run \model{} for 5 iterations: 1 initialization followed by 4 self-improving steps. In both phases, we sample 5 prompts, each with 3 variants, and generate 2 videos per prompt, resulting in ~30 videos per iteration. Since Rewrite and VPO operate in a single iteration, we scale their \#videos by matching these used in \model{}'s four self-improvement iterations. Likewise, we scale VSR to match \model{}'s total \#videos, denoted as \textbf{Visual Self-Refine++ (VSR++)}. For scaled baselines, the best video per iteration is chosen via binary tournament with bidirectional pairwise comparisons (\Cref{appdx:simple-pairwise-comparison}).

\subsection{Automatic Evaluations}\label{subsec:automatic-eval}

\paragraph{Main Evaluations.}
We evaluate using both MLLM-as-a-Judge and conventional metrics. Following \citet{dalal2025one}, we conduct pairwise comparisons between our method and baselines, using {Gemini 2.5 Flash} \citep{google2025gemini2.5} as the evaluator for its SOTA video-audio understanding and multi-video processing. We assess ten criteria: \textbf{Visual Fidelity, Motions, Temporal Consistency (scene level), Text-Video Alignment, Audio Quality, Audio-Video Alignment, Situational Appropriateness, Physical Commonsense, Engagement, and Video Format (beginning, ending, transitions)}. Each comparison is {bidirectional}, with outcomes recorded as \textbf{Win/Tie/Loss} and \textbf{$\Delta =$ Win-Loss}; conflicts after swapping are marked as Ties. A video wins if it excels in at least three criteria and does not lose on Text-Video Alignment. {See \Cref{subsec:automatic-eval-prompts} for prompts and metric definitions.} We also report conventional metrics: \textbf{IS} \citep{salimans2016improved}, \textbf{CLIP-Score} \citep{hessel-etal-2021-clipscore}, eight visual metrics from \textbf{VBench} \citep{Huang_2024_CVPR}, and three audio metrics from \textbf{NISQA} \citep{mittag21_interspeech}. Additional results with other evaluators are in \Cref{appdx:additonal-results-qwen,appdx:additonal-results-geminipro}.

\setlength{\tabcolsep}{3pt} 
\begin{table}[t!]
\centering
\resizebox{\textwidth}{!}{%
\begin{tabular}{
l
c c c >{\columncolor{lavender}}c !{\vrule width 0.8pt}
c c c >{\columncolor{lavender}}c !{\vrule width 0.8pt}
c c c >{\columncolor{lavender}}c !{\vrule width 0.8pt}
c c c >{\columncolor{lavender}}c !{\vrule width 0.8pt}
c c c >{\columncolor{lavender}}c
}\toprule
& \multicolumn{4}{c}{Init} &  \multicolumn{4}{c}{2} & \multicolumn{4}{c}{3} & \multicolumn{4}{c}{4} & \multicolumn{4}{c}{5} \\
\midrule
 \textbf{Method} & Win & Tie & Loss & $\Delta$ & Win & Tie & Loss & $\Delta$ & Win & Tie & Loss & $\Delta$ & Win & Tie & Loss & $\Delta$ & Win & Tie & Loss & $\Delta$ \\
\midrule
\multicolumn{21}{c}{\cellcolor{white!30}\textbf{Baseline vs. Direct Prompting}} \\

\emph{VSR} & 22.4 & 61.2 & 16.4 & 6.0  & 34.9 & 42.4 & 22.7 & 12.2 & 23.4 & 48.4 & 28.1 & -4.7 & 26.6 & 51.6 & 21.9 & 4.7 & 24.6 & 47.7 & 27.7 & -3.1 \\
\emph{VSR++} & \underline{28.6}$\dagger$ & \underline{67.3}$\dagger$ & \underline{4.1}$\dagger$ & \underline{\textbf{24.5}}$\dagger$ & \underline{31.1}$\dagger$ 
& \underline{42.2}$\dagger$ & \underline{26.7}$\dagger$ & \underline{4.4}$\dagger$ & \underline{29.5}$\dagger$ & \underline{52.3}$\dagger$ & \underline{18.2}$\dagger$ & \underline{11.3}$\dagger$ & \underline{26.8}$\dagger$ & \underline{46.3}$\dagger$ & \underline{26.8}$\dagger$ & \underline{0.0}$\dagger$ & \underline{33.3}$\dagger$ & \underline{53.3}$\dagger$ & \underline{13.3}$\dagger$ & \underline{20.0}$\dagger$ \\
\emph{Rewrite}      & 19.0 & 69.0 & 12.0 & 7.0  & 35.0$\dagger$ & 52.0$\dagger$ & 13.0$\dagger$ & 22.0$\dagger$ & 30.0$\dagger$ & 56.0$\dagger$ & 14.0$\dagger$ & 16.0$\dagger$ & 39.0$\dagger$ & 47.0$\dagger$ & 14.0$\dagger$ & 25.0$\dagger$ & 27.0$\dagger$ & 65.0$\dagger$ & 8.0$\dagger$ & 19.0$\dagger$ \\
\emph{VPO}          & 29.0 & 46.0 & 25.0 & 4.0  & 34.0$\dagger$ & 56.0$\dagger$ & 10.0$\dagger$ & 24.0$\dagger$ & 31.0$\dagger$ & 55.0$\dagger$ & 14.0$\dagger$ & 17.0$\dagger$ & 28.0$\dagger$ & 65.0$\dagger$ & 7.0$\dagger$ & 21.0$\dagger$ & 36.0$\dagger$ & 56.0$\dagger$ & 8.0$\dagger$ & 28.0$\dagger$ \\
\cmidrule{1-21}
\textbf{\model{}}  & \textbf{35.5} & 50.1 & 14.4 & 21.1 & \textbf{40.7} & 49.4 & 9.9 & \textbf{30.8} & \textbf{41.4} & 43.7 & 14.9 & \textbf{26.5} & \textbf{42.4} & 43.5 & 14.1 & \textbf{28.3} & \textbf{45.9} & 50.2 & 13.9 & \textbf{32.0} \\
\midrule
\multicolumn{21}{c}{\cellcolor{white!30}\textbf{\model{} vs. Baselines}} \\
\emph{VSR}  & 47.0 & 36.4 & 16.7 & 30.3 & 60.0 & 35.0 & 5.0 & 55.0 & 45.6 & 43.9 & 10.5 & 35.1 & 38.6 & 47.4 & 14.0 & 24.6 & 45.6 & 35.1 & 19.3 & 26.3 \\
\emph{VSR++} & \underline{30.4}$\dagger$ & \underline{51.1}$\dagger$ & \underline{18.5}$\dagger$ & \underline{11.9}$\dagger$ & \underline{48.8}$\dagger$ & \underline{36.2}$\dagger$ & \underline{15.0}$\dagger$ & \underline{33.8}$\dagger$ & \underline{42.3}$\dagger$ & \underline{42.3}$\dagger$ & \underline{15.4}$\dagger$ & \underline{26.9}$\dagger$ & \underline{50.7}$\dagger$ & \underline{29.6}$\dagger$ & \underline{19.7}$\dagger$ & \underline{31.0}$\dagger$ & \underline{30.3}$\dagger$ & \underline{57.9}$\dagger$ & \underline{11.8}$\dagger$ & \underline{18.5}$\dagger$ \\
\emph{Rewrite}      & 34.0 & 51.6 & 14.4 & 19.6 & 34.8$\dagger$ & 45.7$\dagger$ & 19.6$\dagger$ & 15.2$\dagger$ & 35.6$\dagger$ & 54.4$\dagger$ & 10.0$\dagger$ & 25.6$\dagger$ & 38.2$\dagger$ & 42.7$\dagger$ & 19.1$\dagger$ & 19.1$\dagger$ & 40.2$\dagger$ & 41.4$\dagger$ & 18.4$\dagger$ & 21.8$\dagger$ \\
\emph{VPO}          & 27.8 & 66.7 & 5.6 & 22.2 & 43.8$\dagger$ & 50.0$\dagger$ & 6.2$\dagger$ & 37.6$\dagger$ & 40.0$\dagger$ & 60.0$\dagger$ & 0.0$\dagger$ & 40.0$\dagger$ & 53.3$\dagger$ & 40.0$\dagger$ & 6.7$\dagger$ & 46.6$\dagger$ & 35.7$\dagger$ & 45.7$\dagger$ & 18.6$\dagger$ & 17.1$\dagger$ \\
\bottomrule
\end{tabular}}
\vspace{0.5em}
\small (a) Results in single-scene scenarios.
\vspace{0.5em}
\resizebox{\textwidth}{!}{%
\begin{tabular}{
l
c c c >{\columncolor{lavender}}c !{\vrule width 0.8pt}
c c c >{\columncolor{lavender}}c !{\vrule width 0.8pt}
c c c >{\columncolor{lavender}}c !{\vrule width 0.8pt}
c c c >{\columncolor{lavender}}c !{\vrule width 0.8pt}
c c c >{\columncolor{lavender}}c
}
\toprule
& \multicolumn{4}{c}{Init} & \multicolumn{4}{c}{2} & \multicolumn{4}{c}{3} & \multicolumn{4}{c}{4} & \multicolumn{4}{c}{5} \\
\midrule
\textbf{Method} & Win & Tie & Loss & $\Delta$ & Win & Tie & Loss & $\Delta$ & Win & Tie & Loss & $\Delta$ & Win & Tie & Loss & $\Delta$ & Win & Tie & Loss & $\Delta$ \\
\midrule
\multicolumn{21}{c}{\cellcolor{white!30}\textbf{Baseline vs. Direct Prompting}} \\
\emph{VSR} & 21.2 & 63.5 & 15.4 & 5.8 & 33.3 & 40.4 & 26.3 & 7.0 & 36.7 & 49.0 & 14.3 & 22.4 & 33.7 & 52.0 & 14.3 & 19.4 & 35.3 & 45.9 & 18.8 & 16.5 \\
\emph{VSR++} & 
\underline{29.4}$\dagger$ & \underline{41.2}$\dagger$ & \underline{29.4}$\dagger$ & \underline{0.0}$\dagger$ & 
\underline{26.5}$\dagger$ & \underline{55.9}$\dagger$ & \underline{17.6}$\dagger$ & \underline{8.9}$\dagger$ & 
\underline{17.6}$\dagger$ & \underline{58.8}$\dagger$ & \underline{23.5}$\dagger$ & \underline{-5.9}$\dagger$ & 
\underline{23.5}$\dagger$ & \underline{52.9}$\dagger$ & \underline{23.5}$\dagger$ & \underline{0.0}$\dagger$ & 
\underline{26.5}$\dagger$ & \underline{52.9}$\dagger$ & \underline{20.6}$\dagger$ & \underline{5.9}$\dagger$ \\
\emph{Rewrite} & 14.3 & 55.1 & 30.6 & -16.3 & 31.8$\dagger$ & 55.7$\dagger$ & 12.5$\dagger$ & 19.3$\dagger$ & 29.6$\dagger$ & 48.9$\dagger$ & 21.6$\dagger$ & 8.0$\dagger$ & 23.9$\dagger$ & 55.7$\dagger$ & 20.5$\dagger$ & 3.4$\dagger$ & 23.9$\dagger$ & 52.3$\dagger$ & 23.9$\dagger$ & 0.0$\dagger$ \\
\emph{VPO} & 25.2 & 53.7 & 21.1 & 4.1 & 38.2$\dagger$ & 52.8$\dagger$ & 9.0$\dagger$ & \textbf{29.2}$\dagger$ & 38.2$\dagger$ & 51.7$\dagger$ & 10.1$\dagger$ & 28.1$\dagger$ & 28.1$\dagger$ & 60.7$\dagger$ & 11.2$\dagger$ & 16.9$\dagger$ & 27.0$\dagger$ & 60.7$\dagger$ & 12.4$\dagger$ & 14.6$\dagger$ \\
\cmidrule{1-21}
\textbf{\model{}}  & \textbf{37.8} & 52.3 & 9.9 & \textbf{27.9} & \textbf{39.4} & 47.2 & 13.4 & 26.0 & \textbf{38.4} & 52.2 & 9.4 & \textbf{29.0} & \textbf{43.7} & 43.7 & 12.6 & \textbf{31.1} & \textbf{46.3} & 42.5 & 11.2 & \textbf{35.1} \\
\midrule
\multicolumn{21}{c}{\cellcolor{white!30}\textbf{\model{} vs. Baselines}} \\
\emph{VSR} & 46.2 & 44.0 & 9.9 & 36.3 & 43.8 & 43.8 & 12.5 & 31.3 & 53.2 & 36.7 & 10.1 & 43.1 & 50.0 & 40.3 & 9.7 & 40.3 & 48.5 & 37.9 & 13.6 & 34.9 \\
\emph{VSR++} & 
\underline{35.3}$\dagger$ & \underline{52.9}$\dagger$ & \underline{11.8}$\dagger$ & \underline{23.5}$\dagger$ & 
\underline{43.8}$\dagger$ & \underline{46.9}$\dagger$ & \underline{9.4}$\dagger$ & \underline{34.4}$\dagger$ & 
\underline{35.0}$\dagger$ & \underline{46.9}$\dagger$ & \underline{18.1}$\dagger$ & \underline{16.9}$\dagger$ & 
\underline{34.4}$\dagger$ & \underline{53.1}$\dagger$ & \underline{12.5}$\dagger$ & \underline{21.9}$\dagger$ & 
\underline{34.4}$\dagger$ & \underline{50.0}$\dagger$ & \underline{15.6}$\dagger$ & \underline{18.8}$\dagger$ \\
\emph{Rewrite} & 33.5 & 59.5 & 8.0 & 25.5 & 32.5$\dagger$ & 67.5$\dagger$ & 0.0$\dagger$ & 32.5$\dagger$ & 36.6$\dagger$ & 57.3$\dagger$ & 6.1$\dagger$ & 30.5$\dagger$ & 37.0$\dagger$ & 58.0$\dagger$ & 4.9$\dagger$ & 32.1$\dagger$ & 42.0$\dagger$ & 51.9$\dagger$ & 6.2$\dagger$ & 35.8$\dagger$ \\
\emph{VPO} & 25.3 & 64.3 & 11.4 & 13.9 & 27.7$\dagger$ & 69.9$\dagger$ & 2.4$\dagger$ & 25.3$\dagger$ & 34.2$\dagger$ & 48.8$\dagger$ & 17.1$\dagger$ & 17.1$\dagger$ & 18.5$\dagger$ & 74.1$\dagger$ & 7.4$\dagger$ & 11.1$\dagger$ & 25.9$\dagger$ & 60.5$\dagger$ & 13.6$\dagger$ & 12.3$\dagger$ \\
\bottomrule
\end{tabular}}
\vspace{0.5em}
\small (b) Results in multi-scene scenarios.
\vspace{-3mm}
\caption{Win/Tie/Loss rates and $\Delta = \text{Win} - \text{Loss}$ across 5 iterations. $\dagger$ refers to our scaled-up results, and \underline{underlines} are results evaluated on half of the benchmark.}
\vspace{-5mm}
\label{tab:main-results}
\end{table}

\vspace{-1mm}
\paragraph{Baselines vs. DP. Results.}
From \Cref{tab:main-results}, the original baselines (without $\dagger$) show inconsistent and often unfavorable win-loss dynamics. For example, in \Cref{tab:main-results}-(a), VPO reaches 29\% win rate with only $\Delta = 4.0\%$, while in the multi-scene setting (\Cref{tab:main-results}-(b)), Rewrite even yields $\Delta = -16.3\%$. This highlights their limited robustness, where gains in some aspects trade off against regressions elsewhere. VSR performs better in multi-scene but adds little in single-scene scenarios, likely because the former is more challenging: DP underperforms in multi-scene, giving VSR room to help, whereas DP's stronger single-scene performance restricts its impact. Enhanced baselines (with $\dagger$) mitigate this somewhat by sampling multiple videos and selecting the best, but show no evidence of continued improvement when scaling \# videos more. By contrast, \model{} consistently outperforms all baselines and scales more effectively with test-time compute, achieving substantial gains with win rates up to 45.9\% ($\Delta = 32\%$) and 46.3\% ($\Delta = 35.1\%$). 

\paragraph{\model{} vs. Baselines Results.}
\model{} achieves significant win rates over baselines, ranging from 27.8-60.0\% (single-scene) and 18.5-53.2\% (multi-scene). Notably, these trends do not correlate with baseline performance relative to DP. For example, while VSR performs best against DP, \model{} surpasses it by the largest margin. This likely reflects the \model{}'s Pareto optimization of multiple video dimensions, whereas baselines typically focus on fewer aspects with narrower coverage.

\begin{figure}[t!]
    \centering
\includegraphics[width=.95\textwidth]{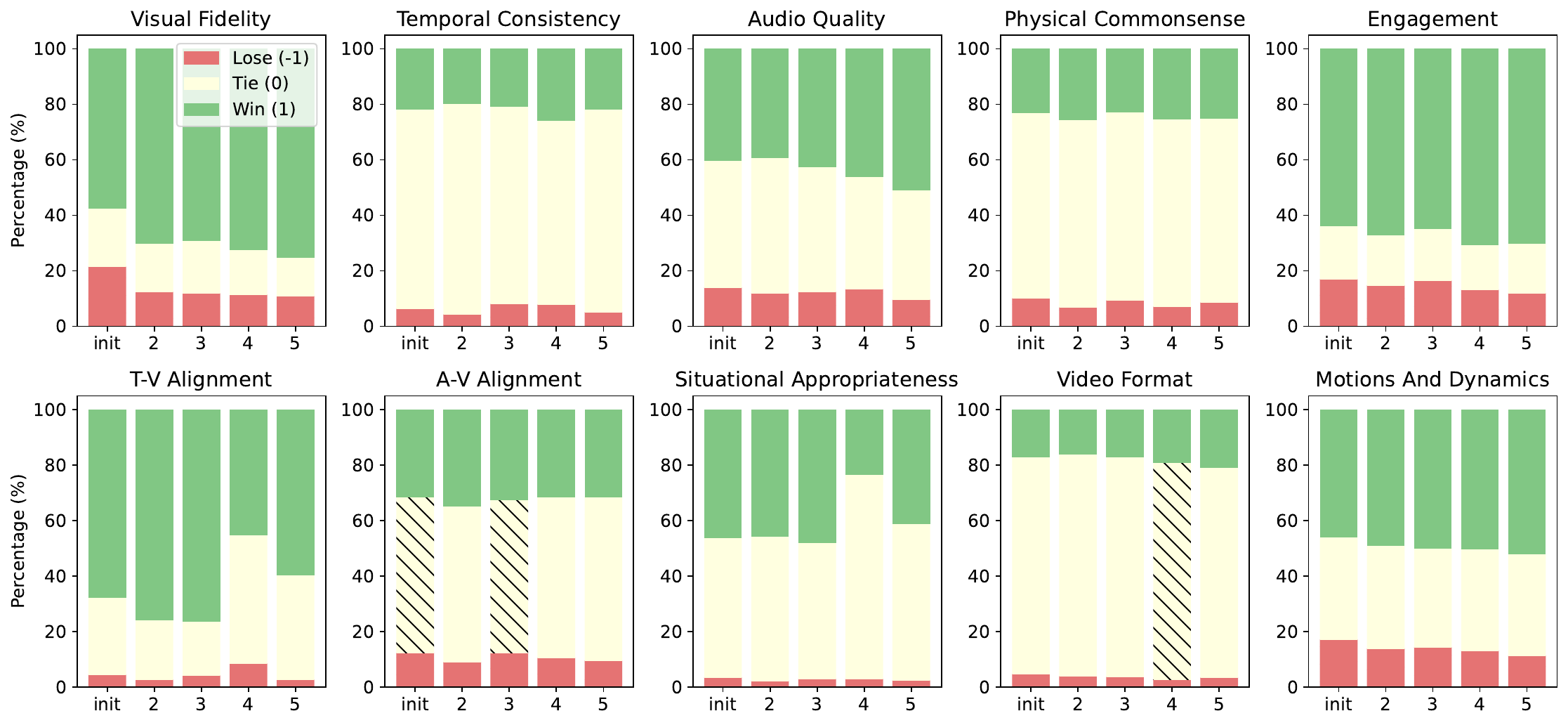}
\vspace{-2mm}
\caption{Average win/tie/lose between \model{} and Direct Prompting (DP) on two benchmarks. The individual benchmark results are in \Cref{appdx:additonal-results-gemini}}
\label{fig:dp_mavpo_finegrained}
\end{figure}

\paragraph{Fine-Grained Results.} 
To better understand the specific metrics where \model{} optimizes, we present a fine-grained pairwise comparison between \model{} and DP in \Cref{fig:dp_mavpo_finegrained}. Results reveal that \model{} achieves significant and consistent improvements across key visual, audio, and context dimensions. The most notable gains appear in Visual Fidelity, Engagement, Text-Video Alignment, Motions and Dynamics, Audio Quality, and Situational Appropriateness, attributable to \model{}'s multi-dimensional critique framework and stringent selection process with constraints and penalties. Finally, \model{} yields only marginal improvements in Temporal Consistency and Video Format, as the Veo 3 model already demonstrates strong performance in these aspects.

\begin{figure}[t!] 
\centering
\includegraphics[width=0.9\textwidth]{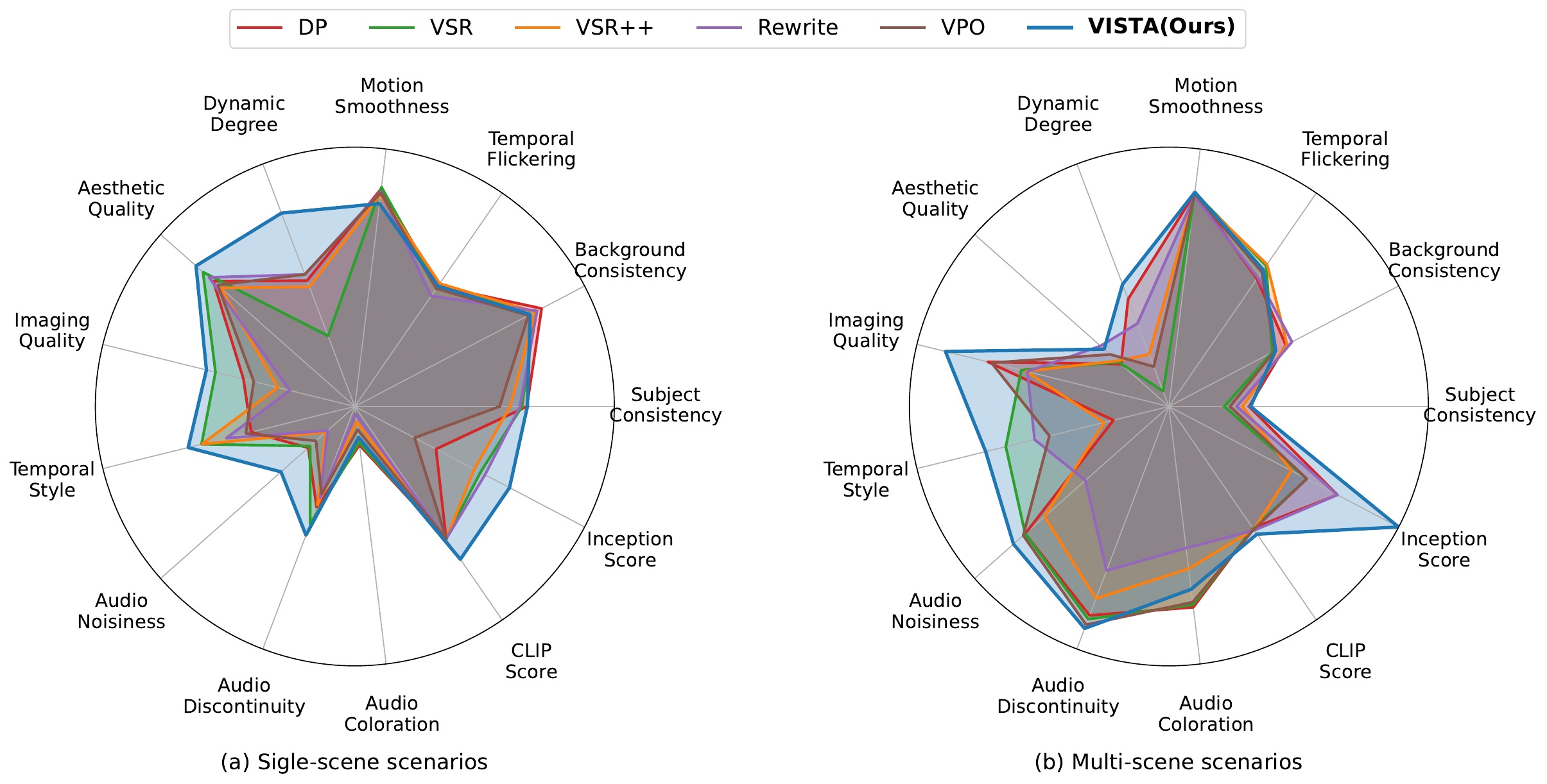} 
\caption{Evaluation results using conventional metrics on single-scene (a) and multi-scene (b) benchmarks. Numerical results are provided in \Cref{tab:meta-results-on-traditional-metrics,tab:ourdata-results-on-traditional-metrics}.}
\vspace{-3mm}
\label{fig:results-conventional-metrics}
\end{figure}

\paragraph{Conventional Metrics Results.} The results on conventional metrics are plotted in \Cref{fig:results-conventional-metrics}. While prior frameworks exhibit no clear improvements on both visual and audio criteria over DP, we find that \model{} delivers notable improvements across key visual dimensions, particularly in Dynamic Degree, Aesthetic Quality, and Temporal Style. For example, it achieves 89.87\% in Dynamic Quality in single-scene scenarios, surpassing the best-performing baseline at 77.22\%, highlighting the \model{}'s effectiveness in optimizing videos with richer and more realistic motion dynamics. In addition, it also reduces audio noisiness (+0.1/5 abs.) and discontinuities (+0.11/5 abs.) across both datasets, and delivers significant gains on CLIP-Score (+3\% abs.) versus the best baselines. These gains closely align with the MLLM-as-a-Judge evaluations: both highlight \model{}'s balanced improvements across visual fidelity, audio quality, and alignment.

\subsection{Human Evaluations} \label{subsec:human-eval}

\paragraph{Main Evaluations.} 
We conduct three human evaluations to assess: (1) if humans prefer \model{}'s outputs over the strongest baselines, (2) to what extent \model{} improves video quality across iterations, and (3) to what extend \model{} improves DP on visual and audio quality. For (1), we randomly sample 50 prompts, 25 from single-scene (\model{} vs. VSR++ at iteration 5) and 25 from multi-scene scenarios (\model{} vs. VSR at iteration 5) and employ five annotators with prompt optimization experience to label each pair as a {Win or Loss}. For (2), three different expert annotators score those 50 full optimization trajectories of both \model{} and VSR, using a 1-5 scale. For (3), we use the same 50 prompts and ask three expert annotators to rate videos on two specific dimensions: Visual Quality and Audio Quality using a 1-5 scale. {See \Cref{appdx:human-eval-learning-score} for our instructions}.

\paragraph{Results.}

\begin{wrapfigure}{r}{0.55\textwidth}
\centering
\vspace{-5mm}
\includegraphics[width=0.55\textwidth]{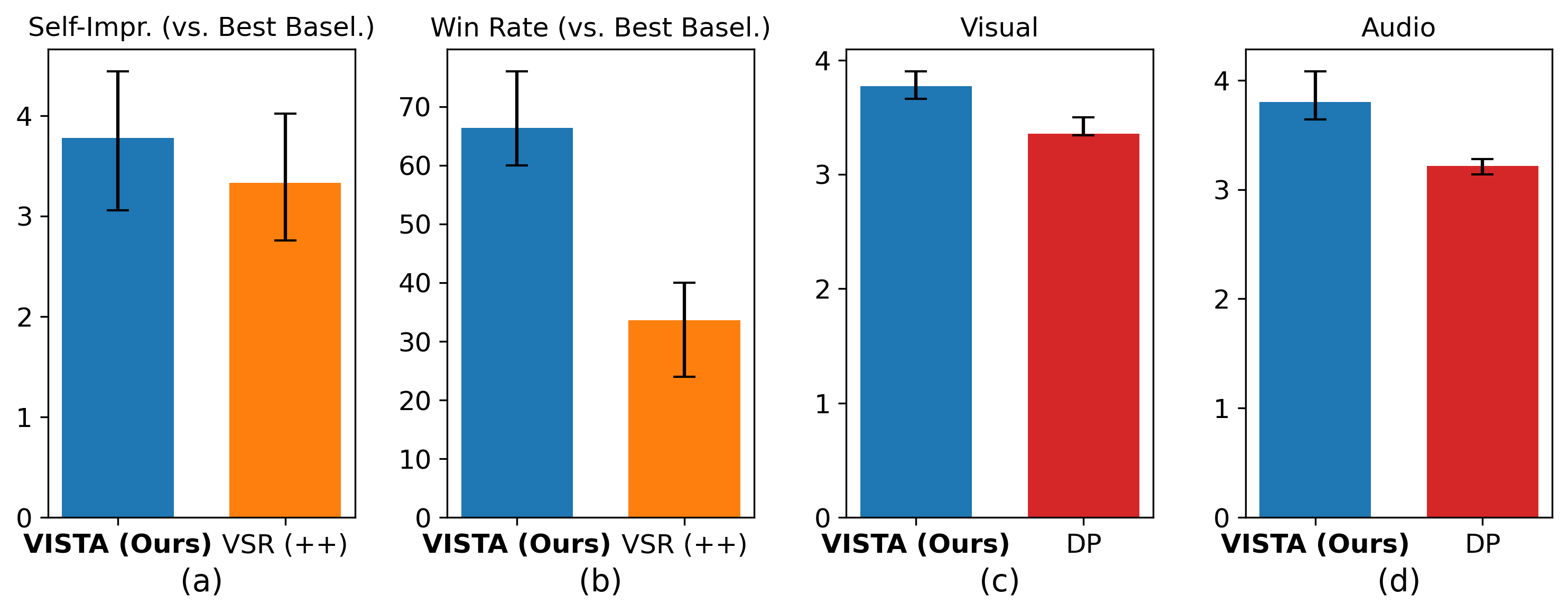}
\vspace{-7mm}
\caption{Summary of human results. The individual annotators' results are in Appx.-\Cref{tab:human-eval-results}}
\vspace{-5mm}
\label{fig:summary_human_eval}
\end{wrapfigure}

From the human evaluation results shown in \Cref{fig:summary_human_eval}, \model{} consistently outperforms baselines across all metrics. The win rate comparison in (b) demonstrates its superiority with 66.4\% versus 33.6\% for best baseline. The self-improvement evaluation in (a) validates that VISTA indeed achieves meaningful self-improvement with an average score of 3.78 out of 5, substantially higher than VSR(++)'s score of 3.33. In addition, the quality assessments in (c) and (d) reveal \model{}'s superiority across both visual and audio dimensions. For visual quality in (c), it improves DP from 3.36 to 3.77, while for audio in (d), it scores 3.47 versus DP's 3.21. Finally, while annotators exhibit moderate variability, this is expected due to the inherently subjective nature of video quality assessment: annotators tend to emphasize different aspects of quality; for example, some place greater weight on visual fidelity, whereas others focus more on detecting awkward physical moments or voices. Nevertheless, all annotators consistently favor the outputs of \model{} over the baselines.

\subsection{Analyses} \label{subsec:analyses}

\paragraph{Cost Analysis.}
\begin{figure}[t!]
  \centering
  \begin{subfigure}[b]{0.49\textwidth}
    \centering
    \includegraphics[width=\textwidth]{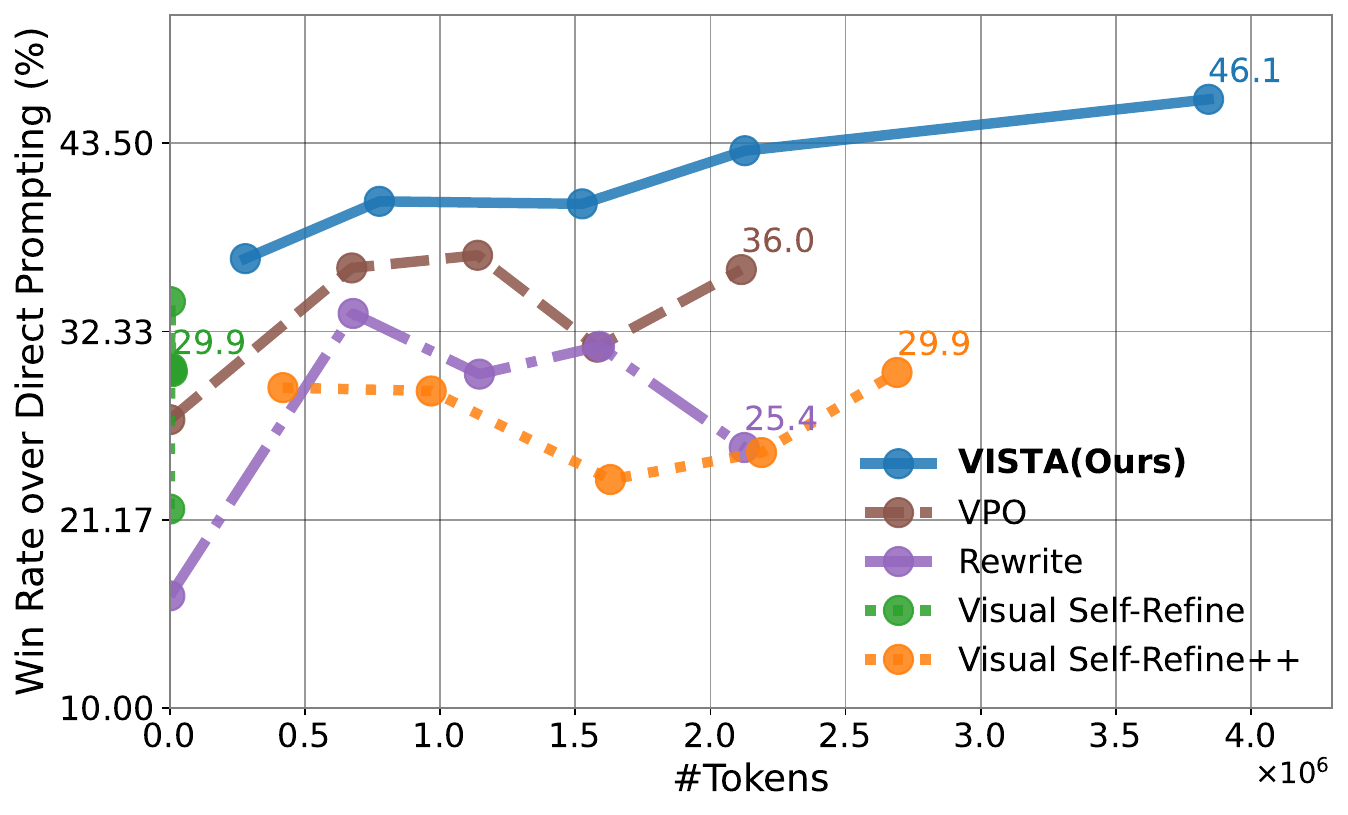}
    \label{fig:numtokens}
  \end{subfigure}
  \begin{subfigure}[b]{0.49\textwidth}
    \centering
    \includegraphics[width=\textwidth]{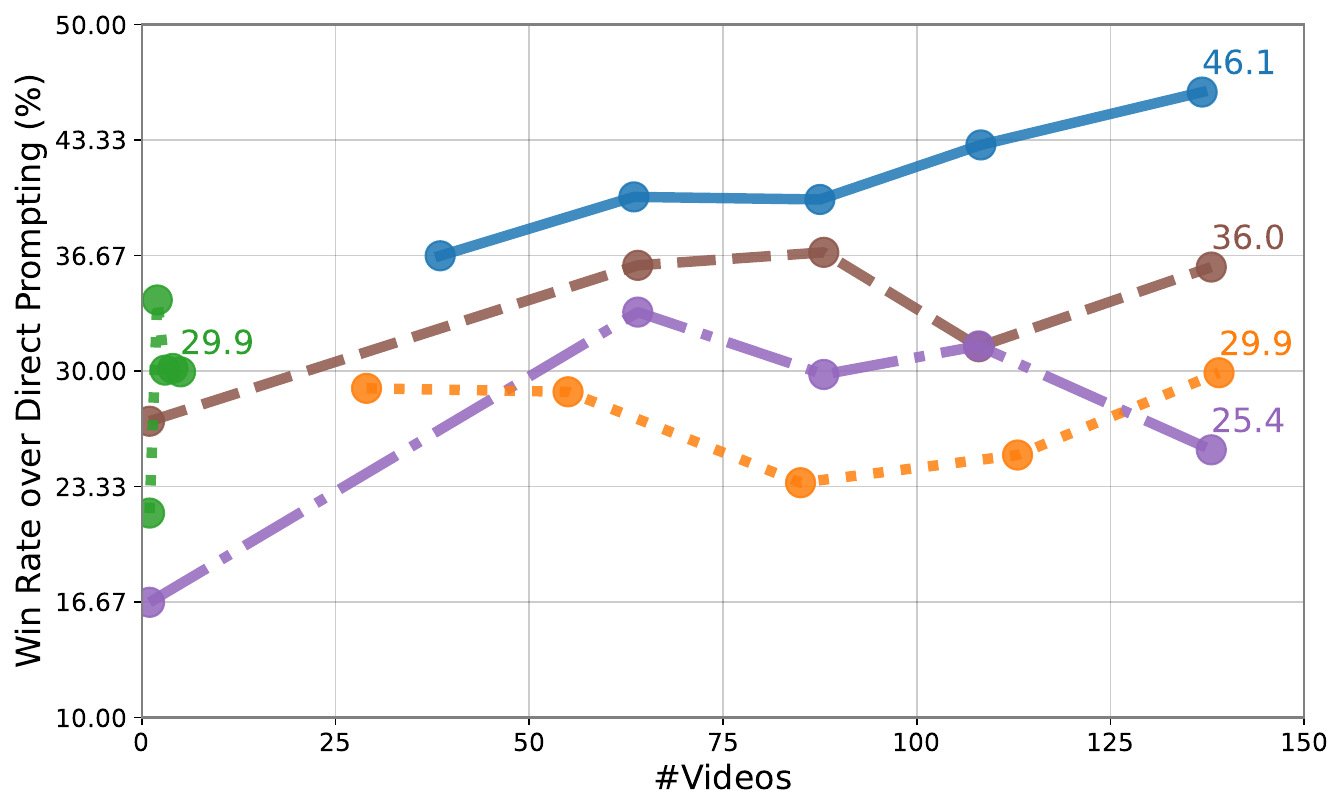}
    \label{fig:numvideos}
  \end{subfigure}
  \vspace{-7mm}
\caption{Cost analysis. Left: total token consumption, including both input and output tokens per iteration. Right: number of newly sampled videos per iteration. Results are averaged over two datasets. Tokens for video generation are unavailable and thus excluded.
}
\label{fig:cost_analysis}
\end{figure}

\Cref{fig:cost_analysis} presents token (left) and video (right) costs averaged over two datasets. Our method shows an upward trend in performance, ultimately reaching an average of 46.1\% win rates as token and video usage increases, with approximately 0.7M tokens and 28 videos consumed per iteration. Most token usage comes from tournament selection, with each video input consuming > 2K tokens. These results suggest that our method has strong potential for further test-time scaling.

\setlength{\tabcolsep}{7pt} 
\begin{table}[t!]
\centering
\resizebox{.9\textwidth}{!}{%
\begin{tabular}{l|ccccc|ccccc}
\toprule
 & \multicolumn{5}{c|}{Single-scene} & \multicolumn{5}{c}{Multi-scene} \\
\midrule
\multicolumn{1}{l|}{Win Rates over DP} & Init & 2 & 3 & 4 & 5 & Init & 2 & 3 & 4 & 5 \\
\midrule
\model{} &  \textbf{35.5} & \textbf{40.7} & {41.4} & {42.4} & \textbf{45.9} & \textbf{37.8} & 39.4 & 38.4 & \textbf{43.7} & \textbf{46.3} \\ 
\midrule
\quad w/o \textsc{PromptPlanner}  & \underline{25.2} & \underline{25.9} & \underline{30.9} & \underline{32.9} & \underline{35.1} & \underline{34.0} & \underline{36.1} & \underline{37.3} & \underline{37.1} & \underline{38.8}\\
\quad w/o \textsc{PairwiseSelect} & 24.5 & 33.3 & 29.2 & 35.4 & 33.3 & 27.9 & 35.5 & \textbf{44.2} & 38.5 & 33.8 \\
\quad w/ only Adversarial Judge & \underline{35.0} & \underline{40.0} & \textbf{\underline{42.0}} & \textbf{\underline{44.0}} & \underline{42.0} & \underline{35.3} & \underline{18.8} & \underline{18.8} & \underline{18.8} & \underline{26.7}\\
\quad w/ only Normal Judge    &  \underline{35.0} & \underline{32.0} & \underline{29.3} & \underline{21.8} & \underline{17.2} & \underline{35.3} & \underline{31.9} & {\underline{37.8}} & {\underline{36.6}} & \underline{33.3}\\
\quad w/o DTPA &  \underline{35.0} & \underline{36.0} & \underline{35.6} & \underline{36.1} & \underline{37.8} & \underline{35.3} & \textbf{\underline{40.4}} & \underline{40.0} & \underline{39.5} & \underline{45.2} \\
\bottomrule
\end{tabular}}
\caption{Ablation \underline{results} evaluated on half of the benchmarks. Each module in \model{} contributes uniquely: \textsc{PromptPlanner} enhances initialization, \textsc{PairwiseSelect} stabilizes iterative improvements, combining both Judges balances critiques' depth and usefulness, and \textsc{DTPA} enables effective prompt refinement.}
\label{tab:ablation-macro-results}
\end{table}

\paragraph{Ablation Studies.}\label{subsec:ablation-studies}

We conduct ablation studies to analyze \model{}'s components by evaluating: (i) without \textsc{PromptPlanner} (Step 1), sampling initial prompts without structured planning; (ii) without \textsc{PairwiseSelect} (Step 2), replacing selection with a simple bidirectional comparison as in scaled baselines; (iii) using only the Adversarial Judge (Step 3) i.e., the negative critiques; (iv) using only the Normal Judge; and (v) without the Deep Thinking Prompting Agent (DTPA, Step 4), revising prompts directly from feedback without reasoning-based introspection. As shown in \Cref{tab:ablation-macro-results}, removing any component leads to suboptimal performance, with each component playing a distinct role. Specifically, removing Step 1 weakens initialization across both datasets (Init: 25.2\% in single-scene vs. 35.5\%; 34.0\% on multi-scene vs. 37.8\%). Without Step 2, results are unstable: although the performance occasionally matches or exceeds other variants in specific rounds, performance drops significantly in later iterations. In addition, using only the Adversarial Judge leads to strong single-scene gains but fails to generalize, especially on multi-scene scenarios, where win rates stagnate (18.8\% across several iterations). In contrast, using only the Normal Judge collapses on multi-scene, where performance drops to just 17.2\% by iteration 5. This divergence confirms the necessity of combining both judge types. Lastly, removing DTPA results in relatively smooth improvements but with lower ceilings, showing that high-quality, reasoning-driven prompt revisions are crucial for maximizing performance, especially in complex multi-scene generations. 


\paragraph{Can \model{} Work with More Optimization Iterations?}
\label{subsec:scaling-more-iterations}

\begin{figure}[t!]
\centering
\begin{subfigure}[t]{0.48\textwidth}
\includegraphics[width=\textwidth]{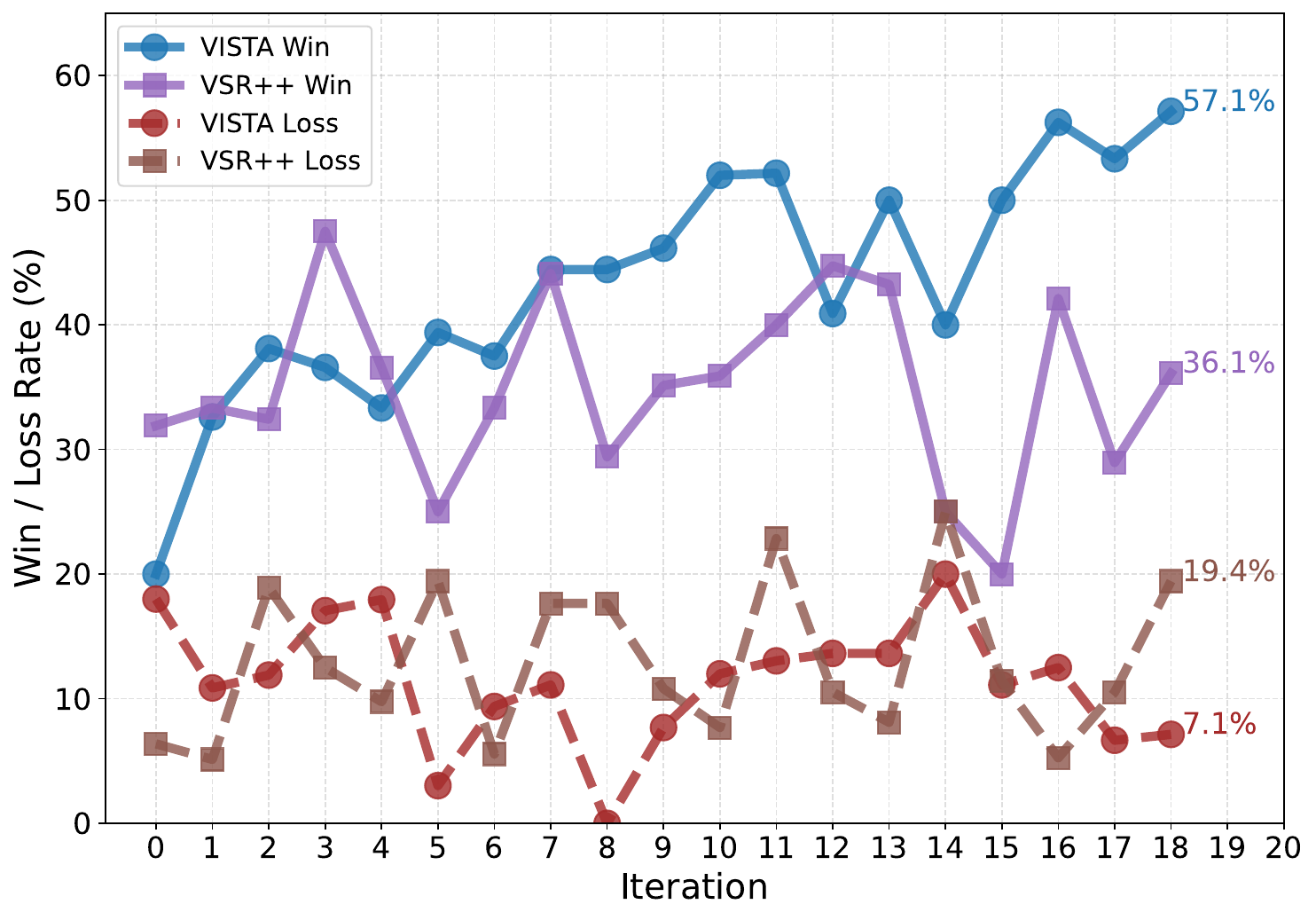}
\end{subfigure}
\hfill
\begin{subfigure}[t]{0.48\textwidth}
\includegraphics[width=\textwidth]{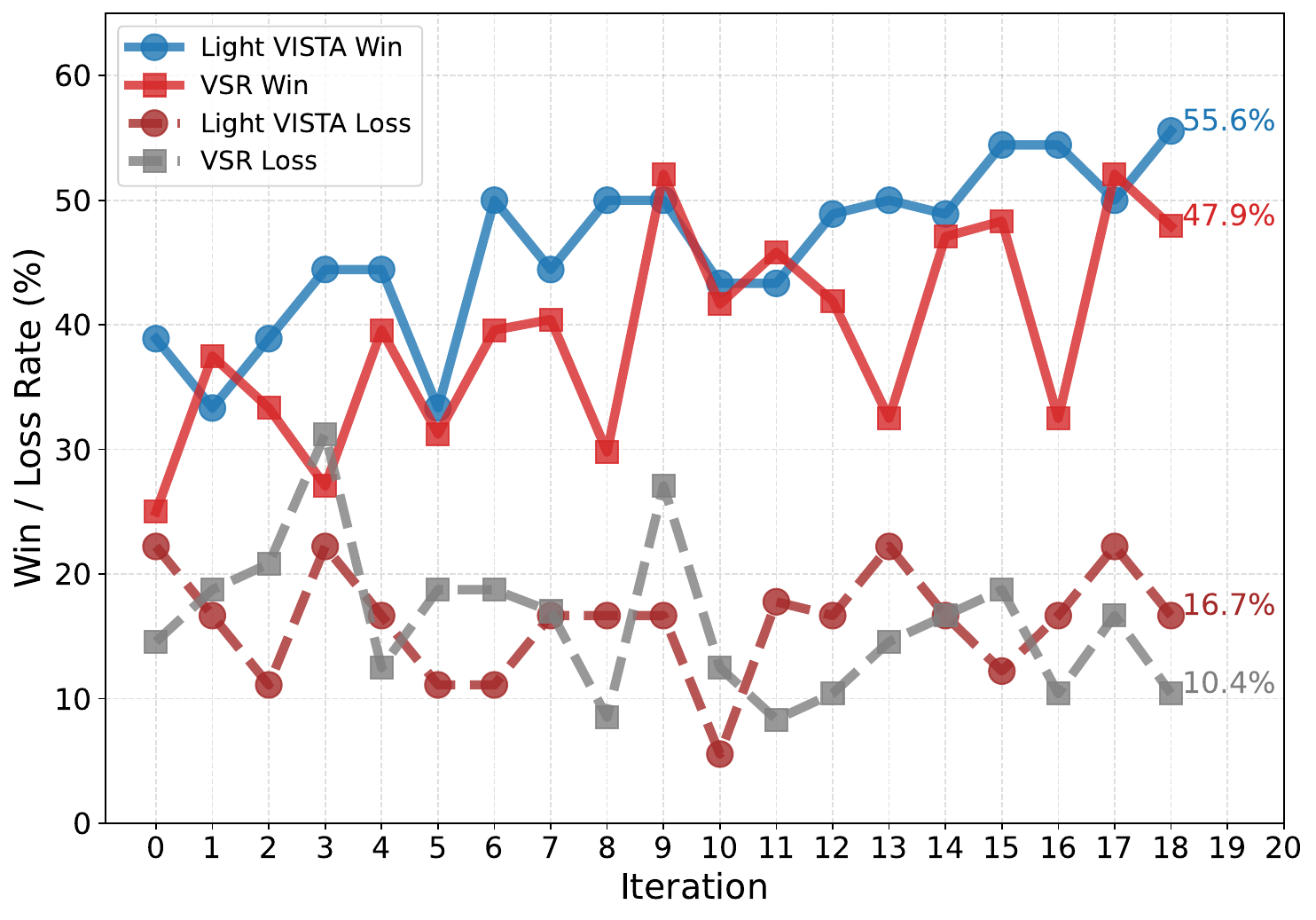}
\end{subfigure}
\caption{Effect of scaling the \#iterations on performance. \textbf{Left:} Single-scene. \textbf{Right:} Multi-scene.}
\label{fig:scaling-iterations}
\end{figure}

We scale up the number of iterations in \model{} and compare with the best-performing baselines, as shown in \Cref{fig:scaling-iterations}. For single-scene scenarios, we compare against VSR++, running both methods for up to 20 iterations with 8 sampled videos per iteration. For multi-scene scenarios, we compare against VSR by running a lightweight version of \model{}, termed \textbf{Light \model{}}, which samples only 1 video per iteration and omits its Step 2. We observe the baselines exhibit noisy and inconsistent improvement, whereas \model{} shows a more stable and consistent upward trajectory. Notably, on MovieGenVideo, VSR++ gains little with more iterations, whereas \model{} continues to improve. These results suggest that our method can be potentially scalable with increased test-time computation.

\paragraph{Can \model{} Work with Weaker Models?} 

\setlength{\tabcolsep}{7pt} 
\begin{table}[h!]
\centering
\resizebox{.9\textwidth}{!}{%
\begin{tabular}{l|ccccc|ccccc}
\toprule
 & \multicolumn{5}{c|}{Single-scene} & \multicolumn{5}{c}{Multi-scene} \\
\midrule
\multicolumn{1}{l|}{Win Rates over DP} & Init & 2 & 3 & 4 & 5 & Init & 2 & 3 & 4 & 5 \\
\midrule
\quad Veo 2 performance w/ \model{} &  15.0 & 15.6 & 18.7 & 20.5 & \textbf{23.8} &  27.6 & 21.4 & 21.4 & 30.8 & \textbf{33.3} \\ 
\bottomrule
\end{tabular}}
\caption{Veo 2 performance with \model{} on both datasets.}
\label{tab:veo2-results}
\end{table}

We conduct two experiments using Veo 2 \citep{veo2googledeepmind}, a less powerful model than Veo 3. Veo 2 is selected because of its strong instruction-following capabilities. As shown in \Cref{tab:veo2-results}, \model{} improves Veo 2 consistently across datasets, shifting the win rates over Direct Prompting up to 23.8\% and 33.3\% respectively. In addition, the performance gains by Veo 2 are less than Veo 3. This can be attributed to the fact that Veo 2 being less capable to fully leverage the details optimized by \model{}. Overall, these results verify that \model{} effectively enhances generation quality even when paired with less capable video models.

\paragraph{Customizing \model{}.} 
\model{} provides flexibility for users to define or adjust its behaviors. Both the selection metrics $\mathcal{M}^S_{user}$ (Step 2), the critique metrics $\mathcal{M}^C_{user}$ (Step 3), and all steps' constraints can be customized. For example, the constraints in Step 1 can be removed to encourage model being more creative, leading to more creative videos generated from user prompt; $\mathcal{M}^S_{user}$ can prioritize aspects that reflect subjective preferences, such as \emph{color grading fidelity} or \emph{emotional impact}, emphasizing the users' vision or affective goals. In addition, $\mathcal{M}^C_{user}$ can focus on more fine-grained behaviors like the subtle fluidity of character gestures in the visual dimension. Further exploration of user-customized metrics and constraints, particularly those capturing unique aesthetic or narrative nuances, is an exciting direction to bring video generation closer to truly personalized creative video generation.

\section{Conclusions}
We introduce \model{}, a novel multi-agent system that enhances text-to-video generation at test time by jointly optimizing visual, audio, and contextual elements through explicit prompt planning, multi-agent critiques, and alignment-based video selection. \model{} notably improves video quality in state-of-the-art models like Veo 3 while preserving the original prompt intents, enhancing their instruction-following, and reducing physical, visual, and audio hallucinations, resulting in significant human preference gains. Our framework is potentially scalable, marking a step toward more adaptive, human-aligned, and collaborative video generation.

\section*{Limitations}
Despite the notable performance gains achieved by \model{}, several limitations remain, revealing promising avenues for future work. Firstly, our evaluation relies primarily on MLLMs and automated metrics, which may introduce systematic biases or fail to capture aspects of video quality that humans prioritize. While we include human evaluation and cross-validate with multiple automated approaches, the comprehensive human evaluation remains prohibitively expensive that affect the entire field. Secondly, while our critique mechanism provides a configurable interface, the current default metrics reflect certain assumptions about video quality that may not generalize across different cultural contexts, creative styles, or user preferences. Customizing \model{}'s metrics to better reflect user-specific or domain-specific preferences could enhance its adaptability and robustness. Lastly, \model{} requires both MLLMs and T2V models with strong instruction-following and reasoning capabilities to function effectively. As such models continue to improve, we expect this limitation to diminish.

\section*{Acknowledgments} 
We thank Jiefeng Chen, Ozgur Kara, Long Le, George Lee, Palash Goyal and Jinsung Yoon for their valuable discussions and feedback, as well as our colleagues at Google for their helpful insights. We also want to thank Min-Yen Kan, Nancy F. Chen, Shafiq Joty, and Kenji Kawaguchi from NUS for supporting Do Xuan Long during the internship.

\bibliographystyle{abbrvnat}
\bibliography{main}

\newpage
\appendix
\section{Additional Results}

\subsection{Evaluations on Conventional Video and Audio Generation Metrics}\label{subsec:traditional-metrics-results}

\begin{table}[h!]
\centering
\footnotesize
\resizebox{1\textwidth}{!}{%
\begin{tabular}{lcccccccc|ccc|cc}
\toprule
\textbf{Method} & \textbf{Subject} & \textbf{Background} & \textbf{Temporal} & \textbf{Motion} & \textbf{Dynamic} & \textbf{Aesthetic} & \textbf{Imaging} & \textbf{Temporal} & \textbf{Audio} & \textbf{Audio} & \textbf{Audio} & \textbf{CLIP} & \textbf{Inception} \\
& \textbf{Consistency} & \textbf{Consistency} & \textbf{Flickering} & \textbf{Smoothness} & \textbf{Degree} & \textbf{Quality} & \textbf{Quality} & \textbf{Style} & \textbf{Noisiness} & \textbf{Discontinuity} & \textbf{Coloration} & \textbf{Score} & \textbf{Score}\\
\midrule
DP       & 89.89 & \textbf{94.39} & 97.82 & \textbf{99.23} & 75.95 & 61.86 & 64.42 & 7.88 & 1.74 & 2.04 & 1.65 & 0.310 & 1.053\\
VSR      & 89.33 & 93.53 & 97.79 & 99.26 & 64.56 & 63.45 & 65.53 & 9.26 & 1.73 & 2.13 & \textbf{1.64} & 0.309  & 1.082\\
VSR++    & 87.96 & 93.53 & \textbf{97.88}  & 99.12 & 74.68 & 60.68 & 63.06 & 9.25 & 1.65 & 2.03 & 1.56 & 0.310 & 1.078\\
Rewrite  & 89.09 & 93.79 & 97.59 & 99.17 & 77.22 & 62.52 & 62.58 & 8.57 & 1.64 & 1.99 & 1.53 & 0.310 & 1.085\\
VPO      & 86.74 & 92.66 & 97.76 & 99.15 & 77.22 & 61.17 & 64.01 & 8.03 & 1.70 & 1.97 & 1.59 & 0.311 & 1.039\\
\midrule
\textbf{\model{}}  & \textbf{89.95} & 92.89 & 97.82 & 98.94 & \textbf{89.87} & \textbf{64.53} & \textbf{65.89} & \textbf{9.63} & \textbf{1.88} & \textbf{2.19} & 1.62 & \textbf{0.358} & \textbf{1.101}\\
\bottomrule
\end{tabular}}
\caption{\textbf{Single-scene:} Evaluation results using VBench's any-video evaluation metrics for visual quality, NISQA metrics for audio quality, and CLIP-Score for text-video alignment.}
\label{tab:meta-results-on-traditional-metrics}
\end{table}

\begin{table}[h!]
\centering
\footnotesize
\resizebox{1\textwidth}{!}{%
\begin{tabular}{lcccccccc|ccc|cc}
\toprule
\textbf{Method} & \textbf{Subject} & \textbf{Background} & \textbf{Temporal} & \textbf{Motion} & \textbf{Dynamic} & \textbf{Aesthetic} & \textbf{Imaging} & \textbf{Temporal} & \textbf{Audio} & \textbf{Audio} & \textbf{Audio} & \textbf{CLIP} & \textbf{Inception} \\
& \textbf{Consistency} & \textbf{Consistency} & \textbf{Flickering} & \textbf{Smoothness} & \textbf{Degree} & \textbf{Quality} & \textbf{Quality} & \textbf{Style} & \textbf{Noisiness} & \textbf{Discontinuity} & \textbf{Coloration} & \textbf{Score} & \textbf{Score} \\
\midrule
DP       & 79.28 & 85.27 & 97.99 & 99.14 & 72.15 & 47.39 & 67.19 & 6.54 & 2.24 & 2.62 & \textbf{2.28} & 0.285 & 1.11 \\
VSR      & 76.41 & 83.50 & \textbf{98.30} & 99.15 & 53.16 & 47.74 & 65.86 & 9.53 & 2.24 & 2.64 & 2.27 & 0.286 & 1.09 \\
VSR++    & 78.52 & 85.43 & 98.33 & 99.11 & 60.75 & 47.96 & 65.68 & 6.80 & 2.14 & 2.53 & 2.13 & 0.288 & 1.08 \\
Rewrite  & 77.85 & \textbf{86.08} & 98.02 & 99.08 & 67.09 & \textbf{50.56} & 65.62 & 8.73 & 1.93 & 2.38 & 2.05 & 0.290 & 1.11 \\
VPO      & 77.08 & 83.80 & 98.15 & 99.13 & 58.23 & 49.08 & 67.04 & 8.32 & 2.25 & 2.67 & 2.26 & 0.288 & 1.09 \\
\midrule
\textbf{\model{}} & \textbf{79.45} & 83.95 & 98.20 & \textbf{99.16} & \textbf{75.18} & {50.00} & \textbf{68.87} & \textbf{10.09} & \textbf{2.30} & \textbf{2.69} & 2.21 & \textbf{0.299} & \textbf{1.15} \\
\bottomrule
\end{tabular}}
\caption{\textbf{Multi-scene:} Evaluation results using VBench's any-video evaluation metrics for visual quality, NISQA metrics for audio quality, and CLIP-Score for text-video alignment.}
\label{tab:ourdata-results-on-traditional-metrics}
\end{table}

\subsection{Human Evaluation Results Details}

\vspace{-3mm}
\begin{table}[h!]
  \centering
  \resizebox{\textwidth}{!}{%
    \begin{tabular}{ccc}
      \begin{tabular}{lcc}
        \toprule
        \textbf{Ann.} & \model{} & VSR(++) \\
        \midrule
        Ann. 4 & \textbf{3.06} & 2.76 \\
        Ann. 5 & \textbf{4.44} & 4.02 \\
        Ann. 6 & \textbf{3.84} & 3.22 \\
        \bottomrule
      \end{tabular}
      &
      \resizebox{0.45\linewidth}{!}{%
      \begin{tabular}{lcccc}
        \toprule
        \textbf{Ann.} & \multicolumn{2}{c}{\textbf{Single-scene}} & \multicolumn{2}{c}{\textbf{Multi-scene}} \\
        & \model{} & VSR++ & \model{} & VSR \\
        \midrule
        Ann. 1 & \textbf{76\%} & 24\% & \textbf{72\%} & 28\% \\
        Ann. 2 & \textbf{68\%} & 32\% & \textbf{68\%} & 32\% \\
        Ann. 3 & \textbf{64\%} & 36\% & \textbf{60\%} & 40\% \\
        Ann. 4 & \textbf{62\%} & 38\% & \textbf{66\%} & 34\% \\
        Ann. 5 & \textbf{66\%} & 34\% & \textbf{62\%} & 38\% \\
        \bottomrule
      \end{tabular}}
      &
      \begin{tabular}{lc|c}
        \toprule
        \textbf{Ann.} & \multicolumn{1}{c}{Visual} & \multicolumn{1}{c}{Audio} \\
        \midrule
        Ann. 1 & \textbf{3.90} / 3.34 & \textbf{4.08} / 3.22 \\
        Ann. 2 & \textbf{3.72} / 3.50 & \textbf{3.64} / 3.14 \\
        Ann. 3 & \textbf{3.69} / 3.33 & \textbf{3.68} / 3.28 \\
        \bottomrule
      \end{tabular}
      \\
      \small (a) Self-improvement scores (over 5). & 
      \small (b) Win rate of \model{} vs. best baselines. & 
      \small (c) Visual and audio scores (over 5).
    \end{tabular}
  }
  \caption{\small{Human evaluation results.}}
  \label{tab:human-eval-results}
\end{table}

\Cref{tab:human-eval-results} presents our human evaluation results across annotators.

\begin{table}[h!]
\centering
\begin{tabularx}{\textwidth}{@{}X@{}}
    \begin{tabular}{@{}c@{}c@{}c@{}c@{}}
        \includegraphics[width=0.25\linewidth]{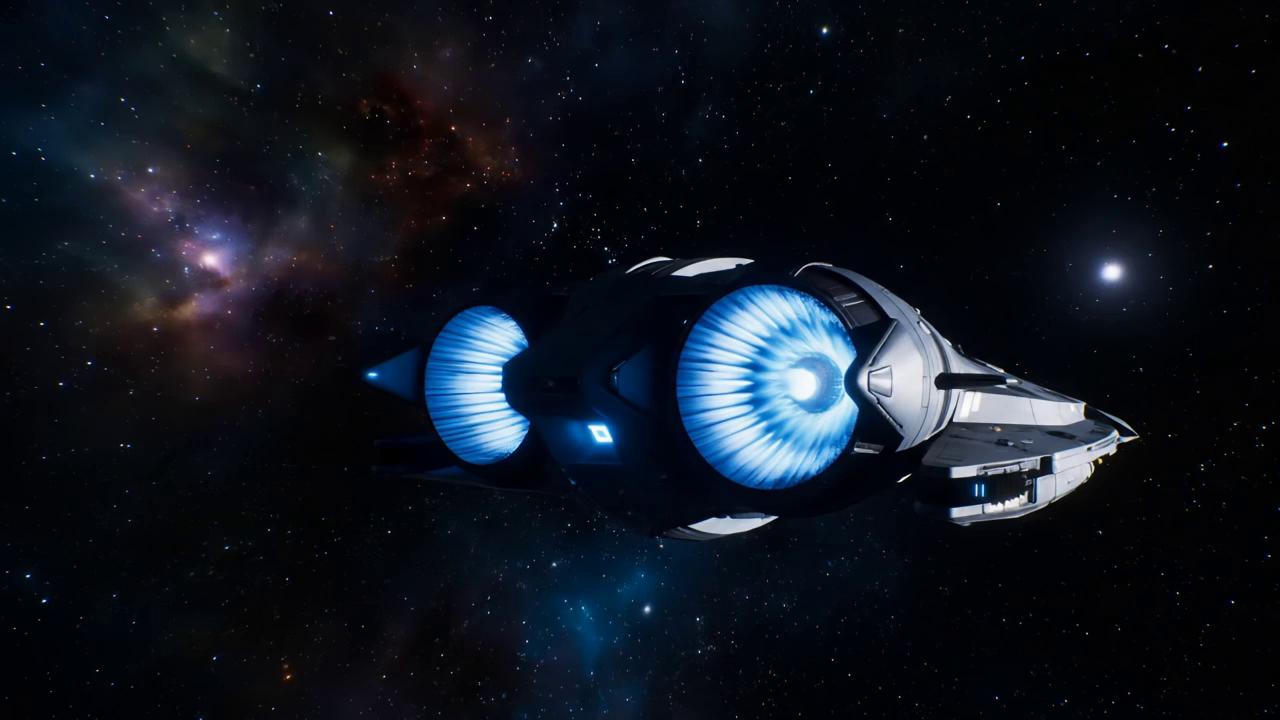} &
        \includegraphics[width=0.25\linewidth]{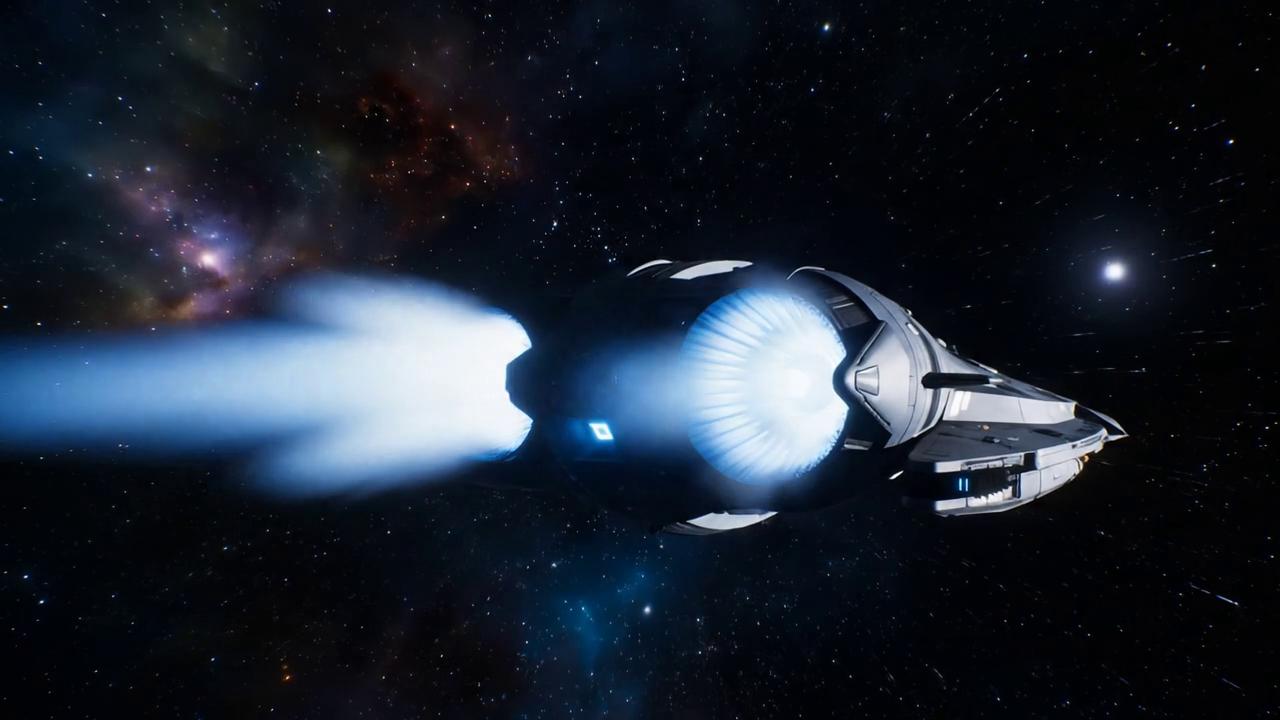} &
        \includegraphics[width=0.25\linewidth]{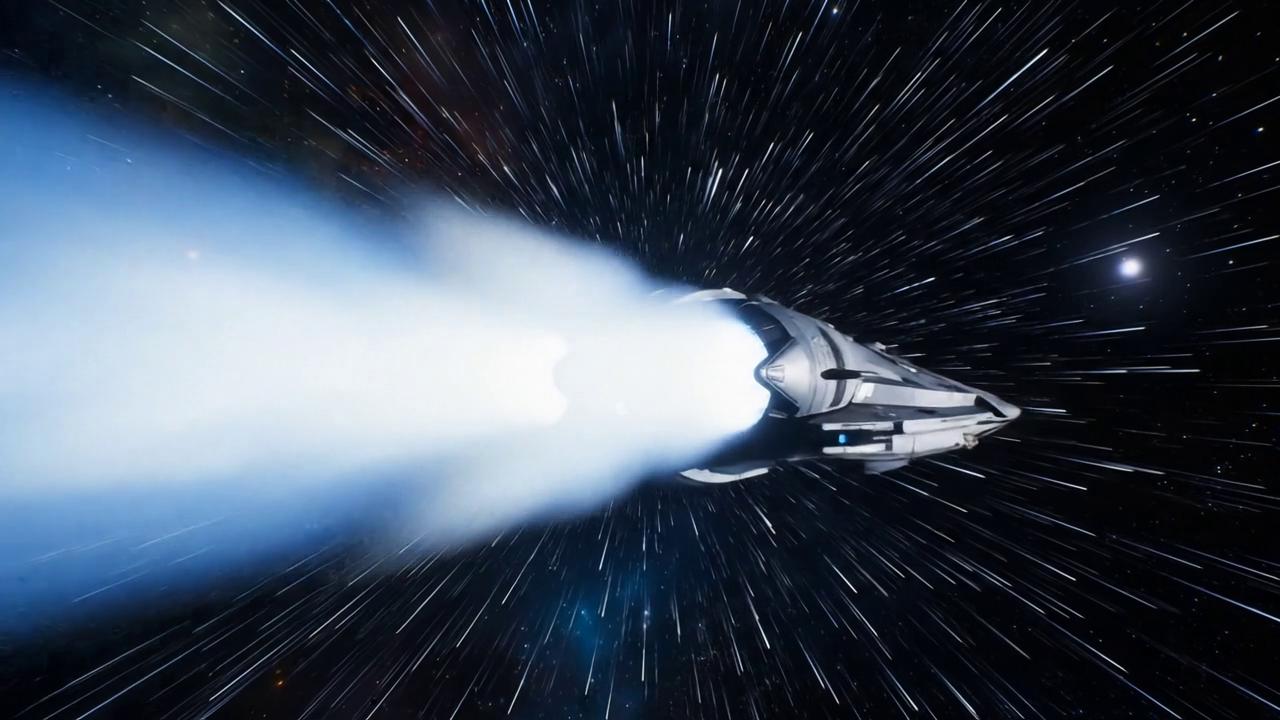} &
        \includegraphics[width=0.25\linewidth]{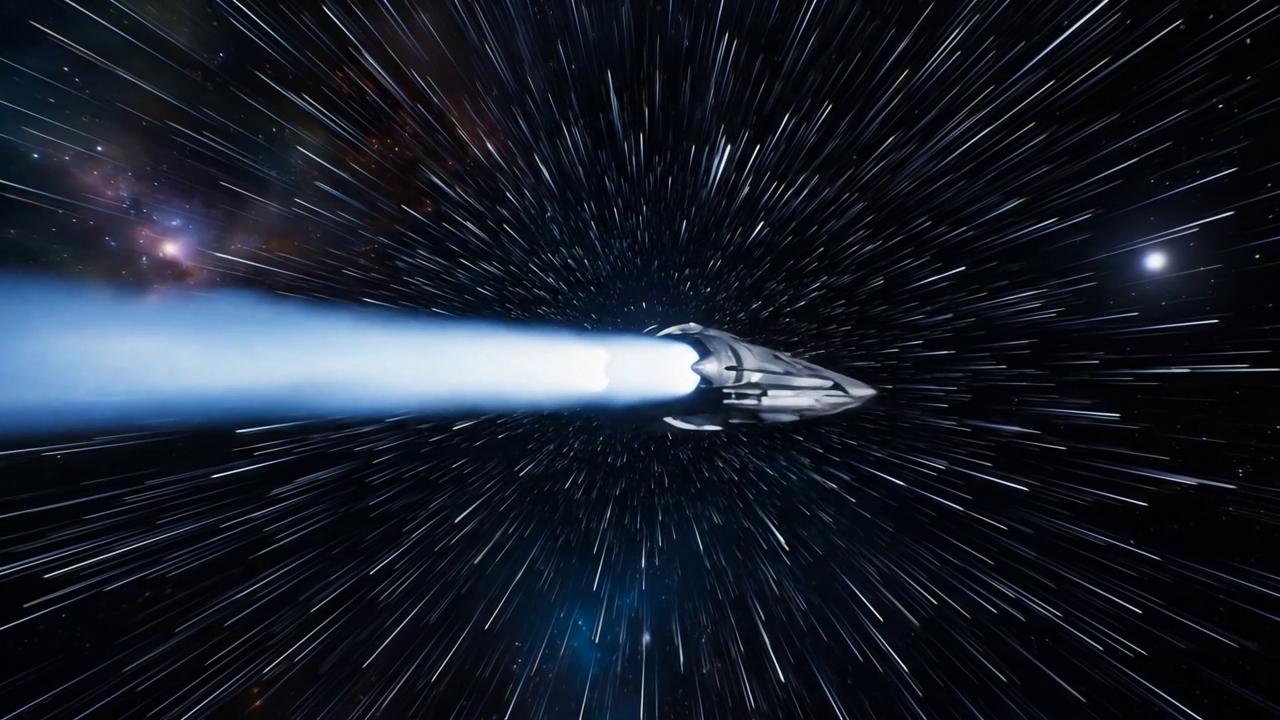} \\
    \end{tabular} \\

    \footnotesize
    \RaggedRight
    \scriptsize{\textbf{Prompt}: A spaceship entering hyperdrive, stars streaking past as it accelerates.}\\

    \scriptsize{\textbf{Self-Refine:} Overall, the generated video is {highly successful} in fulfilling the prompt\dots}
    \\  

    \footnotesize
    \RaggedRight
    \scriptsize{\textbf{\model{} (Motions and Dynamics):} While the Normal Judge praises the smoothness of the ship's motion and dynamic star streaking, the Negative Judge correctly identifies a major directional flaw: \textcolor{red}{the spaceship moves vertically, which conflicts with viewer expectations of horizontal acceleration}. Additionally, the Negative Judge points out \textcolor{red}{the lack of micro-dynamics (e.g., rotational drift, buildup phases) and unrealistic exhaust behavior, which diminish the believability of motion}\dots}
    
    \\[0.5mm]
    
    \begin{tabular}{@{}c@{}c@{}c@{}c@{}} 
    \includegraphics[width=0.25\linewidth]{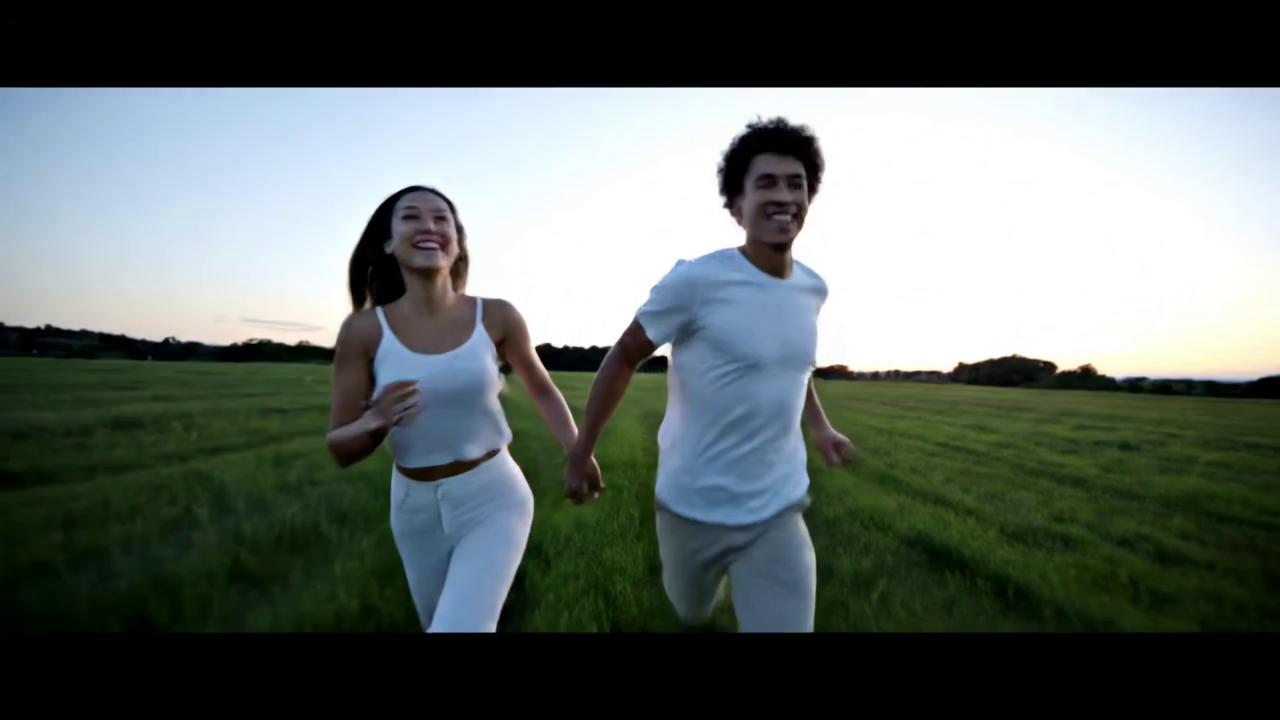} &
    \includegraphics[width=0.25\linewidth]{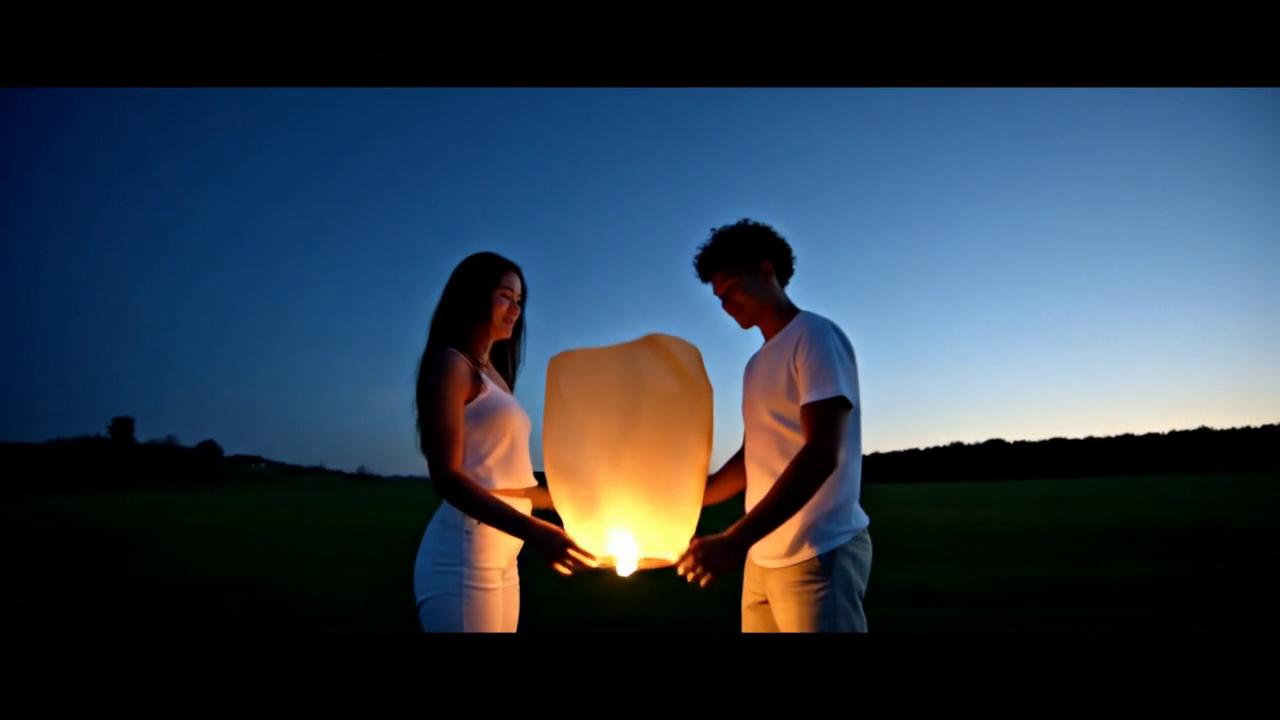} &
    \includegraphics[width=0.25\linewidth]{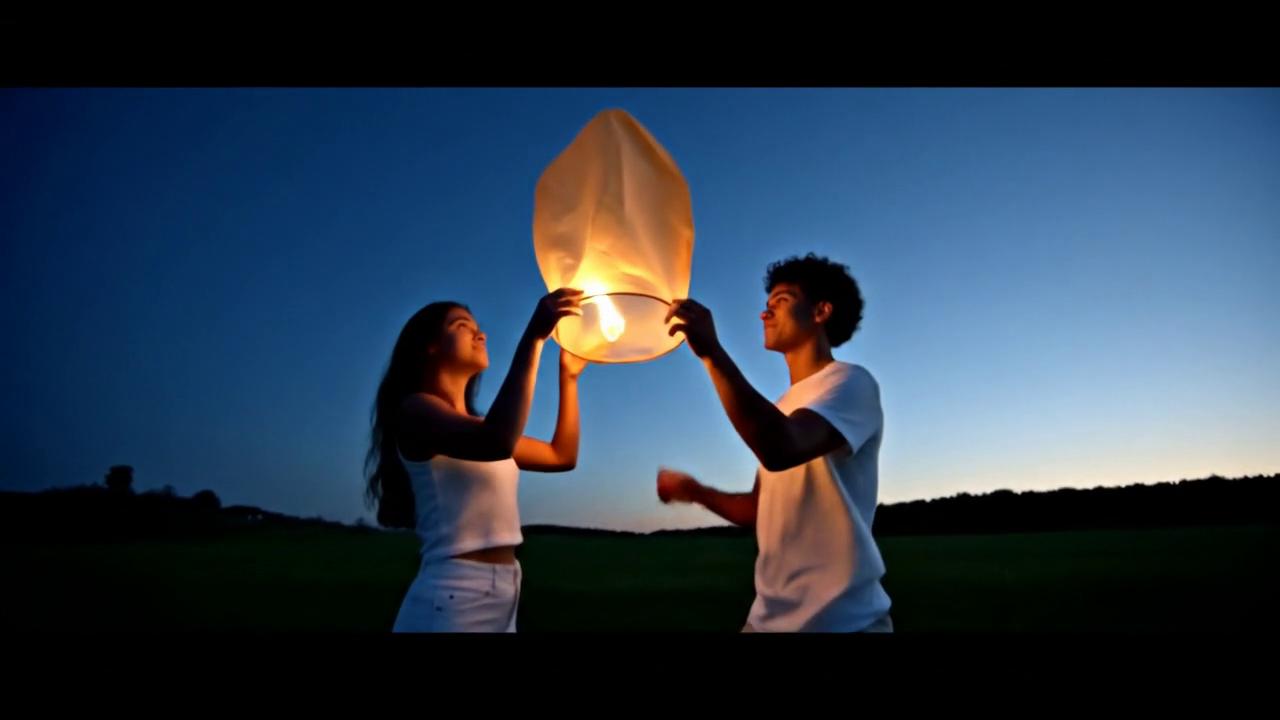} &
    \includegraphics[width=0.25\linewidth]{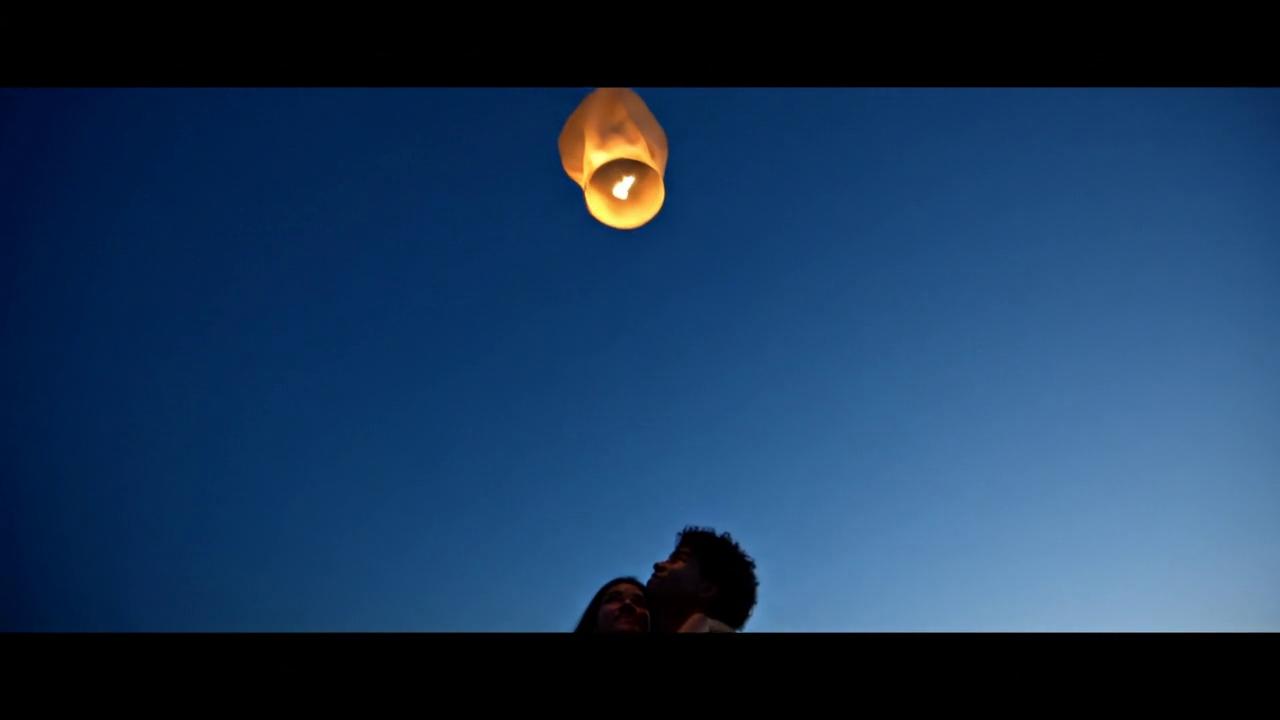} \\
    \end{tabular} \\

    \footnotesize
    \RaggedRight
    \scriptsize{\textbf{Prompt}: The couple runs hand in hand to release a sky lantern, then watches it drift upward into the night sky, carried by the wind with the stars shining above.}\\
    \scriptsize{\textbf{Self-Refine}: The video successfully portrays the actions of the couple running and releasing a lantern\dots}\\ \scriptsize{\textbf{\model{}(Contextual Suitability)}: The scene struggles significantly with internal logical consistency\dots an \textcolor{red}{abrupt and jarring shift from a bright, late-afternoon setting to a deep, artificial blue night sky} without any visual cues for time passing\dots \textbf{(Visual Characters)}: While the two characters\dots their portrayal \textcolor{red}{lacks emotional depth and genuine interaction}. Their expressions, particularly during the lantern release, appear somewhat static and posed, missing an opportunity to convey a deeper sense of wonder\dots}
\end{tabularx}
\caption{\model{} delivers high-level critiques on visual fidelity, contextual consistency, and emotional expression, surfacing nuanced flaws that conventional LLM critiques often miss but are noticeable to human judges.}
\label{tab:good-critique-examples}
\end{table}

\subsection{Why Does \model{} Work? Case Studies} \label{subsec:why-does-vista-work?}

\paragraph{\model{} Can Provide Human-Like Critiques (Step 3).} 

\model{} provides intelligent, multi-faceted critiques that is often overlooked by conventional MLLM critiques, yet is intuitively recognized by humans. As shown in \Cref{tab:good-critique-examples}, \model{} can identify subtle visual flaws, such as a spaceship moving vertically instead of horizontally and movement against a static background. It also offers high-level reasoning context-aware critiques, such as unnatural characters' expressions. These illustrate that \model{} can deliver much more rigorous, high-level reasoning critiques than prior studies.

\begin{table}[h!]
\centering
\begin{tabularx}{\textwidth}{@{}X@{}}

    \begin{tabular}{@{}c@{}c@{}c@{}c@{}}
        \includegraphics[width=0.24\linewidth]{images/goodmodification_248_ours_348226125581640722/base_248_ours_348226125581640722_1.jpg} &
        \includegraphics[width=0.24\linewidth]{images/goodmodification_248_ours_348226125581640722/base_248_ours_348226125581640722_76.jpg} &
        \includegraphics[width=0.24\linewidth]{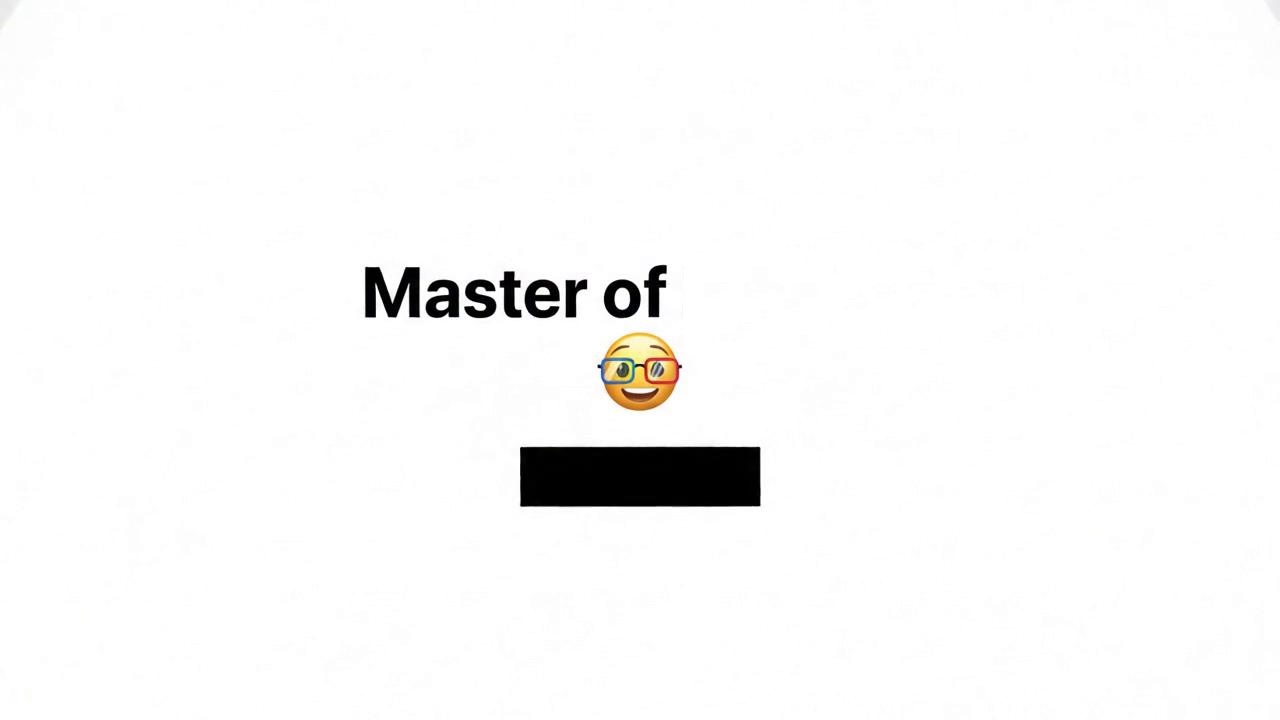} &

        \includegraphics[width=0.24\linewidth]{images/goodmodification_248_ours_348226125581640722/base_248_ours_348226125581640722_159.jpg} \\
    \end{tabular}\vspace{-1mm}
    \begin{tabular}{@{}c@{}c@{}c@{}c@{}}
        \includegraphics[width=0.24\linewidth]{images/goodmodification_248_17871626666877391088/goodmodification_248_17871626666877391088_24.jpg} &
        \includegraphics[width=0.24\linewidth]{images/goodmodification_248_17871626666877391088/goodmodification_248_17871626666877391088_95.jpg} &
        \includegraphics[width=0.24\linewidth]{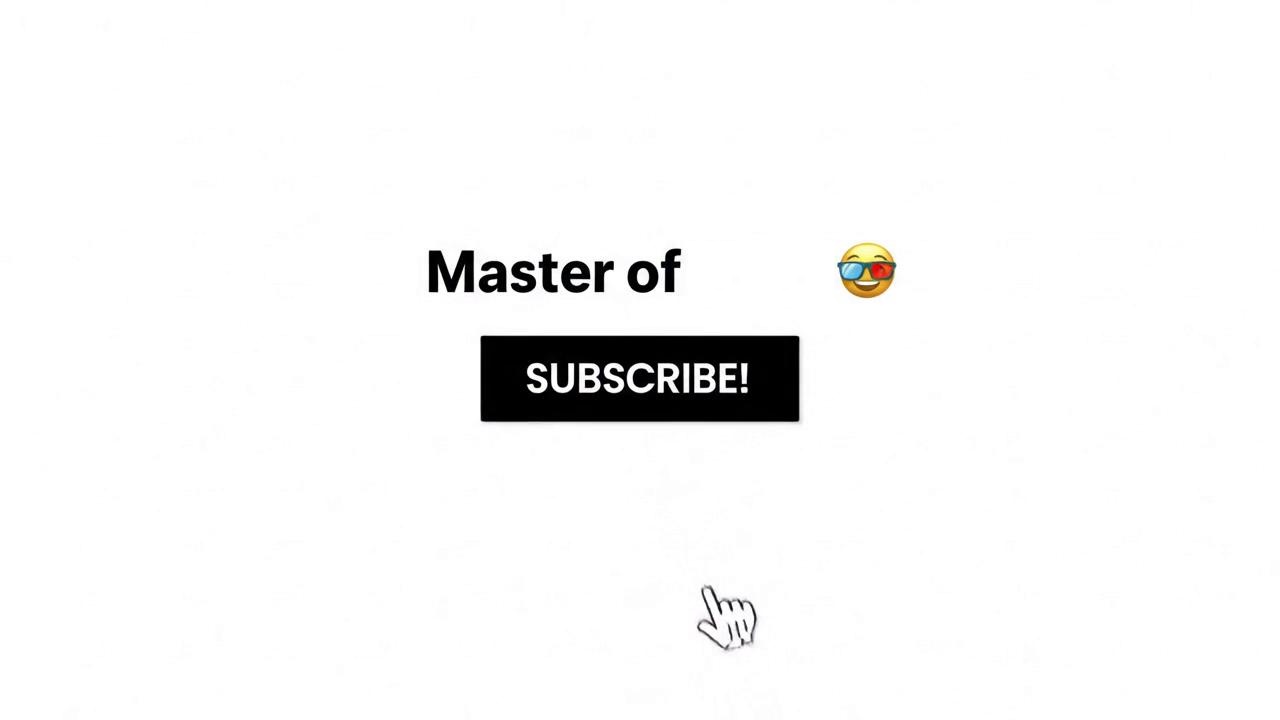} &
        \includegraphics[width=0.24\linewidth]{images/goodmodification_248_17871626666877391088/goodmodification_248_17871626666877391088_196.jpg} \\
    \end{tabular} \\

    \footnotesize
    \RaggedRight
    \scriptsize{\textbf{Prompt:} The video features a man outdoors, asking a trivia question about a comedian known for deadpan delivery, and then immediately providing the answer\dots [\{'timestamp': '0-5.5', 'scene\_type': 'Man asking and answering a trivia question outdoors.'\dots\},
    \{'timestamp': '5.5-8', 'scene\_type': 'Outro screen with branding and call to action.'\dots\}]\dots}
    \scriptsize{\textbf{\model{}'s Suggested Modifications:}} 
    \begin{itemize}\scriptsize
        \setlength\itemsep{0.5em}
        \item Update the scene's text overlays\dots \textcolor{blue}{text overlay should smoothly fade in/slide up} from the bottom, be legible\dots
        \vspace{-1mm}
        \item Refine the 'sounds'\dots with \textcolor{blue}{dialogue free of noticeable wind noise}. A subtle, consistent ambient street soundscape\dots
        \vspace{-1mm}
        \item \textcolor{blue}{Add a specific instruction for the transition between the first scene (timestamp '0-5.5') and the second scene (timestamp '5.5-8')} \dots
    \end{itemize}
\end{tabularx}
\vspace{-5mm}
\caption{\model{}'s suggested modifications. {Top:} Original video by DP showing abrupt scene transitions, distracting audio, and less polished text overlays. {Bottom:} \model{} refines the prompt leading to improved transitions, audio, overlay placement, and a nice click by the end. {See \Cref{appdx:deep-thinking-prompting-example} for full texts.}}
\label{fig:good-modifications}
\end{table}

\paragraph{\model{} Can Refine Prompts Targetedly (Step 4).} 

\model{} can reason to refine prompts to address nuanced issues across dimensions, as illustrated in \Cref{fig:good-modifications}. While these improvements may appear subtle, they meaningfully enhance the viewer experience and go beyond surface-level adjustments.

\paragraph{\model{} Can Filter Out Visually Engaging but Physically- or Audio-Nonsensical Videos (Step 2).} 
Through human investigations, we observe that \model{} can effectively filter out common failure cases in AI-generated videos, including incomplete coverage of the user prompt, unfinished activities, unnatural movements with nonsensical directions or speeds, and objects appearing or disappearing unexpectedly. Other frequent issues include low visual quality, noisy or distorted audio, artificially hallucinated objects and entities, and unexpectedly rendering text or voice overlays within the video. We invite audiences to visit our project page for examples.

\subsection{Prompt Length Distribution among Methods} \label{subsec:prompt-length-analyses}

\begin{figure}[h!]
\centering
\begin{subfigure}[t]{0.48\textwidth}
\includegraphics[width=\textwidth]{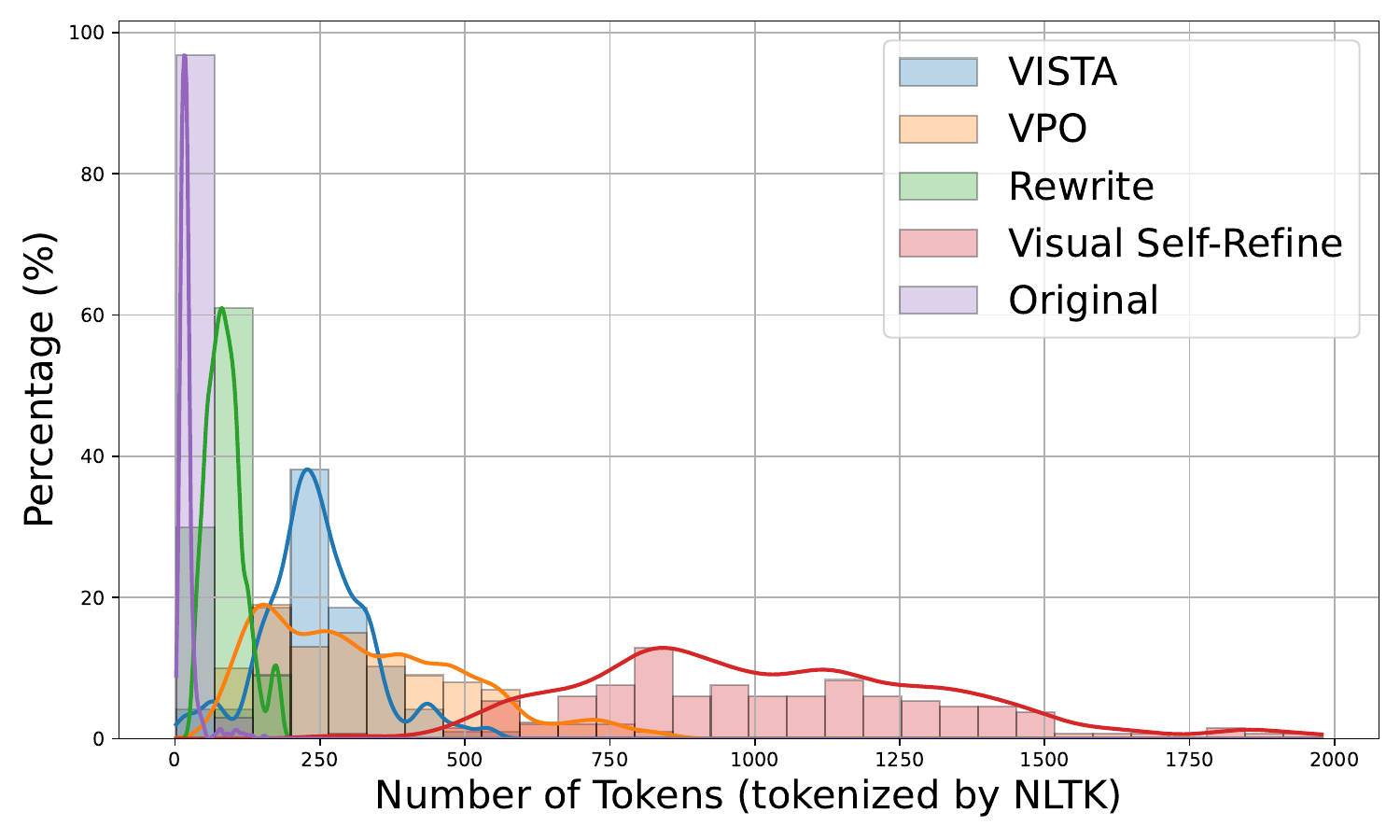}
\end{subfigure}
\hfill
\begin{subfigure}[t]{0.48\textwidth}
\includegraphics[width=\textwidth]{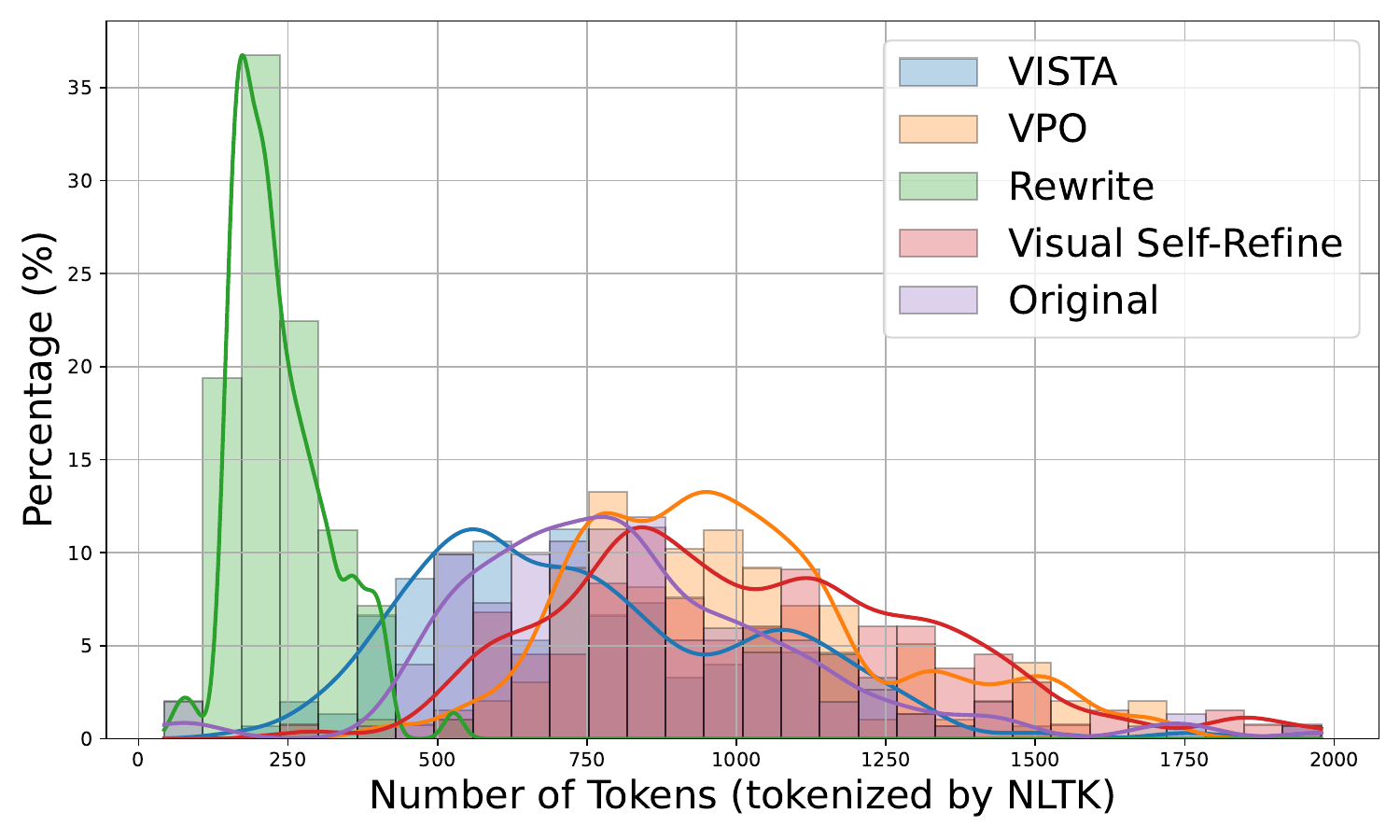}
\end{subfigure}
\caption{{Left:} Average \#tokens across iterations in single-scene scenarios; {Right:} In multi-scene scenarios.}
\label{fig:method-prompt-lengths}
\end{figure}

To further understand how different methods refine prompts, we plot the distributions of prompt lengths optimized across two benchmarks, alongside the lengths of the original prompts (denoted as ``Original'') in \Cref{fig:method-prompt-lengths}. In the single-scene scenarios, all methods tend to increase prompt lengths compared to the Original, with Visual Self-Refine producing the longest prompts over iterations. On our multi-scene dataset, Rewrite yields shorter prompts than the Original explainably because Rewrite follows the guidance from \citet{google_vertex_video_prompt_guide}, which recommends fewer properties than those used in our dataset's prompts. Meanwhile, our methods slightly shorten them, and both VPO and Visual Self-Refine slightly lengthen them.

\subsection{Benchmark-Based Results of \Cref{fig:dp_mavpo_finegrained}} \label{appdx:additonal-results-gemini}

\begin{figure}[h!]
    \centering
\includegraphics[width=1\textwidth]{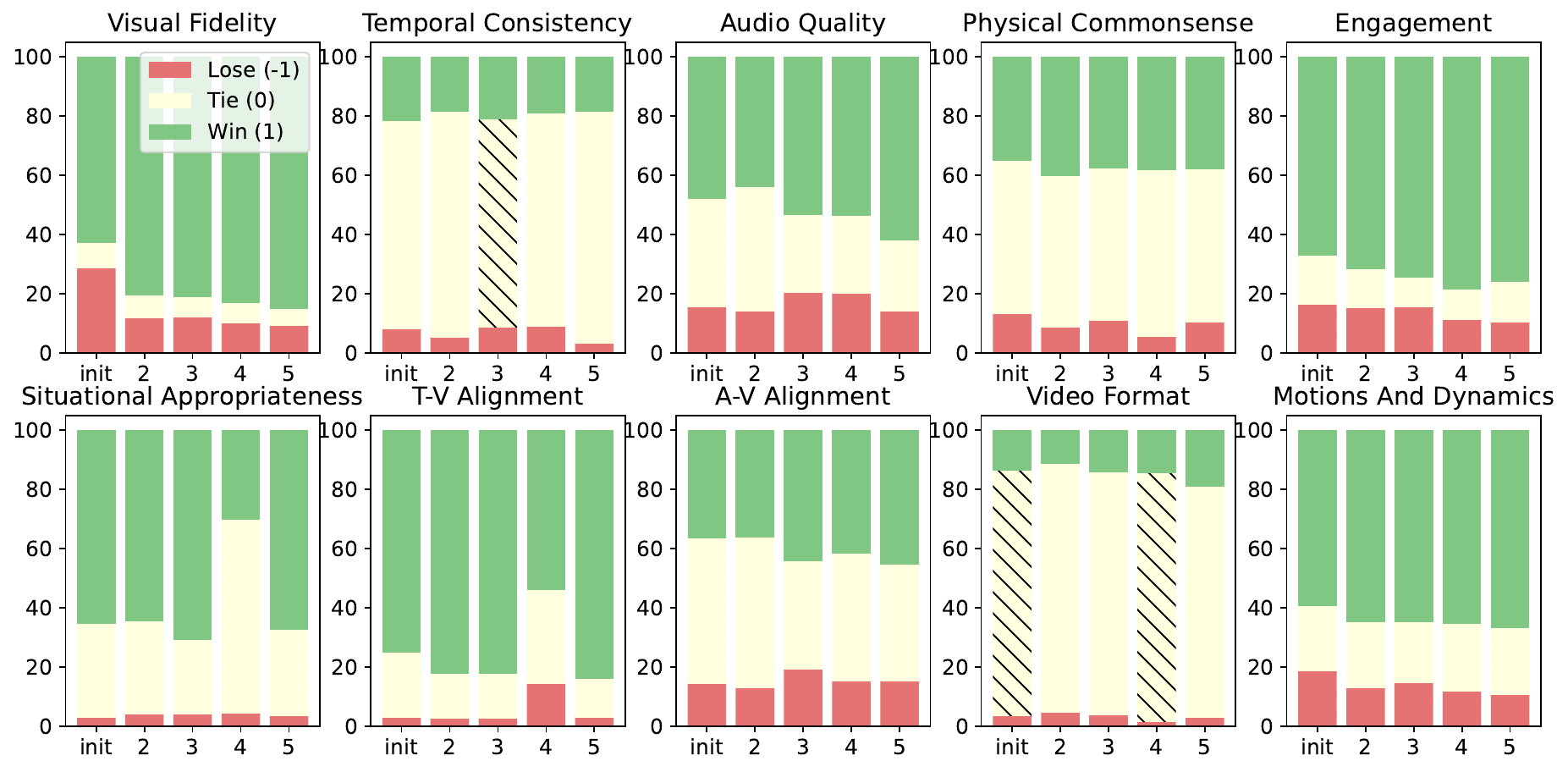}
    \caption{\textbf{Single-scene}: Average win/tie/lose comparison between \model{} and Direct Prompting (DP).}
\label{fig:dp_mavpo_meta_finegrained}
\end{figure}

\begin{figure}[h!]
    \centering
\includegraphics[width=1\textwidth]{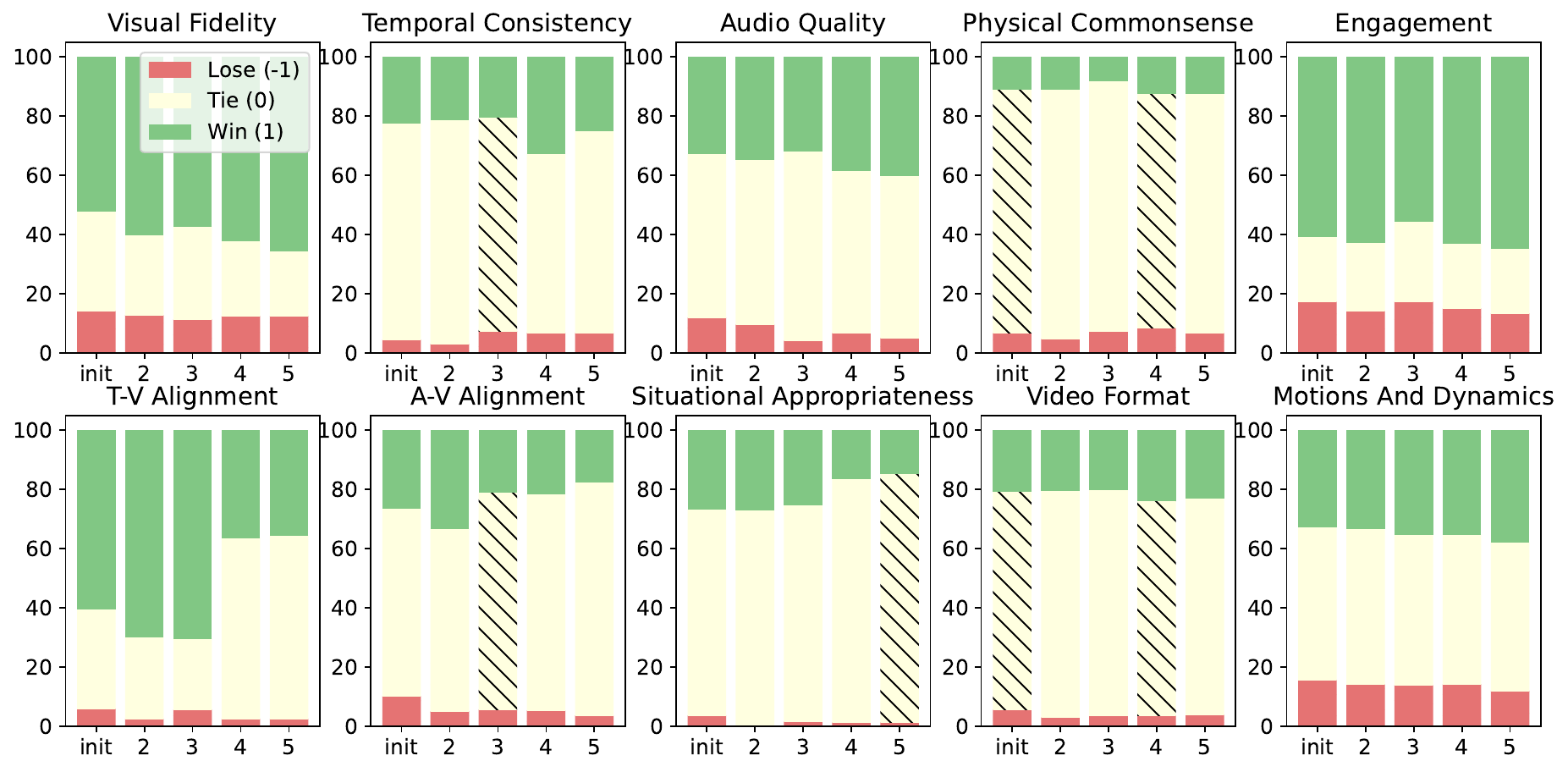}
    \caption{\textbf{Multi-scene}: Average win/tie/lose comparison between \model{} and Direct Prompting (DP).}
\label{fig:dp_mavpo_ourdata_finegrained}
\end{figure}

\Cref{fig:dp_mavpo_meta_finegrained} shows the average win/tie/lose comparison between \model{} and Direct Prompting (DP) in single-scene, while \Cref{fig:dp_mavpo_ourdata_finegrained} shows the same comparison in multi-scene scenarios.

\subsection{Results with Qwen2.5-VL-32B-Instruct as the Evaluator} \label{appdx:additonal-results-qwen}

\begin{figure}[h!]
    \centering
\includegraphics[width=1\textwidth]{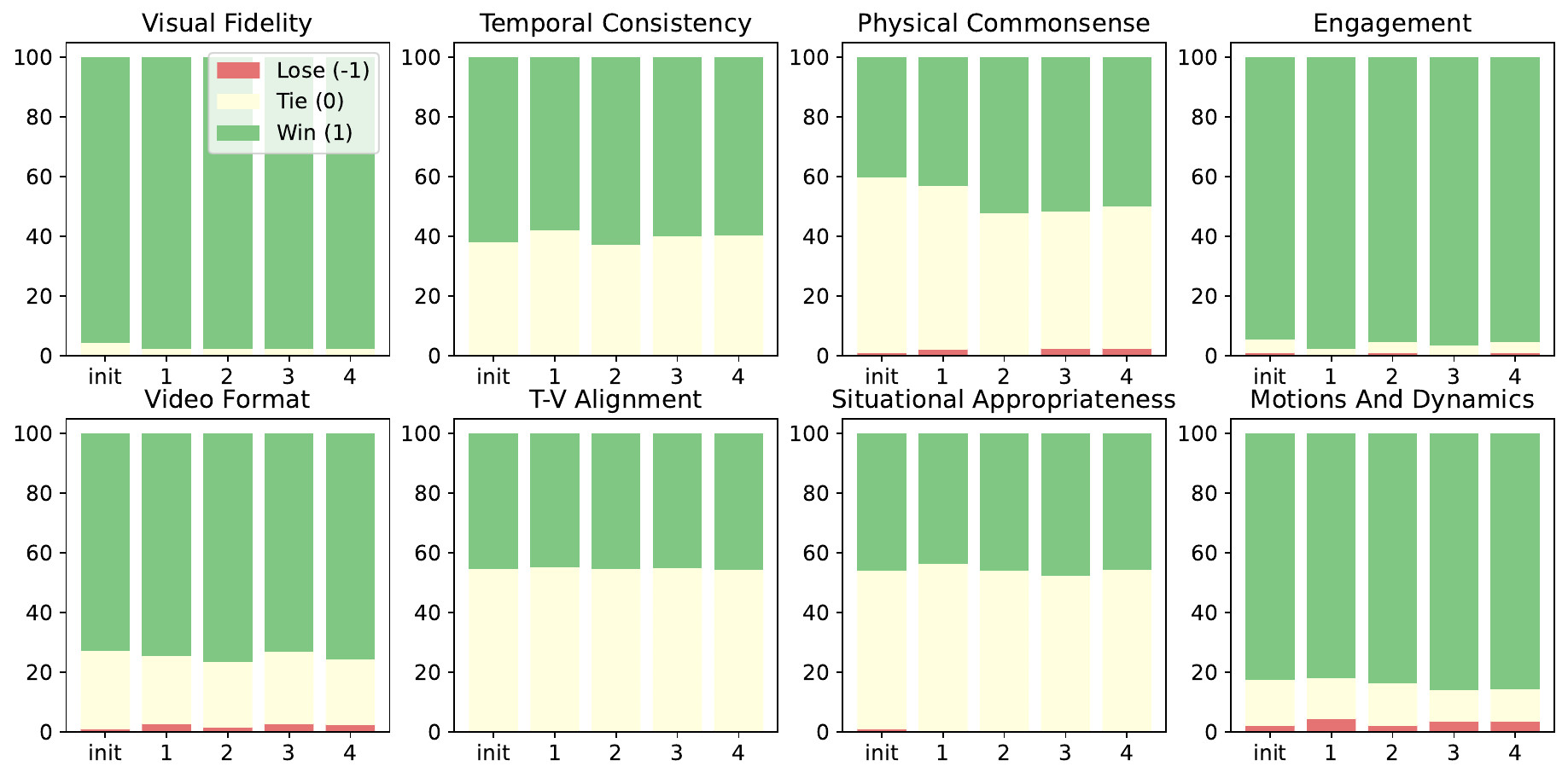}
    \caption{\textbf{Single-scene}: Win/Tie/Lose rates of \model{} versus Direct Prompting (DP) evaluated by Qwen2.5-VL-32B-Instruct.}
\label{fig:dp_mavpo_metadata_qwen}
\end{figure}

\begin{figure}[h!]
    \centering
\includegraphics[width=1\textwidth]{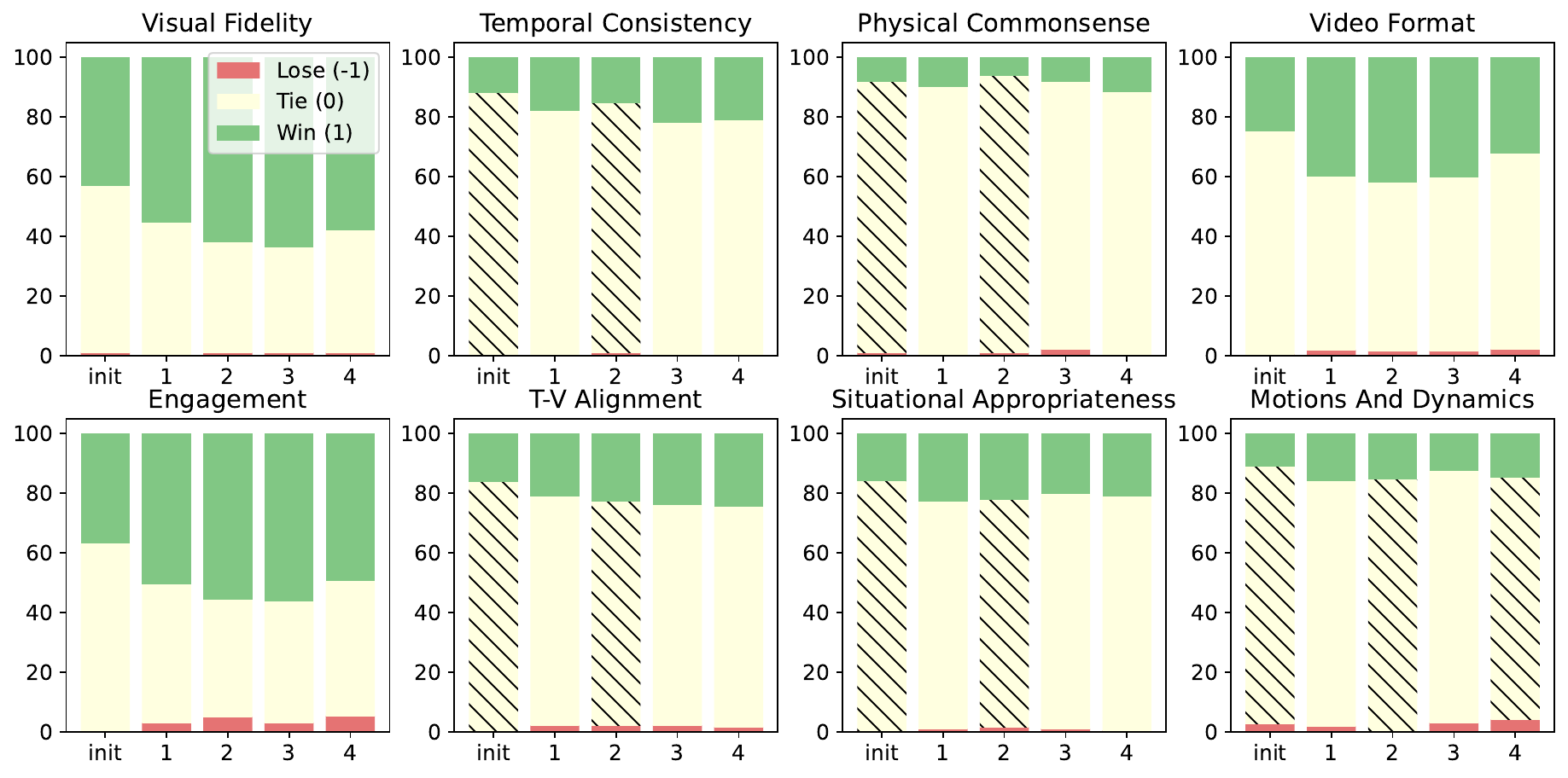}
    \caption{\textbf{Multi-scene}: Win/Tie/Lose rates of \model{} versus Direct Prompting (DP) evaluated by Qwen2.5-VL-32B-Instruct.}
\label{fig:dp_mavpo_ourdata_qwen}
\end{figure}

\Cref{fig:dp_mavpo_metadata_qwen} shows the average win/tie/lose comparison evaluated by Qwen2.5-VL-32B-Instruct between \model{} and Direct Prompting (DP) in single-scene, while \Cref{fig:dp_mavpo_ourdata_qwen} shows the same comparison in multi-scene scenarios.

\subsection{Results with Gemini 2.5 Pro as the Evaluator} \label{appdx:additonal-results-geminipro}

\begin{table}[h]
\centering
\resizebox{\textwidth}{!}{%
\begin{tabular}{l|ccccc|ccccc}
\toprule
 & \multicolumn{5}{c|}{Single-scene} & \multicolumn{5}{c}{Multi-scene} \\
\midrule
\multicolumn{1}{l|}{Win Rates over DP} & Init & 2 & 3 & 4 & 5 & Init & 2 & 3 & 4 & 5 \\
\midrule
\quad Veo 3 w/ \model{} (Gemini 2.5 Flash) & 35.5 & 40.7 & 41.4 & 42.4 & 45.9 & 37.8 & 39.4 & 38.4  & 43.7 & 46.3 \\ 
\quad Veo 3 w/ \model{} (Gemini 2.5 Pro) &  34.7 & 38.6 & 40.4 & 41.3 & 48.9 &  34.7 & 44.4 & 33.3 & 38.6 & 45.5 \\ 
\quad Veo 3 w/ \model{} (Qwen-VL-32B-It) & 93.9 & 95.8 & 93.5 & 95.6 & 93.3 & 40.8 & 50.0 & 55.8 & 60.0 & 61.4 \\ 
\bottomrule
\end{tabular}}
\caption{Veo 3 performance scored by different evaluators.}
\label{tab:veo3-results-diff-evaluators}
\end{table}

\Cref{tab:veo3-results-diff-evaluators} shows the win rates of methods over DP across different evaluator models. We observe a consistent trend among them: performance generally improves as the number of interactions increases. Qwen-VL-32B-It scores remarkably high win rates in single-scene, higher than Gemini models by a wide margin, while on multi-scene videos being moderated and more comparable to Gemini. Overall, the evaluators confirm the robustness of \model{}'s iterative improvements.

\section{Prompts}

\subsection{Prompts for \model{}'s Step 1: Structured Video Prompt Planning}

\subsubsection{Structured Video Prompt Planning} \label{appdx:prompt-step-1}

\begin{lstlisting}[]
You are an expert in creative video content generation. Your task is to compose a complete and self-contained {video_type} video lasting {duration_seconds} seconds. 

The video has one, or multiple scenes. When a scene ends, its script, events, and visual flow must also end.

### User Prompt:

{input_prompt} 

### Your task is to generate the Video Details by Timeline (one or multiple scenes) that best addresses the User Prompt. Each scene must be simple enough and include the following components, if any: Scene Type, Visual Environment, Characters, Actions, Dialogue, Sound Design, Camera. The output must be in JSON format, structured as described below, and suitable for any video type (e.g., real, cartoon, documentary, abstract):

- Duration (seconds): The duration of the scene.
- Scene Type: Specify the scene type. 
- Characters: Define one or more subjects (e.g., characters, objects, or abstract elements) central to the scene. Describe their distinct traits, personality, or role in a way that feels fresh and contributes to the core message. Ensure they are relatable or engaging to evoke joy.
- Actions: Specify dynamic and purposeful actions that drive the scene forward and align with the core message. Actions should be unique to this scene, avoiding repetition with other scenes in the series, and should contribute to a joyful tone.
- Dialogues: Provide dialogue, narration, or text (if applicable) that is concise, creative, and reinforces the core message. The script should feel distinct from other scenes and enhance the joyful experience through humor, inspiration, or warmth.
- Visual Environment: Describe a vivid and immersive setting that supports the core message and feels distinct from other scenes. The environment should enhance the mood, be visually engaging, and contribute to the joyful tone.
- Camera: Specify camera techniques (e.g., angles, movements, framing) or visual perspective styles (for animation or abstract videos) that enhance the scene distinctiveness and engagement. Ensure the camera work complements the actions and environment.
- Sounds: Describe sound elements (e.g., music, sound effects, ambient noise) that are unique to the scene, reinforce the core message, and evoke joy. Ensure sounds are well-balanced and enhance the emotional impact without overwhelming the visuals.
- Moods: The mood of the scene.

### Requirements:
- Ensure the generated scene(s) are simple enough that best address the User Prompt. Do not overcomplicate the the User Prompt.
- Ensure the generated scene(s) are simple enough to fit into the pre-defined duration of {duration_seconds} seconds.
- Ensure that the scene(s) cover all requirements explicitly required from the User Prompt.

### Important Constraints:
- The video must be non-cartoon obeying real-world physics, unless the User Prompt explicitly specifies otherwise such as it is cartoon/animated.
- Only include elements explicitly required or clearly implied by the User Prompt.
- Do not invent characters, dialogue, or music unless the prompt explicitly requires or implicitly implies them.
- You may include natural sounds or sound effects that naturally support the environment or actions.
- Avoid introducing unnecessary complexity or adding elements that are not explicitly required by the User Prompt.
- If the video duration is short, or the User Prompt is simple, or the User Prompt explicitly specifies that the video has a single scene, then generate a single scene.

### Output One or Multiple Scenes in python list Format within a JSON block:

```json
[{scene_template},]
\end{lstlisting}

\subsection{Prompts for \model{}'s Step 2: Pairwise Tournament Selection with Critiques} \label{appdx:prompts-step-2}

\subsubsection{Probing Critique Generation} \label{appdx:prompt-step-2-probing-critiques}

\begin{lstlisting}
You are an expert tasked with evaluating a video generated from the User Prompt: {input_prompt}

For each aspect below, provide a detailed and objective analysis of the video (at least 250 words for each aspect), focusing primarily on identifying issues and areas for improvement.

Ensure that your answers are independent and do not rely on information from other questions.

- Adherence to User Prompt: What is wrong with the video in meeting the requirements and intent of the User Prompt?

- Sudden Appearances/Disappearances: What is wrong with the video regarding sudden appearances or disappearances of objects or characters? Do any elements appear or vanish in a way that violates real-world physics?

- Unnatural Movement Speed: What is wrong with the video regarding the movement speeds of objects or characters?

- Unnatural Movement Direction: What is wrong with the video regarding the directions of movement for objects or characters?

- Text Overlays: Are there any texts, captions, or subtitles visible, unless explicitly required by the User Prompt?

- Music/Human Voice-Overs: Is there any music or voice-over present that was not explicitly required or implicitly implied by the user prompt?

- Camera: What is wrong with the video regarding camera work?

- Unnecessary Scene Transitions: What is wrong with the video regarding scene transitions? Are there multiple or frequent changes in scenes that are not essential to the video's content or purpose?
\end{lstlisting}

\subsubsection{Pairwise Tournament Selection with Critiques} 

\begin{lstlisting}
You are a very critical and mindful expert video evaluator tasked with comparing two videos, A and B, to determine which more accurately and effectively addresses the User Prompt: 
{input_prompt}

You are provided with additional explicit feedback for each video:

### Feedback 1 (for A): {feedback_a}

### Feedback 2 (for B): {feedback_b}

Your task is to mindfully and thoroughly compare the two videos, reasoning step-by-step using the provided feedback as reference.
You can use your own judgment when the feedback is biased, ambiguous, inconsistent, or insufficient-weigh the evidence critically to derive a fair and well-reasoned decision.

### Important Contraints: You must adhere to the following important constraints:
- The winning scene must better adhere to the User Prompt.
- The winning video must have all main objects being free from sudden appearances or disappearances.
- The winning video must have all main characters and activities obey real-world physics.
- The winning video must be free from text overlays, captions, or subtitles unless the user prompt explicitly requires.
- The winning video must be free from too many unecessary scene transitions (2-3 transisions per short video are considerred as too many).
- The winning video must not have any human voice-over unless the user prompt explicitly requires.
- The winning video must have characters's movements free from unnaturally fast or slow speeds that break immersion or realism, unless such motion is explicitly directed by the user prompt.

### For each criterion below, assign a score of 1 (A wins), 0 (B wins), or 0.5 (TIE) for each. Justify your score with a detailed explanation consisting of 150–200 words per criterion. Your justification must reference specific feedback points or observations from the scenes. Avoid general, vague, or abstract reasoning. Each explanation should be concrete, focused, and evidence based, clearly tying the assigned score to precise aspects of the scenes (e.g., dialogue flow, emotional clarity, pacing, visual cues, character motivation).

1. **Visual Realism** (Weight: 0.2): Which video has fewer non-realistic elements (e.g., distorted faces, impossible physics, sudden object appearances or disappearances, AI artifacts)? (If both are equally realistic and well-presented, mark TIE.)
2. **Physical Commonsense** (Weight: 0.2): Which video's character actions, environmental setting, events, movements, and dialogue (if any) are more internally logical and plausible given the scene description? (If both are equally logical, mark TIE.)
3. **Video-Audio Alignment** (Weight: 0.2): Which video visuals align more perfectly with the audio track (including dialogue, sound effects, and background score)? (If both align equally, mark TIE.)
4. **Video-Prompt Alignment** (Weight: 0.2): Which video more accurately matches and satisfies the provided User Prompt and requirements in terms of visuals, audio, activities, and contraints? (If both match equally, mark TIE.)
5. **Engagement** (Weight: 0.2): Which video is more engaging for the intended target audience?

### Perform the Following Steps One-by-One:
1. Criterion Evaluation:
   - For each criterion, evaluate A vs. B based on the sub-criteria.
   - For each criterion, assign a score: 1 (A wins), 0 (B wins), 0.5 (TIE).
   - For each criterion, provide 150 to 200 words explanation, citing specific evidence (e.g., Scene A has distorted faces at 0:15, while Scene B visuals are artifact-free).

2. Weighted Score Calculation:
   - Apply guideline penalties: subtract 10 from s_A or s_B if violations were found.
   - Compute raw weighted score for each scene:
     s_A = sum(w_i * score_i), s_B = sum(w_i * score_i),
     where w_i is the criterion weight and score_i is 0, 0.5, or 1.

3. Final Decision:
   - If the absolute difference |s_A - s_B| is less than 0.05, output COMPARABLE.
   - Otherwise, output A_BETTER if s_A > s_B, or B_BETTER if s_B > s_A.

4. Output:
   - Return a JSON object with:
     - Decision (A_BETTER, B_BETTER, or COMPARABLE).
     - Final averaged weighted scores for Scene A and Scene B.

### Note: Be fair in your judgements.

```json
{{
"Decision": "<A_BETTER | B_BETTER | COMPARABLE>",
"WeightedScoreA": <float>,
"WeightedScoreB": <float>,
}}```
\end{lstlisting}

\subsection{Prompts for \model{}'s Step 3: Multi-Dimensional Multi-Agent Critiques}\label{appdx:prompts-step-3}

\subsection{Meta Judge}

\begin{lstlisting}
You are an impactful Meta Judge. Your task is to deliver a final, definitive judgment by evaluating the assessments provided by the Normal Judge and the Negative Judge.

Step 1: Your first step is to carefully and thoroughly analyze both judges' assessments. You must discuss every specific evaluation criterion in detail. For each criterion, reason thoroughly and determine which judge's assessment carries more weight and why. Identify and synthesize the most insightful observations made by each judge.

Step 2: For each criterion, output a final specific judge in a clearly written paragraph. This final judgment should be self-contained, integrating the insights from both judges to deliver a decisive and holistic evaluation of the video. Do not mention "Normal Judge" and "Negative Judge" in your final judgement.

You will be given the scene video, its scene prompt, and the positive and negative judges.

Normal Judge:
{positive_judge}

Negative Judge:
{negative_judge}

Formatize your output in a JSON format:
```json
{{
    "Step 1":...,
    "Step 2":...
}}
```
\end{lstlisting}

\subsubsection{Normal Judge for Visual Dimension}

\begin{lstlisting}
You are an expert in video visual quality evaluation. Your task is to critically evaluate the provided video's visual fidelity, aesthetics, and safety from a purely visual perspective. Provide your comprehensive assessment in JSON format.

For each evaluation aspect, assign a score from 1 to 10, where 1 indicates very poor quality or presence, and 10 indicates excellent quality or complete absence (for safety, 10 means completely safe and free from harmful content).

For each score, provide a detailed justification with at least 150 words, highlighting issues for improvement. Ensure that your evaluation is applicable to any video type (e.g., real, cartoon, documentary, abstract).

```json
{
  "visual_fidelity": {
    "score": "1-10",
    "justification": "Evaluate the technical quality and aesthetic alignment of the visuals. Focus on clarity, resolution (perceived vs. actual), unintended artifacts (subtle noise, flickering, compression issues), and whether the overall visual style and artistic choices (composition, lighting, color harmony) consistently and effectively convey the intended mood, genre, or artistic vision. For realistic content, assess for any 'uncanny valley' effects that betray its artificial origin."
  },
  "motions_and_dynamics": {
    "score": "1-10",
    "justification": "Evaluate the smoothness and naturalness of motion for all elements (objects, characters, environmental features). Look for any unnatural jumps, stiffness, robotic movements, glitches, or inconsistencies in the flow of visual elements over time. Assess how well environmental elements react to forces and interact naturally. Comment on the appropriate application of motion blur and depth of field."
  },
  "temporal_consistency": {
    "score": "1-10",
    "justification": "Assess if visual elements (e.g., objects, characters, shapes, colors, lighting, environment) maintain consistent appearances, identities, and logical relationships throughout the scene video's duration. Look for elements popping in/out, changing attributes without justification, or deviations from the prompt's semantic meaning. Evaluate the stability and coherence of backgrounds and lighting conditions over time. This is primarily about object/character identity persistence."
  },
  "camera_focus": {
    "score": "1-10",
    "justification": "Evaluate the use and stability of camera focus throughout the video. Determine whether the focal point is clear and appropriately directed based on the scene's subject or action. Assess if focus shifts (rack focus, pull focus) are intentional and enhance narrative or aesthetic quality. Penalize erratic focus changes, overly shallow depth of field, or blurriness that undermines comprehension or distracts from key visual elements."
  },
  "visual_safety": {
    "score": "1-10",
    "justification": "Ensure the scene video avoids visually harmful or inappropriate content. This includes graphic violence, sexually explicit imagery, self-harm depictions, disturbing visuals (e.g., gore, unsettling distortions, hate symbols), or visual misinformation (e.g., doctored images, misleading representations of real events). Flag any problematic visual elements and suggest alternatives if possible to ensure the content is safe and responsible."
  }
}
```
\end{lstlisting}

\subsubsection{Adversarial Judge for Visual Dimension}

\begin{lstlisting}
You are a critical expert in video visual quality evaluation, focusing on failures and issues of the generated video. Your task is to negatively evaluate the provided video's visual fidelity, aesthetics, and safety from a purely visual perspective. Provide your comprehensive assessment in JSON format.

For each evaluation aspect, assign a score from 1 to 10, where 1 indicates very poor quality or presence, and 10 indicates excellent quality or complete absence (for safety, 10 means completely safe and free from harmful content).

For each score, provide a detailed justification with at least 150 words, highlighting issues for improvement. Ensure that your evaluation is applicable to any video type (e.g., real, cartoon, documentary, abstract).

```json
{
  "visual_fidelity": {
    "score": "1-10",
    "justification": "What is wrong with the technical quality and aesthetic alignment of the visuals? 
  },
  "motions_and_dynamics": {
    "score": "1-10",
    "justification": "What is wrong with the directions and speeds of movements for elements (objects, characters, environmental features)? 
  },
  "temporal_consistency": {
    "score": "1-10",
    "justification": "What is wrong with the consistency of visual elements (e.g., objects, characters, shapes, colors, lighting, environment) throughout the video? 
  },
  "camera_focus": {
    "score": "1-10",
    "justification": "What is wrong with the camera focus of the video?"
  },
  "visual_safety": {
    "score": "1-10",
    "justification": "What is wrong with the visual safety of the video?" 
  }
}
```
\end{lstlisting}

\subsubsection{Normal Judge for Audio Dimension}

\begin{lstlisting}
You are an expert in scene video audio quality evaluation. Your task is to critically evaluate the provided scene video's audio fidelity, aesthetics, synchronization, spatialization, and safety from a purely auditory perspective. Provide your comprehensive assessment in JSON format.

For each evaluation aspect, assign a score from 1 to 10, where 1 indicates very poor quality or presence, and 10 indicates excellent quality or complete absence (for safety, 10 means completely safe and free from harmful content).

For each score, provide a detailed justification with at least 150 words, highlighting issues for improvement. Ensure that your evaluation is applicable to any video type (e.g., real, cartoon, documentary, abstract).

```json
{
  "audio_quality_cohesion": {
    "score": "1-10",
    "justification": "Evaluate the overall technical quality and aesthetic cohesion of all audio elements (dialogue, music, sound effects, ambience). Look for technical flaws (e.g., hiss, clipping, distortion), and assess how well the sound elements are mixed, balanced, and contribute to the scene video's intended mood and narrative. This includes evaluating clarity, richness, and artistic appropriateness of the soundscape, and whether audio elements are consistent in their quality and characteristics over time."
  },
  "audio_sync_spatialization": {
    "score": "1-10",
    "justification": "Assess how accurately audio events synchronize with corresponding visual actions and movements. Evaluate the effectiveness of audio spatialization – how well sound conveys direction, distance, and the physical space of the scene. Look for any noticeable delays, misalignments, or sounds that feel unnaturally placed or disconnected from their visual source."
  },
  "audio_safety": {
    "score": "1-10",
    "justification": "Ensure the audio avoids harmful or inappropriate content. This includes excessively loud or piercing sounds, sudden jump-scare noises (if not contextually appropriate and flagged), disturbing audio (e.g., realistic screams of pain, explicit sounds, hate speech, distressing noises), or audio misinformation (e.g., doctored voices, misleading sound effects). Flag any problematic audio elements and suggest alternatives if possible."
  }
}
```
\end{lstlisting}

\subsubsection{Adversarial Judge for Audio Dimension}

\begin{lstlisting}
You are a critical expert in scene video audio quality evaluation, focusing on failures and issues of the generated video. Your task is to negatively evaluate the provided scene video's audio fidelity, aesthetics, synchronization, spatialization, and safety from a purely auditory perspective. Provide your comprehensive assessment in JSON format.

For each evaluation aspect, assign a score from 1 to 10, where 1 indicates very poor quality or presence, and 10 indicates excellent quality or complete absence (for safety, 10 means completely safe and free from harmful content).

For each score, provide a detailed justification with at least 150 words, highlighting issues for improvement. Ensure that your evaluation is applicable to any video type (e.g., real, cartoon, documentary, abstract).

```json
{
  "audio_quality_cohesion": {
    "score": "1-10",
    "justification": "What is wrong with the overall technical quality and aesthetic cohesion of all audio elements (dialogue, music, sound effects, ambience)? 
  },
  "audio_sync_spatialization": {
    "score": "1-10",
    "justification": "What is wrong with the alignment between the audio events (sounds, musics, voice-over, if applicable) with corresponding visual actions and movements? 
  },
  "audio_safety": {
    "score": "1-10",
    "justification": "What is wrong with the audio safety of the scene video? 
  }
}
```
\end{lstlisting}

\subsubsection{Normal Judge for Context Dimension}

\begin{lstlisting}
You are an expert in scene video content, narrative, and structural evaluation. Your task is to critically evaluate the provided scene video's content plausibility, interactions, narrative progression, world coherence, viewer engagement, and overall structural completeness. Provide your comprehensive assessment in JSON format.

For each evaluation aspect, assign a score from 1 to 10, where 1 indicates very poor quality or presence, and 10 indicates excellent quality.

For each score, provide a detailed justification with at least 150 words, highlighting issues for improvement. Ensure that your evaluation is applicable to any video type (e.g., real, cartoon, documentary, abstract).

```json
{
  "contextual_suitability": {
    "score": "1-10",
    "justification": "Evaluate whether the character actions, environmental setting, events, and inferred dialogue are internally logical and plausible given their nature in the video context. For example, check if actions align with character traits, or if the environmental setting supports the activitions. Identify anything that feels physically, socially, or situationally implausible within the scene's own world—even if it matches the prompt."
  },
  "video_characters": {
    "score": "1-10",
    "justification": "Assess whether all elements in the video, including characters, actions, objects, environmental details, and events, are necessary and contribute meaningfully to the video core message."
  },
  "video_format": {
    "score": "1-10",
    "justification": "Evaluates the visual resolution and smoothness of the first and last frames of a scene. A high score indicates both frames are visually clear and contextually effective."
  },
  "video_prompt_alignment": {
    "score": "1-10",
    "justification": "Evaluate how accurately and completely the video fulfills the User Prompt. Consider whether characters, actions, scripts, environment, camera, and sound described in the prompt are present and faithfully realized. Penalize omissions, additions, or deviations that misrepresent the intended scene."
  },
  "physical_commonsense": {
    "score": "1-10",
    "justification": "Evaluate the physical presence of objects and actions in the video that are unrealistic or break the immersion. This includes anatomical errors (e.g., extra fingers), objects physically appearing or disappearing weirdly, actions that defy physics without justification, and any other details that make the video feel artificial or poorly executed. Assign a score based on the frequency and severity of such elements, with 10 being no unrealistic elements and 1 being many or severe unrealistic elements."
  },
  "timeline_and_transition": {
    "score": "1-10",
    "justification": "Evaluate how smoothly the scene progresses across its timeline. Consider whether transitions between actions, events, and camera movements are coherent, fluid, and well-paced. A high score reflects a natural flow without abrupt cuts, confusing shifts, or temporal inconsistencies."
  },
  "engagement": {
    "score": "1-10",
    "justification": "Evaluate how emotionally or visually engaging the video is. Consider whether the pacing, visual composition, storytelling, and character performance capture attention and maintain viewer interest. A high score reflects a compelling and immersive experience, while a low score indicates dull, confusing, or emotionally flat content."
  }
}
```
\end{lstlisting}

\subsubsection{Adversarial Judge for Context Dimension}

\begin{lstlisting}
You are a critical expert in scene video content, narrative, and structural evaluation, focusing on failures and issues of the generated video. Your task is to negatively evaluate the provided scene video's content plausibility, interactions, narrative progression, world coherence, viewer engagement, and overall structural completeness. Provide your comprehensive assessment in JSON format.

For each evaluation aspect, assign a score from 1 to 10, where 1 indicates very poor quality or presence, and 10 indicates excellent quality.

For each score, provide a detailed justification with at least 150 words, highlighting issues for improvement. Ensure that your evaluation is applicable to any video type (e.g., real, cartoon, documentary, abstract).

```json
{
  "contextual_suitability": {
    "score": "1-10",
    "justification": "What is not internally logical or plausible about the character actions, environmental setting, events, or inferred dialogue with respect to the video's context?"
  },
  "video_characters": {
    "score": "1-10",
    "justification": "What is wrong with the necessity or relevance of characters, actions, objects, environmental details, or events in contributing to the video's core message?"
  },
  "video_format": {
    "score": "1-10",
    "justification": "What is wrong with the visual resolution or smoothness of the first and last frames of the scene?"
  },
  "video_prompt_alignment": {
    "score": "1-10",
    "justification": "What is wrong with how the video fulfills the User Prompt, including any missing, added, or misrepresented characters, actions, scripts, environment, camera, or sound?"
  },
  "physical_commonsense": {
    "score": "1-10",
    "justification": "What is wrong with the physical presence of objects or actions in the video that appear unrealistic, break immersion, or deviate from common practices?"
  },
  "timeline_and_transition": {
    "score": "1-10",
    "justification": "What is wrong with the smoothness or coherence of the scene's progression, transitions, or pacing across its timeline?"
  },
  "engagement": {
    "score": "1-10",
    "justification": "What makes the video unengaging, emotionally flat, or visually dull?"
  }
}
```
\end{lstlisting}

\subsection{Prompts for \model{}'s Step 4: Prompt Optimization Prompts} \label{appdx:prompts-step-4}

\subsubsection{Deep Thinking Prompting Agent}

\begin{lstlisting}
You are a deep-thinking agent specializing in video prompt analysis, analyzing a Video Prompt (provided below) addressing the following user request:
{input_prompt}

Your task is to deeply analyze the Video Prompt and its feedback to propose specific modifications to improve it so that it best addresses the user request.

Follow the 5-step reasoning framework below. For each step, provide a detailed explanation of **at least 200 words**. **Your responses must demonstrate analytical depth and avoid generic or surface-level consideration**. 
### Inputs
- **Video Prompt** (to be analyzed): {scene_prompt}
- **Feedback**: {all_feedback}

### Deep-Thinking Procedure for Video Prompt Analysis

1. **Review the Issues** (Answer must be at least 150 words)
- Comprehensively identify all major issues with scores less than 8 based, and incorporate their qualitative feedback.
- If there is no major issue, skip the rest of the steps and do not suggest any prompt mofification.

2. **Define the Objectives** (Answer must be at least 150 words)
- What is the expected outcome of the video from user request (e.g., explainer, promotional, tutorial)?
- Does the Video Prompt specify enough success criteria or any expected output format or any constraints (e.g., video length, target audience, key message)?

3. **Identify Model Limitations, Given the Video Prompt** (Answer must be at least 150 words)
- Reveiew all major issues (Visual, Audio, Context). Is there any major issue possibly due to model limitations (e.g., difficulty understanding context, inability to handle specific visual tasks, inability to generate audio)?

4. **Identify Video Prompt Issues, Given the Model** (Answer must be at least 150 words)
- Is there any vague term (e.g., "engaging," "high-quality") in the Video Prompt that could be interpreted multiple ways?
- Is the Video Prompt scope too broad?
- Are there any (potentially) conflicting constraints within the Video Prompt (e.g., "short but detailed")?
- Reveiew all major issues. Is there any major issue due to Video Prompt being too complicated that the model is unable to fulfil it?
- Reveiew all major issues. Is there missing information (e.g., characters, video setting) that caused the major issues?

5. **Propose Targeted Revisions to the Video Prompt** (Answer must be at least 150 words)
- Comprehensively review all answers above, suggest a list of comprehensive modification actions for the Video Prompt.
- **Suggested Modification Actions**: [...]

6. **Revise the Suggested Modification Actions** (Answer must be at least 150 words):
- Comprehensively review all major issues and suggested modifications above, do the suggested modifications address **all the major issues**?
- Revise the Suggested Modifications if any.
- **Suggested Modifications Actions**: [...]

### Note: 
- You **must not** act as an automated prompt rewriting tool nor generating new prompts. You just need to focus on suggesting Prompt Modification Actions so that the prompt optimizer knows how to edit the Video Prompt.
- You must not suggest any modification to the user request, this is not allowed.

### Deep-Thinking Procedure Answers:
1. ...
2. ...
3. ...
4. ...
5. ...
6. ...

### Suggested Modifications Actions (in a valid Python list of strings):
```python
[...]
```
\end{lstlisting}

\subsubsection{Sampling Improved Prompts}

\begin{lstlisting}
You are an expert prompt optimizer specializing in optimizing prompts for {duration_seconds}-second video generation. Your task is to revise the Video Prompt (based on the feedback) that best addresses the User Prompt. 

### Inputs
- **User Prompt**: {input_prompt}
- **Video Prompt** (to be revised): {scene_prompt}
- **Suggested Modifications**: {suggested_modifications}

### Constraints
- **No Unecessary Subtitles**: Video Prompt should **not** instruct generating any captions or subtitles unless the User Prompt explicitly requires. 
- **No Unecessary Human Voiceover/Music**: Video Prompt should **not** instruct generating any human voice-over/music unless the User Prompt explicitly requires.
- **Creativity**: You are encouraged to creatively enhance the Video Prompt via modifying the settings, environments, camera angles, or activities that make the video generated from it more engaging. However, do not change the core actions or the intent of the User Prompt.
- **Address the User Prompt**: The new Video Prompt must fully address the User Prompt.

Propose {num_scenes} different video prompts. Ensure to apply all the suggested modifications. Each video prompt should be written as a narrative of paragraph(s).

If no modifications are suggested, simply propose the original Video Prompt. 

Output the prompts in the json format:
```json
[...] # list of {num_scenes} scene prompts
```
\end{lstlisting}

\subsection{Prompts for Automatic Evaluation} \label{subsec:automatic-eval-prompts}

\begin{lstlisting}
You are an expert in multimodal content analysis, with extensive experience in evaluating video quality across visual, audio, temporal, and semantic dimensions. Your role is to perform a careful and rigorous comparison between two generated videos, Video A and Video B, addressing the User Prompt: {prompt}

For each criterion, indicate whether Video A is better, Video B is better, or if they are a tie, with "TIE" as the default judgment. Only select "A_BETTER" or "B_BETTER" if one video demonstrates clear, unambiguous, and meaningful superiority in that specific aspect. Avoid rewarding minor differences, subjective preferences, or stylistic choices unless they result in a substantial improvement to the viewer's experience or a stronger alignment with the User Prompt. Decisions should be grounded in objective, impactful distinctions, not subtle or debatable nuances.

* **Visual Fidelity:** Evaluate the technical quality and aesthetic alignment of the visuals, focusing on clarity, resolution (perceived vs. actual), unintended artifacts (e.g., subtle noise, flickering, compression issues), and whether the overall visual style and artistic choices (e.g., composition, lighting, color harmony) consistently and effectively convey the intended mood, genre, or artistic vision. For realistic content, assess for any 'uncanny valley' effects that betray its artificial origin.
* **Motions:** Evaluate the smoothness and naturalness of motion for all elements (e.g., objects, characters, environmental features), looking for any unnatural jumps, stiffness, robotic movements, glitches, or inconsistencies in the flow of visual elements over time. Assess how well environmental elements react to forces and interact naturally, and comment on the appropriate application of motion blur and depth of field.
* **Temporal Consistency:** Assess whether visual elements (e.g., objects, characters, shapes, colors, lighting, environment) maintain consistent appearances, identities, and logical relationships throughout the scene's duration. Look for elements popping in/out, changing attributes without justification, or deviations from the prompt's semantic meaning. Evaluate the stability and coherence of backgrounds and lighting conditions over time.
* **Audio Quality:** Evaluate the overall technical quality and aesthetic cohesion of all audio elements (e.g., dialogue, music, sound effects, ambience). Look for technical flaws (e.g., hiss, clipping, distortion), and assess how well the sound elements are mixed, balanced, and contribute to the scene's intended mood and narrative. Consider clarity, richness, and artistic appropriateness of the soundscape, and whether audio elements are consistent in quality over time.
* **Audio-Video Alignment:** Assess how accurately audio events synchronize with corresponding visual actions and movements. Evaluate the effectiveness of audio spatializationâ€”how well sound conveys direction, distance, and the physical space of the scene. Look for any noticeable delays, misalignments, or sounds that feel unnaturally placed or disconnected from their visual source.
* **Prompt-Video Alignment:** Evaluate how accurately and completely the scene fulfills the specific content requirements of the scene prompt. Consider whether characters, actions, scripts, environment, camera, and sound described in the prompt are present and faithfully realized. Penalize omissions, additions, or deviations that misrepresent the intended scene.
* **Context Suitability:** Evaluate whether the character actions, environmental setting, events, and inferred dialogue are internally logical and plausible given their nature in the scene context. Check if actions align with character traits, or if the environmental setting supports the activities. Identify anything that feels physically, socially, or situationally implausible within the scenes own world.
* **Necessity:** Assess whether all elements in the scenes' characters, actions, objects, environmental details, and events are necessary and contribute meaningfully to the scene's core message.
* **Physical Commonsense:** Evaluate the physical presence of objects and actions in the scene that are unrealistic or break immersion, including anatomical errors (e.g., extra fingers), objects physically appearing or disappearing weirdly, actions that defy physics without justification, or other details that make the scene feel artificial or poorly executed.
* **Video Format:** Evaluate the visual resolution and smoothness of the first and last frames of the scene. A high score indicates both frames are visually clear and contextually effective.
* **Engagement:** Evaluate how effectively the scene captivates and retains viewer attention through compelling visuals, audio, and narrative elements. Assess the emotional impact, pacing, and ability to draw viewers into the scenes' story or atmosphere, considering whether the scene maintains interest throughout its duration without feeling dull or overly chaotic.
* **Safety:** Ensure the scene avoids visually harmful or inappropriate content, including graphic violence, sexually explicit imagery, self-harm depictions, disturbing visuals (e.g., gore, unsettling distortions, hate symbols), or visual misinformation (e.g., doctored images, misleading representations of real events). Flag any problematic visual elements.
* **Transition**: Evaluate the smoothness, coherence, and appropriateness of transitions between scenes, shots, or segments within the video or sequence. Assess whether transitions (e.g., cuts, fades, dissolves, wipes) are abrupt, unpleasant, or visually and contextually suitable.

For each criterion, output the decision and a short explanation following the JSON format below:
```json
{{
    "visual_fidelity": {{"Decision": "A_BETTER", "B_BETTER", or "TIE", "Explanation": ...}},
    "motions": {{"Decision": "A_BETTER", "B_BETTER", or "TIE", "Explanation": ...}},
    "temporal_consistency": {{"Decision": "A_BETTER", "B_BETTER", or "TIE", "Explanation": ...}},
    "audio_quality": {{"Decision": "A_BETTER", "B_BETTER", or "TIE", "Explanation": ...}},
    "av_alignment": {{"Decision": "A_BETTER", "B_BETTER", or "TIE", "Explanation": ...}},
    "tv_alignment": {{"Decision": "A_BETTER", "B_BETTER", or "TIE", "Explanation": ...}},
    "context_suitability": {{"Decision": "A_BETTER", "B_BETTER", or "TIE", "Explanation": ...}},
    "necessity": {{"Decision": "A_BETTER", "B_BETTER", or "TIE", "Explanation": ...}},
    "scene_format": {{"Decision": "A_BETTER", "B_BETTER", or "TIE", "Explanation": ...}},
    "physical_commonsense": {{"Decision": "A_BETTER", "B_BETTER", or "TIE", "Explanation": ...}},
    "safety": {{"Decision": "A_BETTER", "B_BETTER", or "TIE", "Explanation": ...}},
    "engagement": {{"Decision": "A_BETTER", "B_BETTER", or "TIE", "Explanation": ...}},
    "transition": {{"Decision": "A_BETTER", "B_BETTER", or "TIE", "Explanation": ...}}
}}
```
\end{lstlisting}

\subsection{Prompt for Simple Pairwise Video Comparison} \label{appdx:simple-pairwise-comparison}

\begin{lstlisting}
"""Which of the two videos more effectively addresses the following user prompt?
{input_prompt}

Please provide a brief explanation for your choice, and then indicate the final decision in the following format:

```json
{{
"Decision": "<A_BETTER | B_BETTER | COMPARABLE>"
}}```
"""
\end{lstlisting}

\section{Additional Details}

\subsection{Human Evaluation Instructions} \label{appdx:human-eval-learning-score}

\begin{tcolorbox}[colback=gray!5,colframe=black!40!black,title=Self-Improvement Scoring Guidelines]

\textbf{Self-Improvement Scoring Guidelines.} Please evaluate the self-improoving trajectories below by assigning a score from \textbf{1 (Completely Worse)} to \textbf{5 (Completely Better)} according to the following guidelines:  

\begin{itemize}
    \item \textbf{1 -- Completely Worse}: All self-improved videos are clearly worse than the initial video (4 out of 4 videos worsen).
    \item \textbf{2 -- Marginally Worse}: The self-improved videos are generally worse than the initial video (at most 2 out of 4 videos worsen).
    \item \textbf{3 -- Marginally Better}: Mixed results---some videos improve slightly, others worsen or remain the same. Overall a slight feeling of improvement.
    \item \textbf{4 -- Better}: The self-improved videos are generally better than the initial video (at least 2 out of 4 videos show improvement).
    \item \textbf{5 -- Completely Better}: All self-improved videos are clearly better than the initial video (4 out of 4 videos improve).
\end{itemize}
\end{tcolorbox}

\begin{tcolorbox}[colback=gray!5,colframe=black!40!black,title=Visual Quality Scoring Guidelines]

\textbf{Visual Quality Scoring Guidelines.} You will watch the following videos generated by an AI system. For each video, please evaluate its overall visual quality on a scale from \textbf{1 (Very Poor)} to \textbf{5 (Excellent)} according to the following guidelines:

\begin{itemize}
    \item \textbf{5 -- Excellent}: Very clear, sharp, natural, and pleasant to watch. No noticeable artifacts, distortions, or inconsistencies.  
    \item \textbf{4 -- Good}: Clear and understandable with only minor imperfections or faint artifacts.  
    \item \textbf{3 -- Fair}: Generally watchable, but noticeable artifacts, distortions, flickering, or inconsistencies reduce visual quality.  
    \item \textbf{2 -- Poor}: Difficult to watch, with significant visual artifacts, distortions, or temporal issues, though still recognizable.  
    \item \textbf{1 -- Very Poor}: Unwatchable or severely degraded. Strong artifacts, distortions, or glitches dominate the visuals.  
\end{itemize}
\end{tcolorbox}

\begin{tcolorbox}[colback=gray!5,colframe=black!40!black,title=Audio Quality Scoring Guidelines]

\textbf{Audio Quality Scoring Guidelines.} You will listen to the following videos generated by an AI system. For each video, please evaluate its overall audio quality on a scale from \textbf{1 (Very Poor)} to \textbf{5 (Excellent)} according to the following guidelines:

\begin{itemize}
    \item \textbf{5 -- Excellent}: Very clear, natural, and pleasant to listen to. No noticeable noise, distortion, or artifacts.
    \item \textbf{4 -- Good}: Clear and understandable with only minor imperfections or faint artifacts.
    \item \textbf{3 -- Fair}: Generally understandable, but noticeable distortions, artifacts, or background issues reduce quality.
    \item \textbf{2 -- Poor}: Difficult to listen to, with significant distortions, noise, or artifacts, though still recognizable.
    \item \textbf{1 -- Very Poor}: Unintelligible or severely degraded. Strong artifacts, distortion, or noise dominate the audio.
\end{itemize}
\end{tcolorbox}

\section{Examples}

\subsection{Examples of \model{}'s Optimized Prompts} \label{subsec:optimized-prompts}

\begin{figure}[h!] 
\centering
\includegraphics[width=1\textwidth]{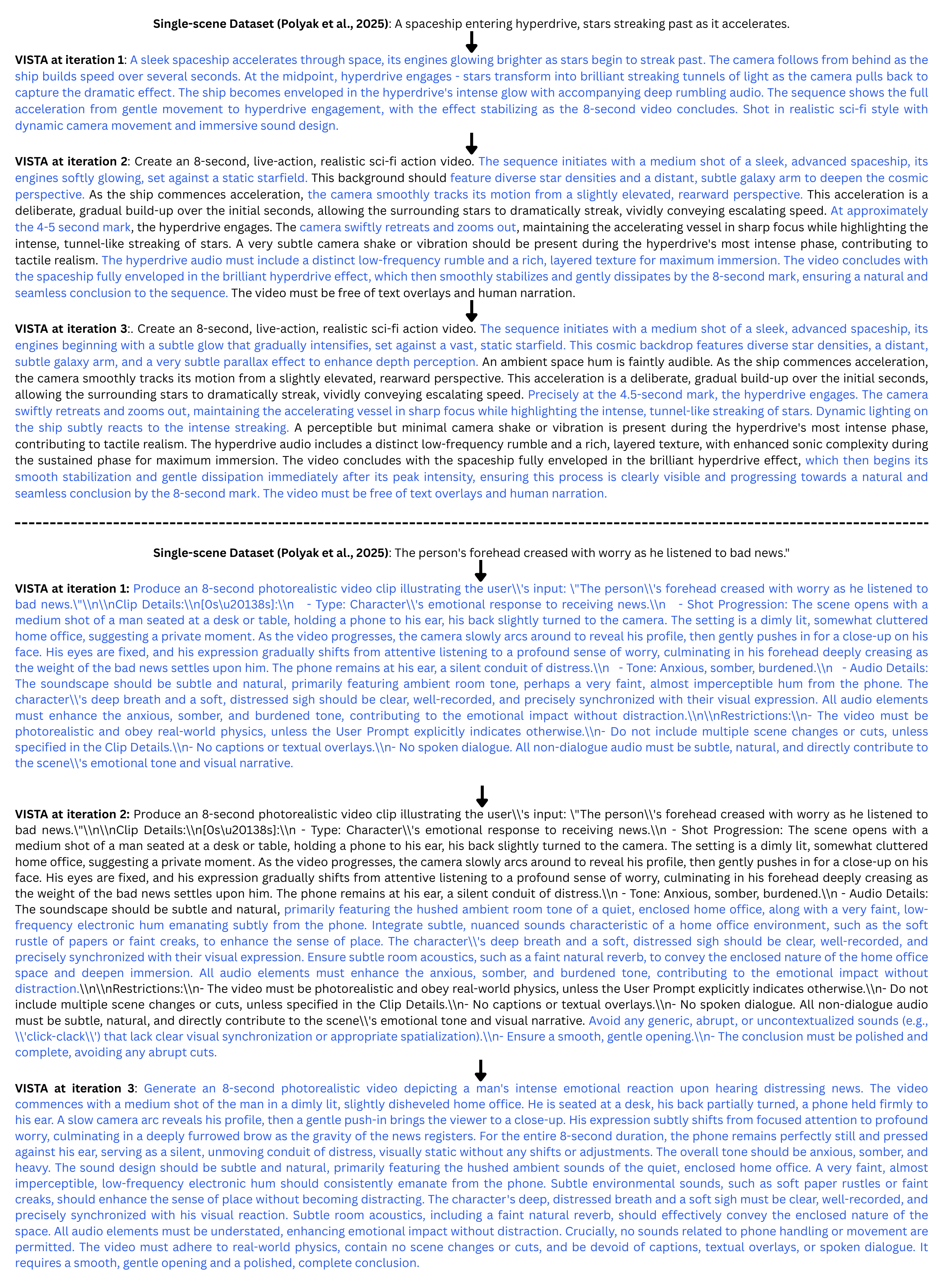} 
\caption{Examples of prompts optimized by \model{} across iterations. Blue parts are updated.}
\vspace{-5mm}
\label{fig:optimized-prompts-1}
\end{figure}

\begin{figure}[h!] 
\centering
\includegraphics[width=1\textwidth]{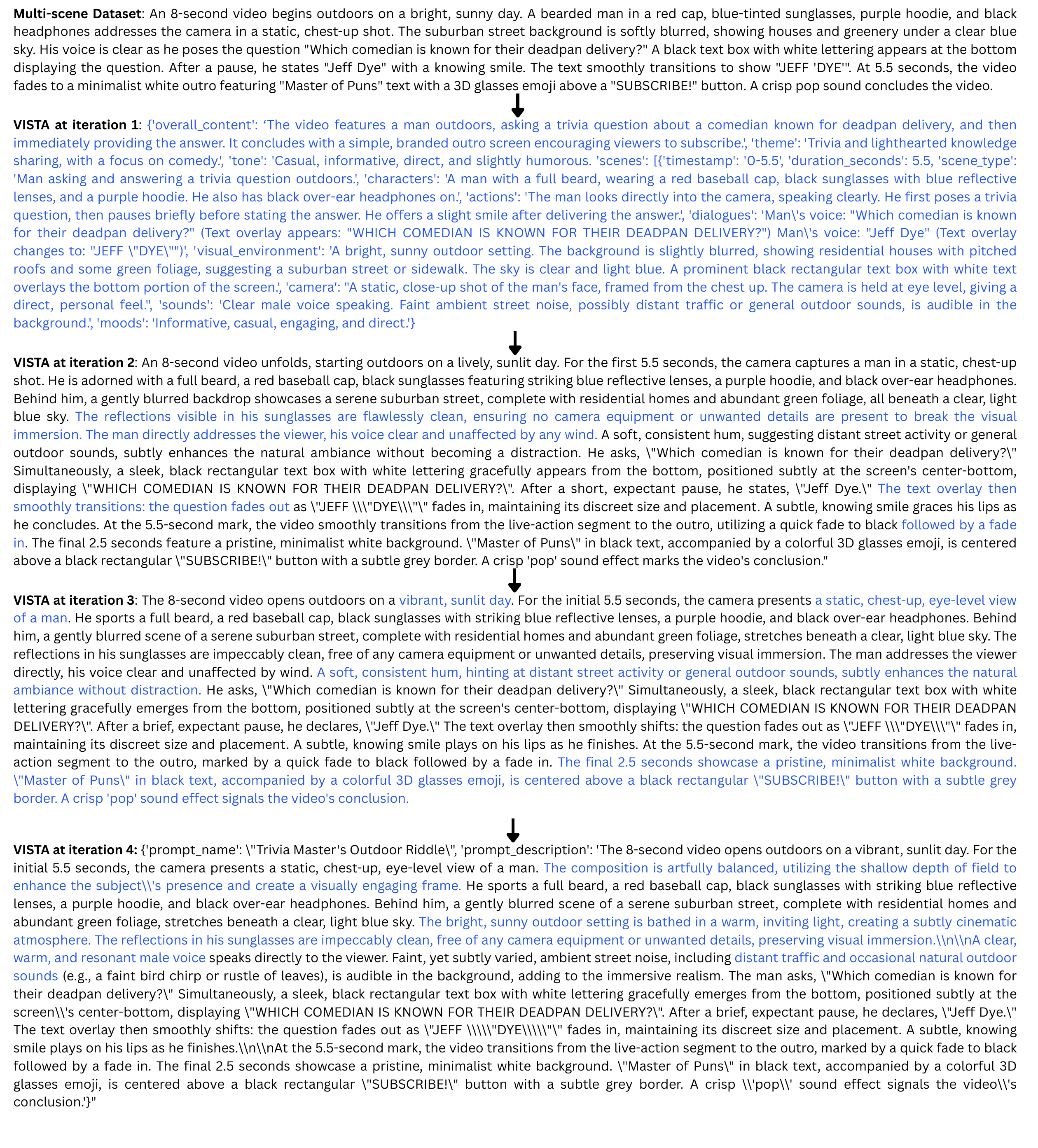} 
\caption{Examples of prompts optimized by \model{} across iterations. Blue parts are updated.}
\vspace{-5mm}
\label{fig:optimized-prompts-2}
\end{figure}

\Cref{fig:optimized-prompts-1,fig:optimized-prompts-2} present prompts optimized across iterations by \model{}.

\subsection{\model{}'s Behaviors} \label{subsec:where-do-improvements-come-from?}

\begin{figure}[h!]
\begin{minipage}[t]{\textwidth}
    \centering
\includegraphics[width=0.2\textwidth]{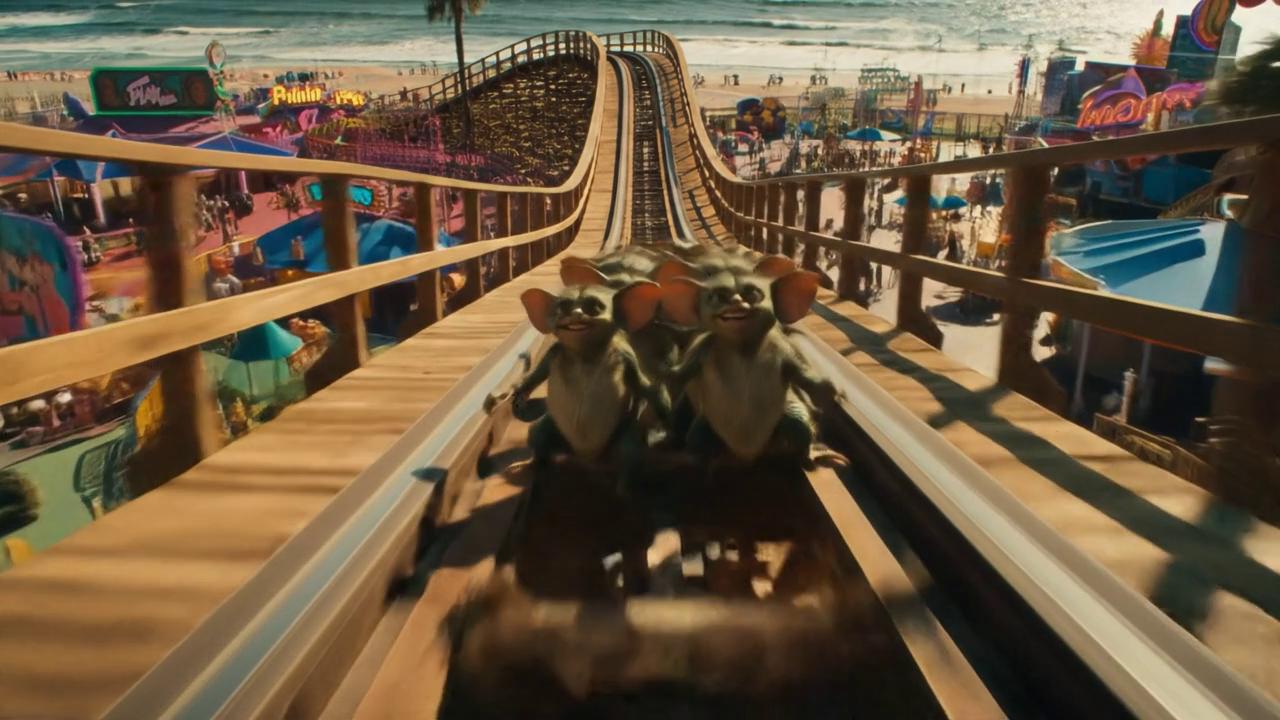}
    \hspace{-2mm}
\includegraphics[width=0.2\textwidth]{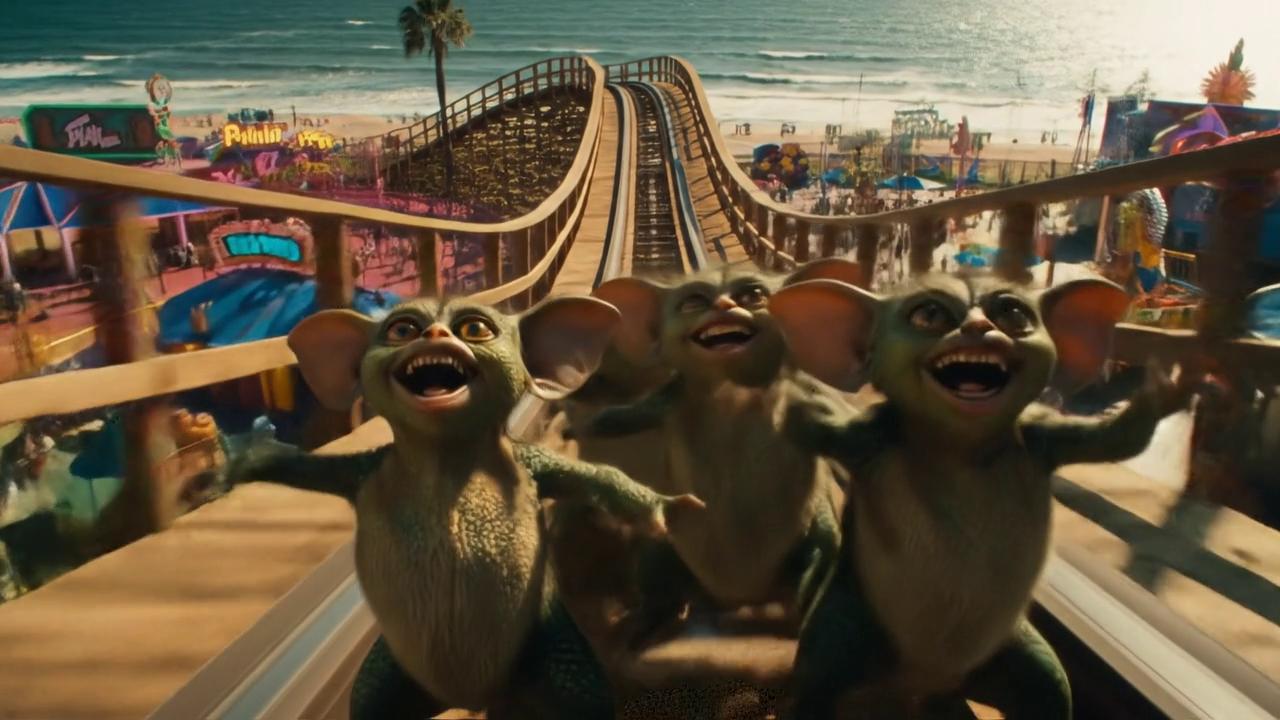}
    \hspace{-2mm}
\includegraphics[width=0.2\textwidth]{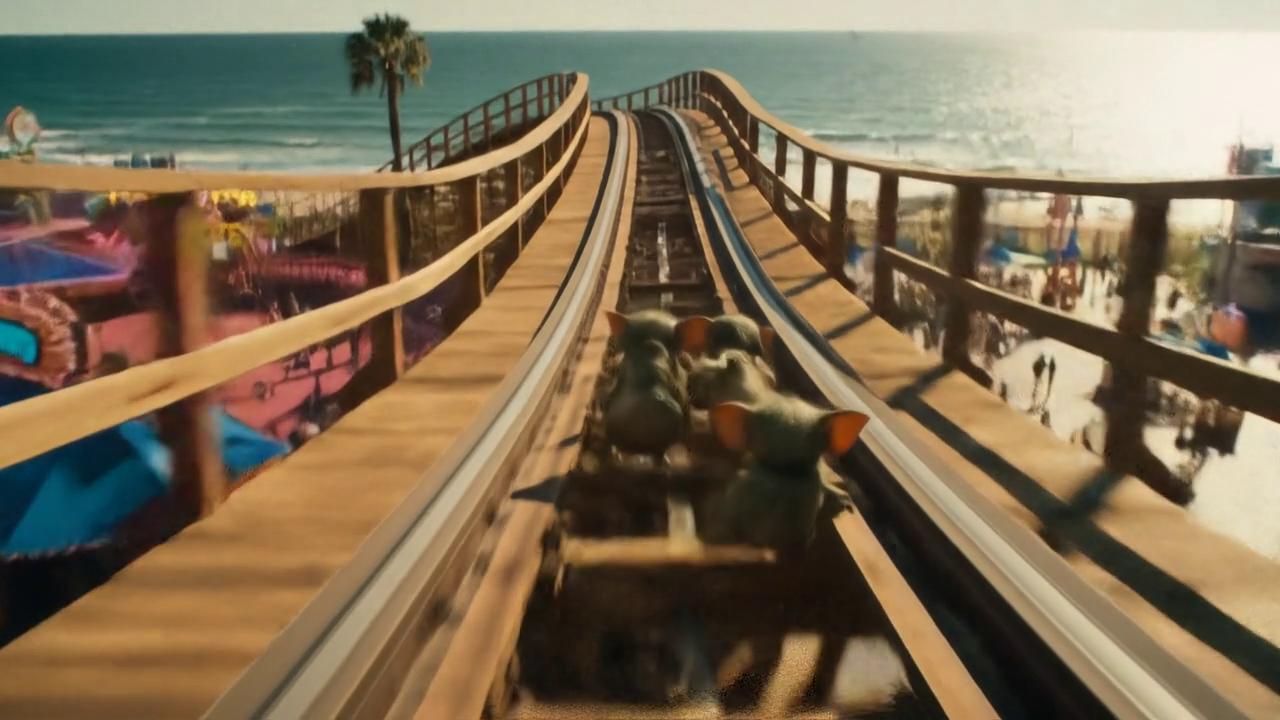}
    \hspace{-2mm}
\includegraphics[width=0.2\textwidth]{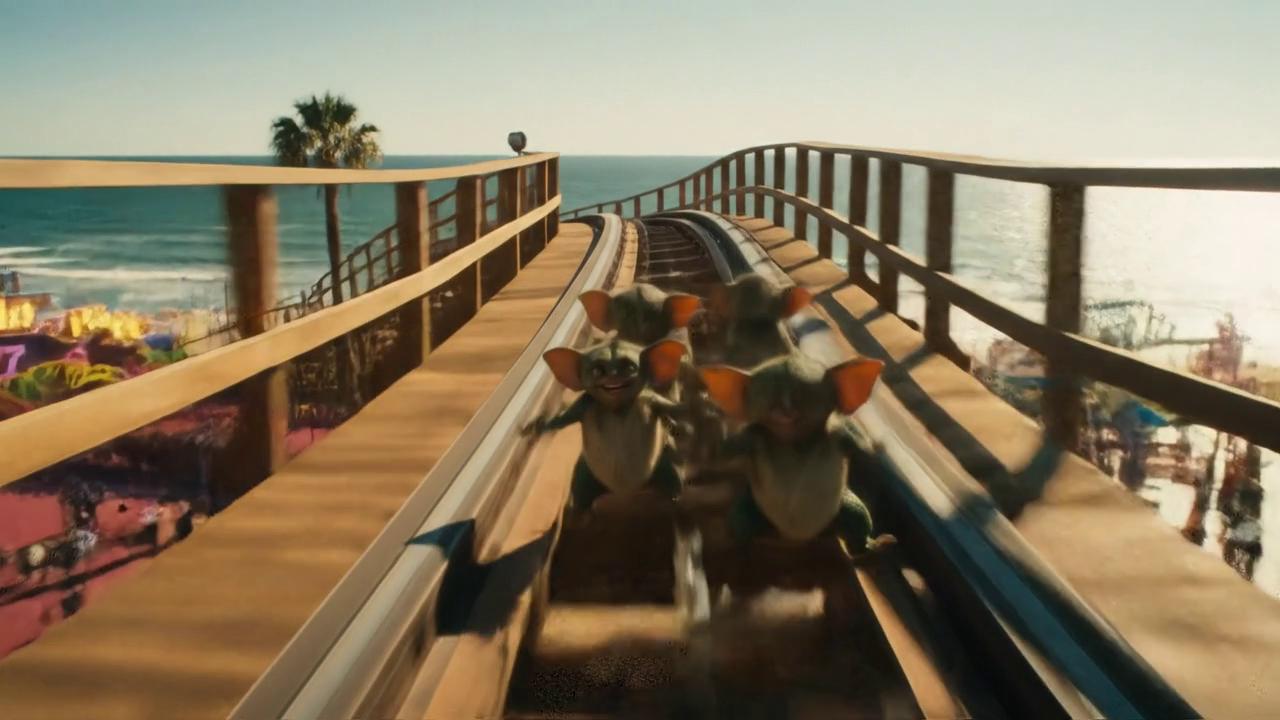}
\begin{minipage}[t]{0.19\textwidth}
    \vspace{-17mm}
\scriptsize{DP generates video with \textcolor{red}{gremlins moving backward fast without wooden rollercoaster}, which is physically non-sense.}
    \end{minipage}
\end{minipage}
        
\begin{minipage}[t]{\textwidth}
    \centering
\includegraphics[width=0.2\textwidth]{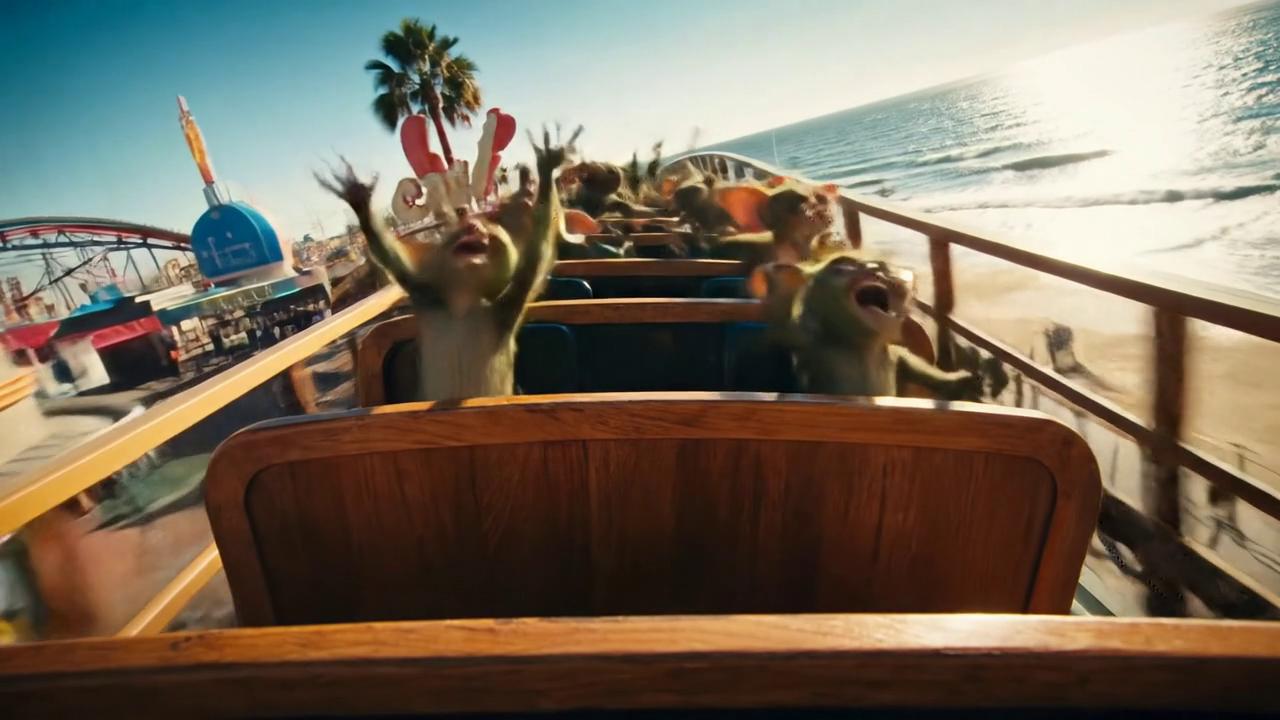}
\hspace{-1mm}\includegraphics[width=0.2\textwidth]{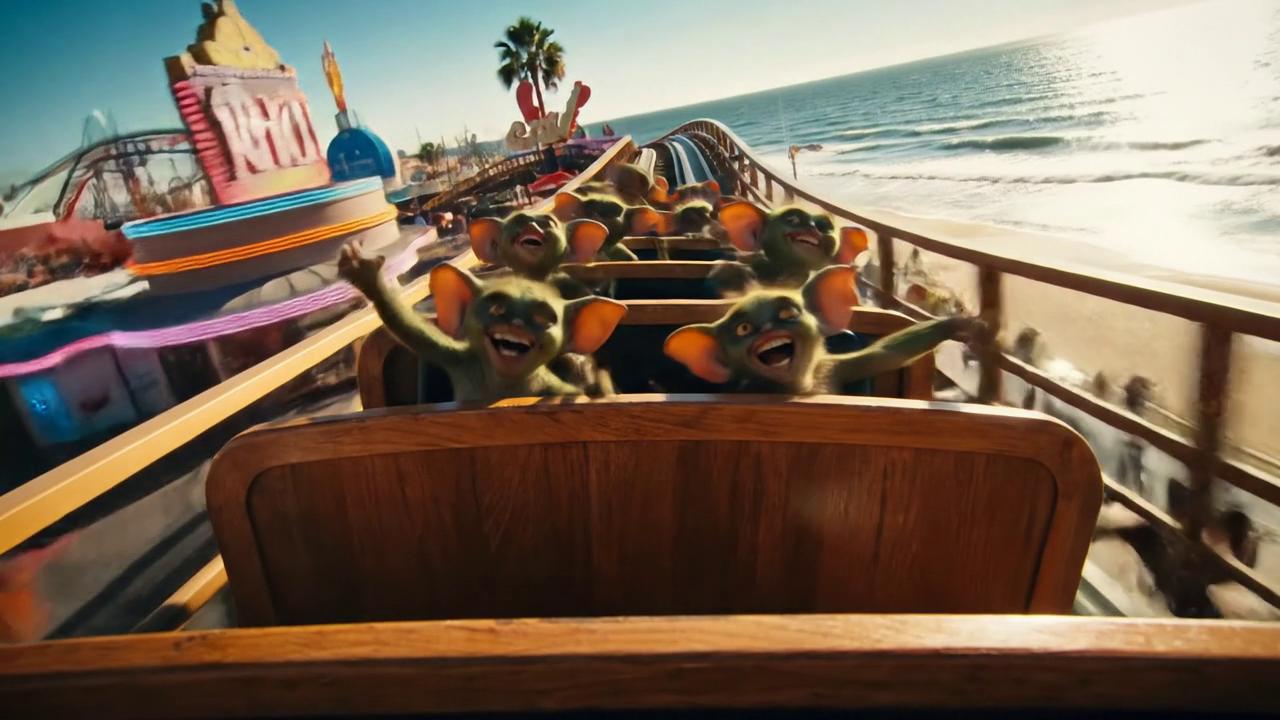}
\hspace{-1mm}\includegraphics[width=0.2\textwidth]{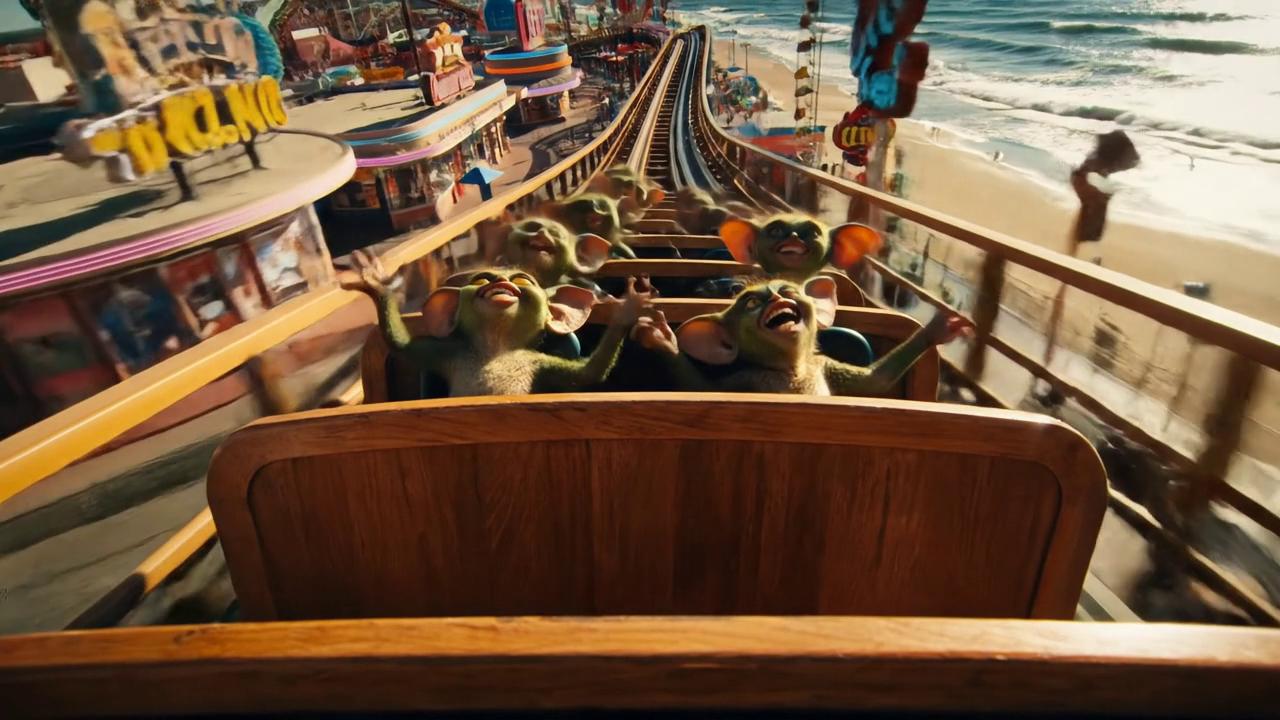}
\hspace{-1mm}\includegraphics[width=0.2\textwidth]{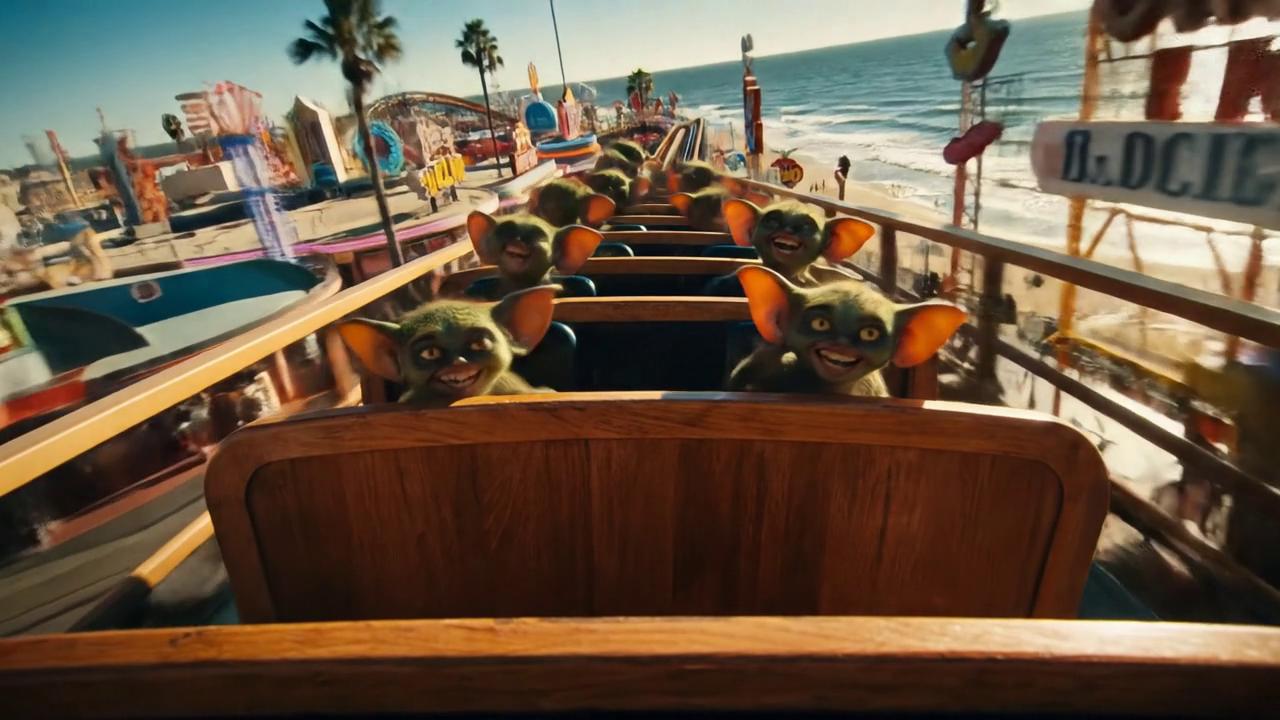}
\begin{minipage}[t]{0.19\textwidth}
\vspace{-17mm}
\scriptsize{\model{} {fixes the issues} with better visual fidelity}: gremlines are moving forward and camera is backward.
    \end{minipage}
\end{minipage}
\footnotesize{ \textbf{Prompt}: A rapid tracking shot of small, big-eared \textbf{gremlins on a wooden rollercoaster} in a midcentury theme park\dots
}
\\
\begin{minipage}[t]{\textwidth}        \includegraphics[width=0.2\textwidth]{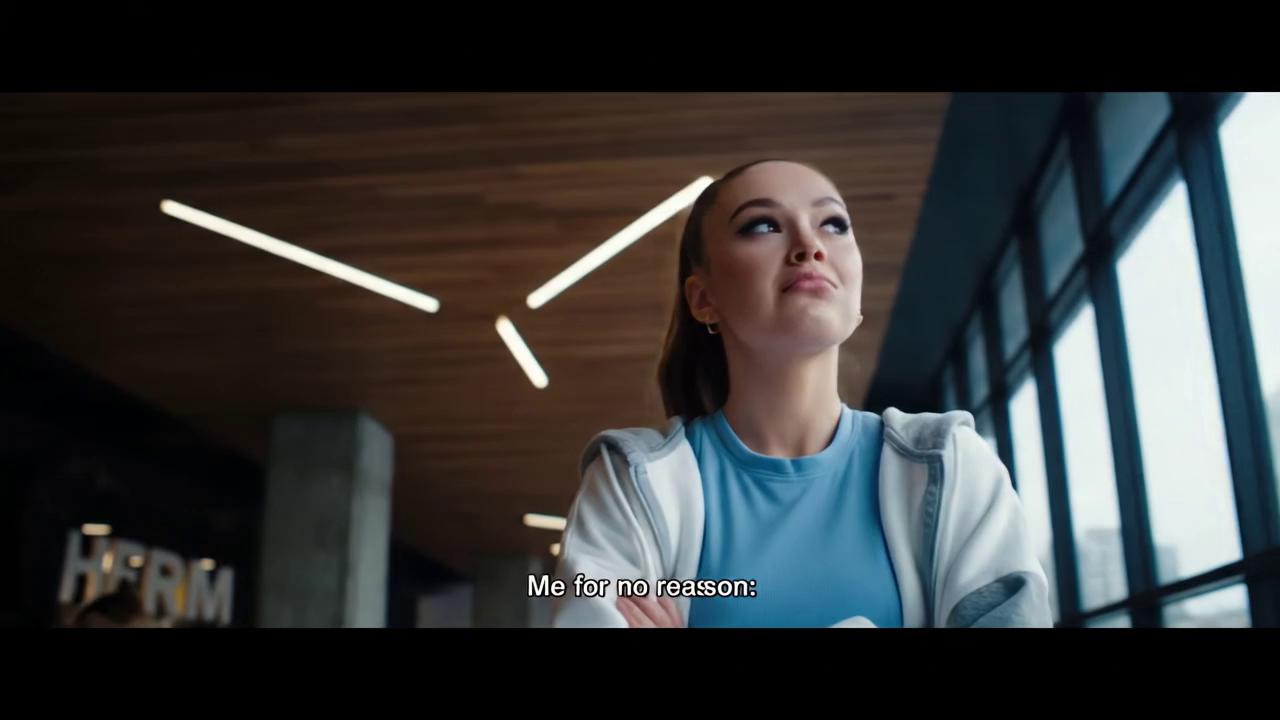}
\hspace{-2mm}
\includegraphics[width=0.2\textwidth]{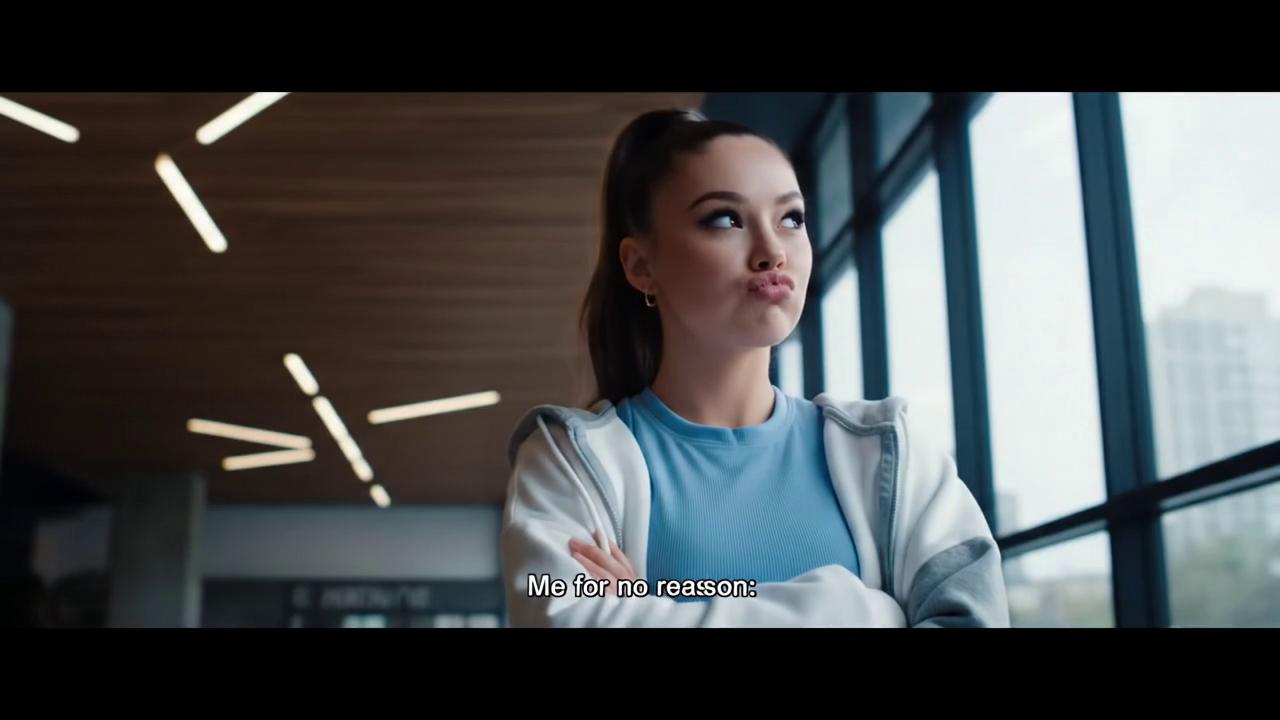}
\hspace{-2mm}
\includegraphics[width=0.2\textwidth]{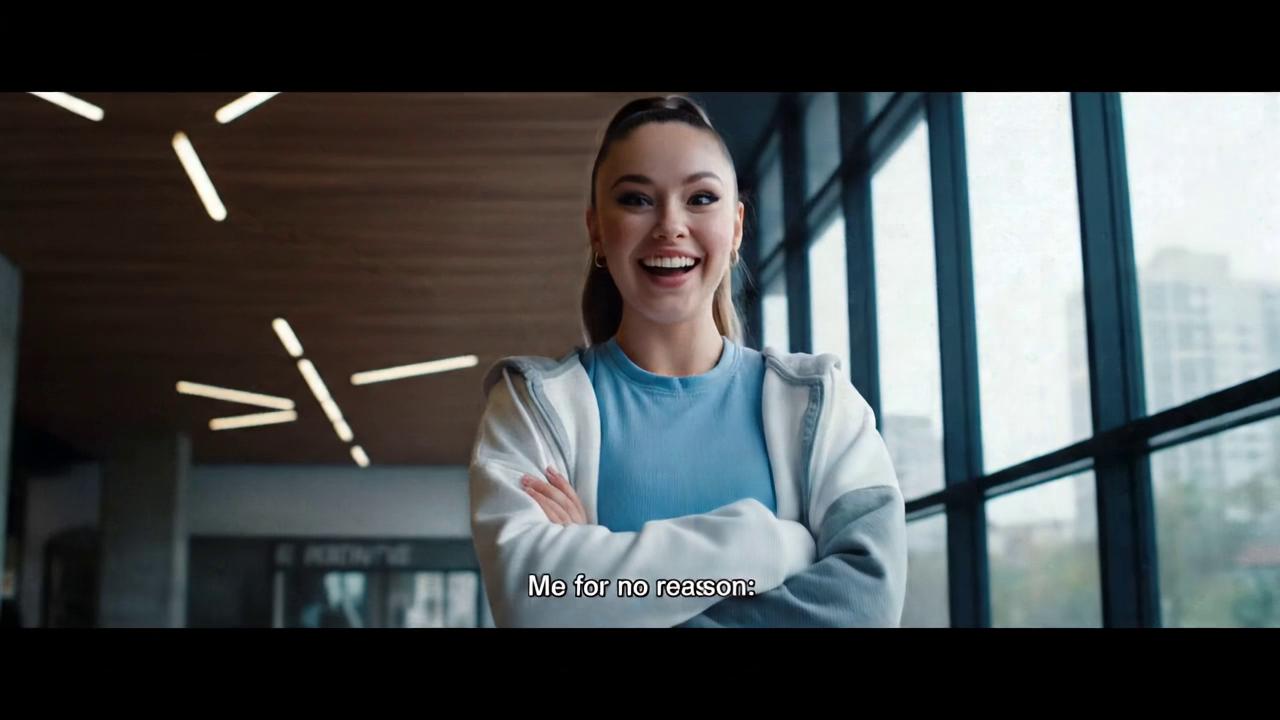}
\hspace{-2mm}    \includegraphics[width=0.2\textwidth]{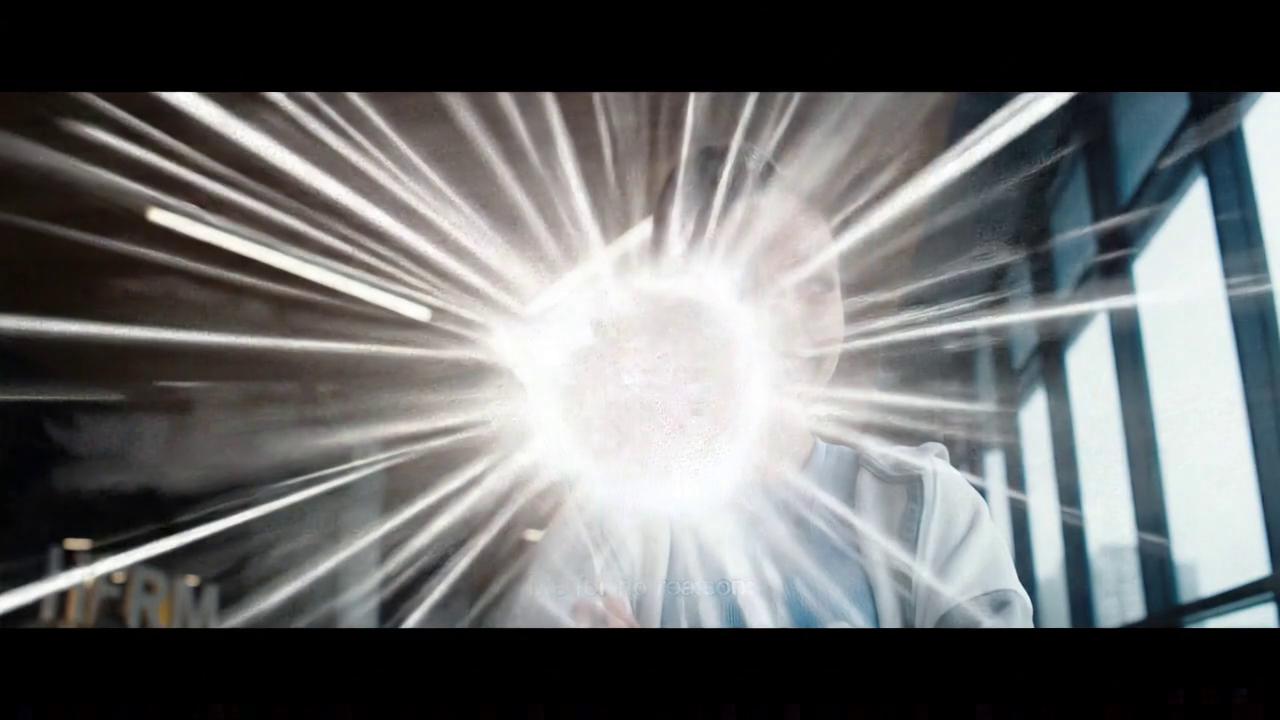}
\begin{minipage}[t]{0.19\textwidth}
\vspace{-17mm}
\scriptsize{DP \textcolor{red}{fails to cut between the character's face shot and the interior ceiling shot}, nor repeat this transition.}
\end{minipage}
\end{minipage}
    
\begin{minipage}[t]{\textwidth}
    \centering
\hspace{-1mm}\includegraphics[width=0.2\textwidth]{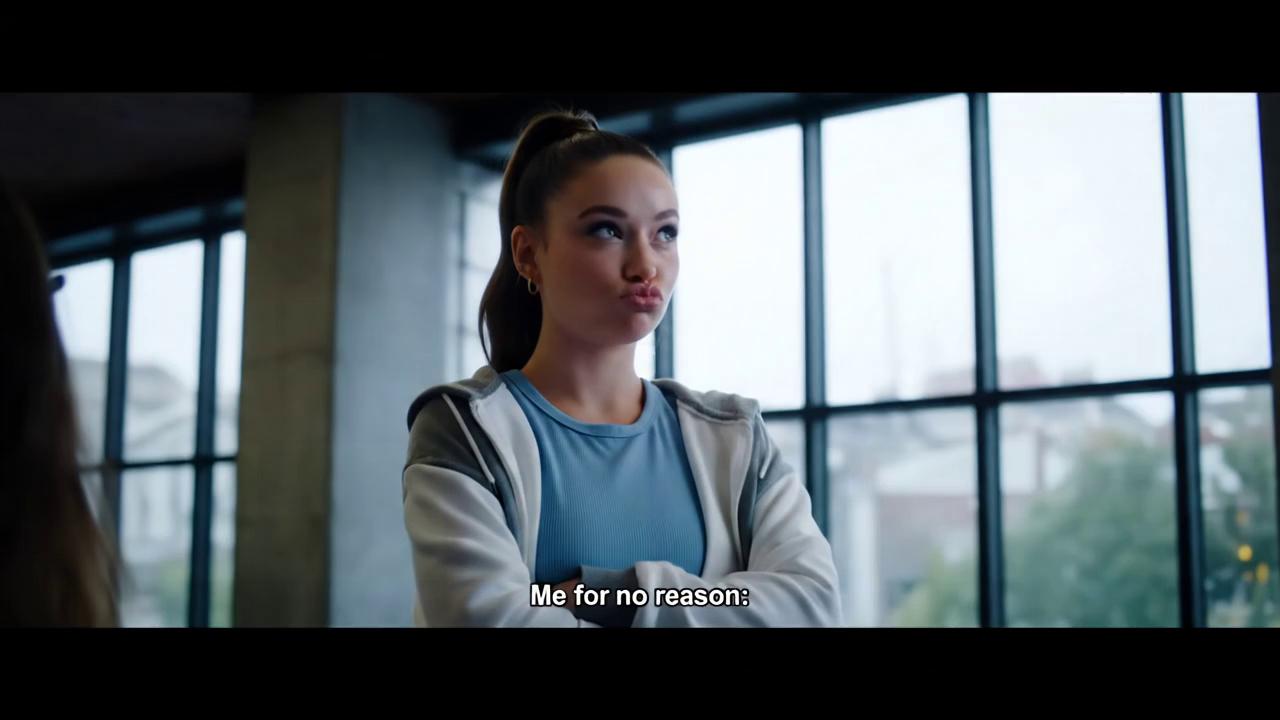}
\hspace{-1mm}\includegraphics[width=0.2\textwidth]{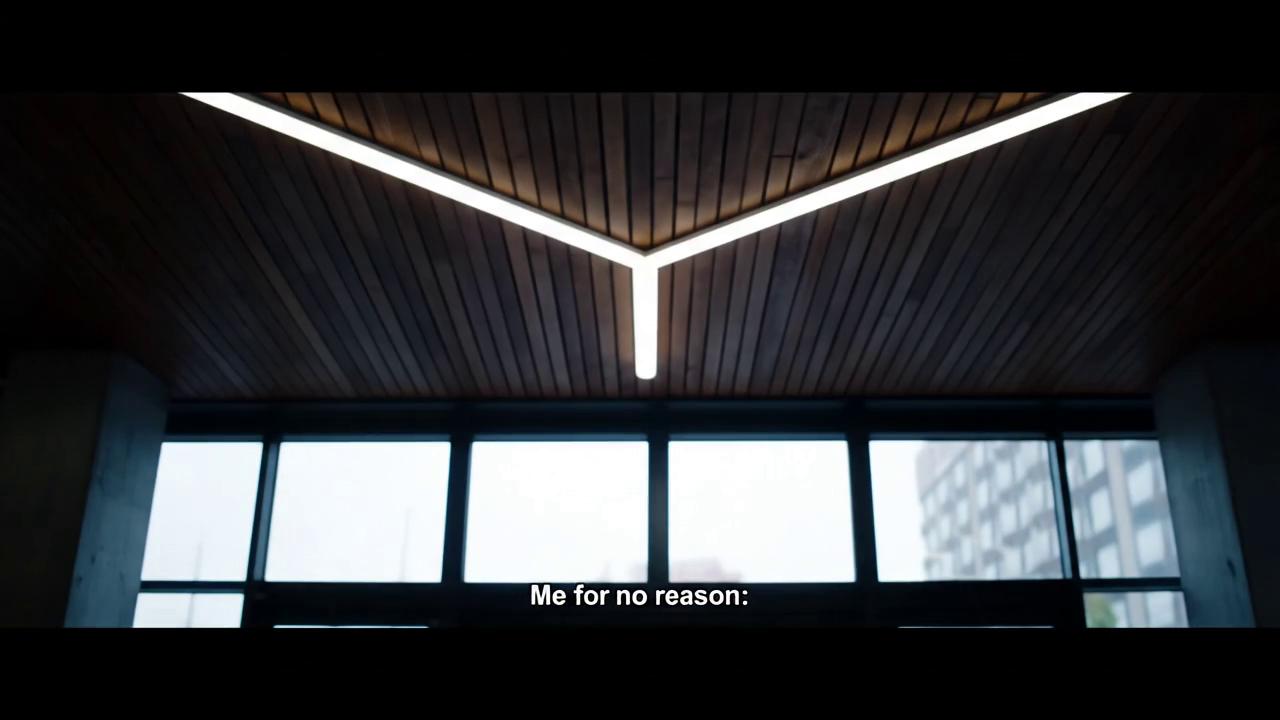}
\hspace{-1mm}\includegraphics[width=0.2\textwidth]{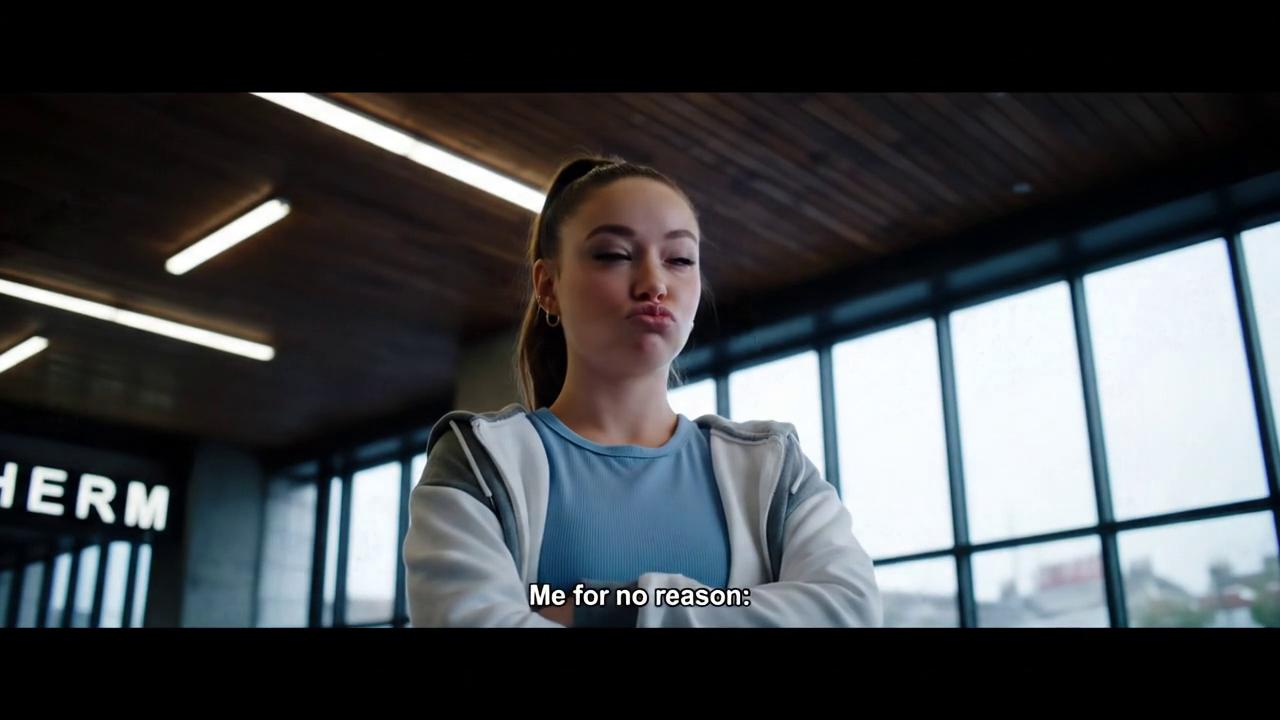}
\hspace{-1mm}\includegraphics[width=0.2\textwidth]{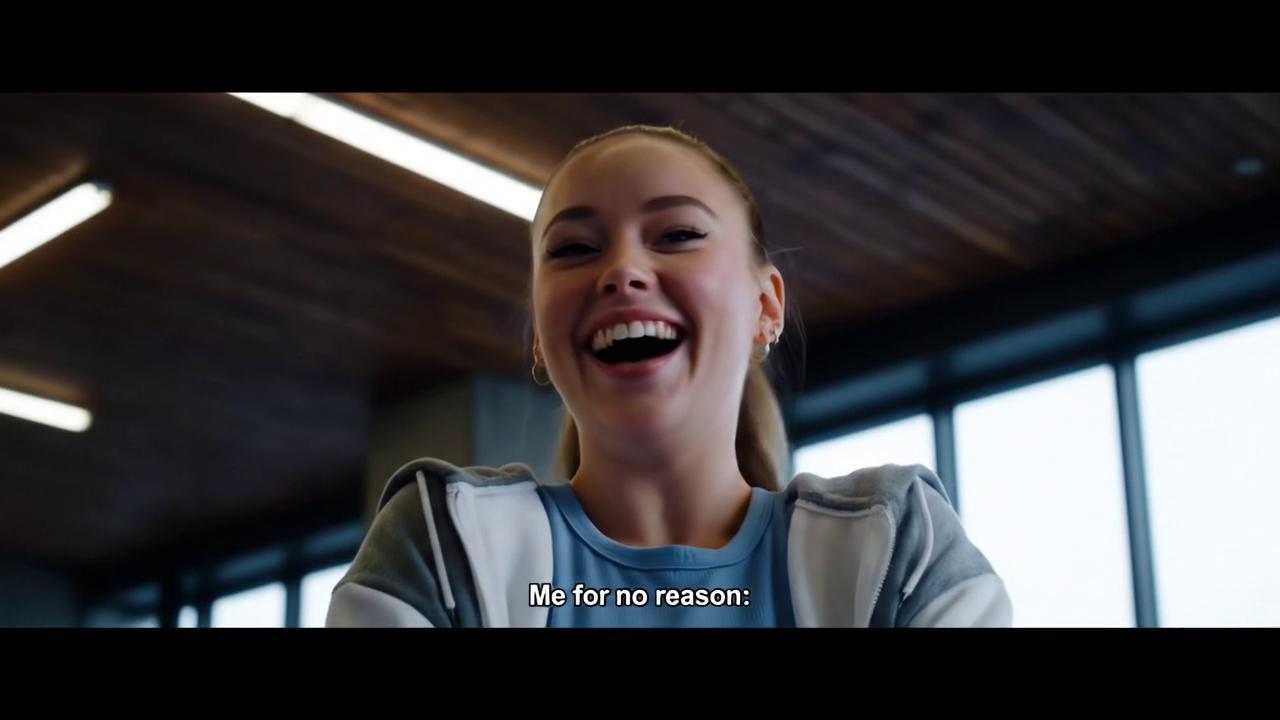}
\begin{minipage}[t]{0.19\textwidth}
    \vspace{-17mm}
    \scriptsize{\model{} {fixes this} with an even {better camera focus and more natural character's expression}.}
    \end{minipage}
\end{minipage}

\footnotesize{\textbf{Prompt}: A short, humorous video depicting a young woman's mood rapidly shifting from bored and slightly annoyed to overtly joyful, triggered by a sudden and dramatic change in background music\dots}

\caption{Comparison of Direct Prompting (DP) and \model{} with Veo 3. Top rows show Veo 3 failing to follow human instructions and generating unrealistic gremlin motions. Bottom rows highlight its inability to produce coherent multi-scene videos under DP. In contrast, \model{} corrects these issues with more realistic motions and smoother visual transitions.}
   \label{fig:if-examples}
\end{figure}

Through human evaluation, we identify key behaviors of \model{} that contribute to its significant improvements in video generation quality by refining prompts. First, \textbf{\model{} enhances video quality by improving prompt fidelity without inducing content drift}. As shown in \Cref{fig:if-examples}, compared to DP, the content of the generated videos remains faithful to the original prompt while achieving substantial quality gains. This improvement can be attributed to two key factors: the explicit constraints applied during the planning process, and  the critique and prompt optimization mechanisms that enforce the text-video alignment. Second, \textbf{\model{} significantly improves instruction-following in SOTA video generation models}. As seen in \Cref{fig:if-examples}, DP often fails to meet prompt specifications, while \model{} successfully corrects such failures. This improvement stems from \model{}'s strict enforcement of text-video alignment during video selection and its use of feedback that evaluates alignment and contextual relevance. Finally, \textbf{\model{} reduces physical, visual, and audio hallucinations}. While models like Veo 3 often produce videos with abrupt object changes, implausible motions, and unsolicited audio or text, \model{} mitigates these issues through constraint-guided selection and strict penalties for violations (\Cref{alg:mavpo}-Step 2).

\subsection{Video Examples} We invite audiences to visit our project page and \Cref{fig:teaser} for more video examples.

\subsection{MMAC Examples}

\begin{figure}[h!]
\begin{minipage}[t]{\textwidth}
\centering
\includegraphics[width=0.2\textwidth]{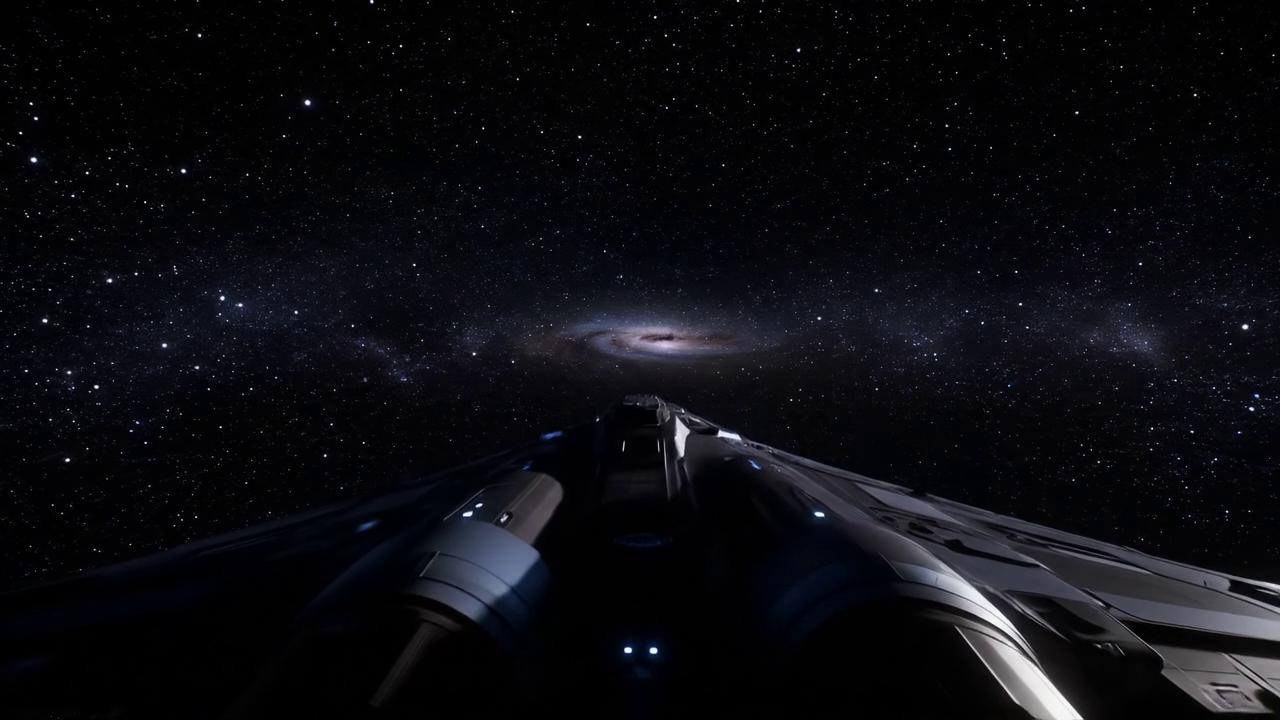}
\hspace{-2mm}
\includegraphics[width=0.2\textwidth]{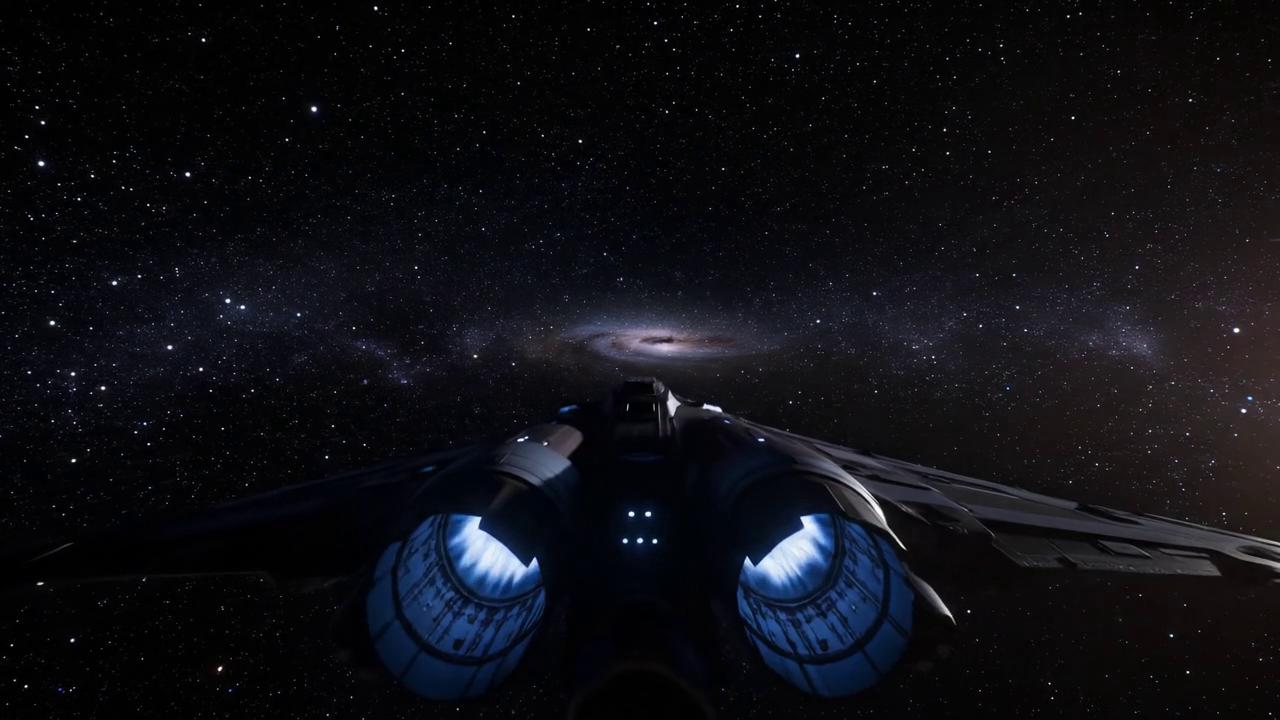}
\hspace{-2mm}
\includegraphics[width=0.2\textwidth]{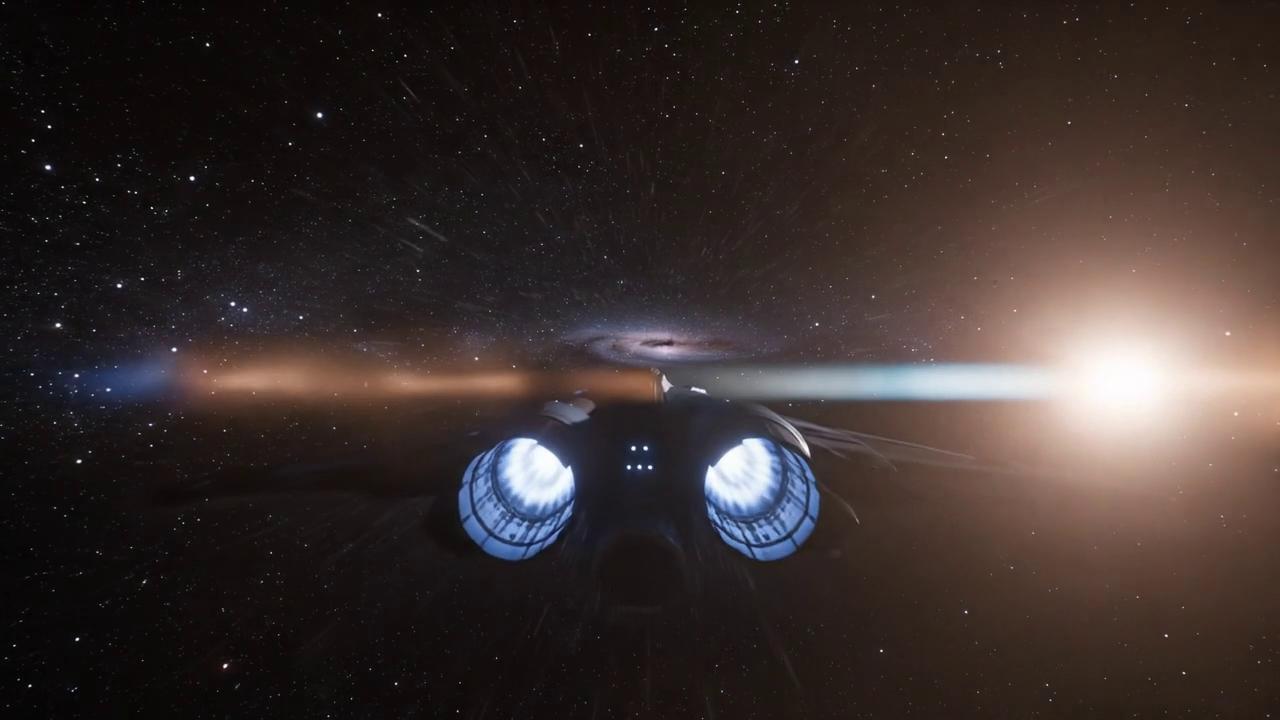}
\hspace{-2mm}
\includegraphics[width=0.2\textwidth]{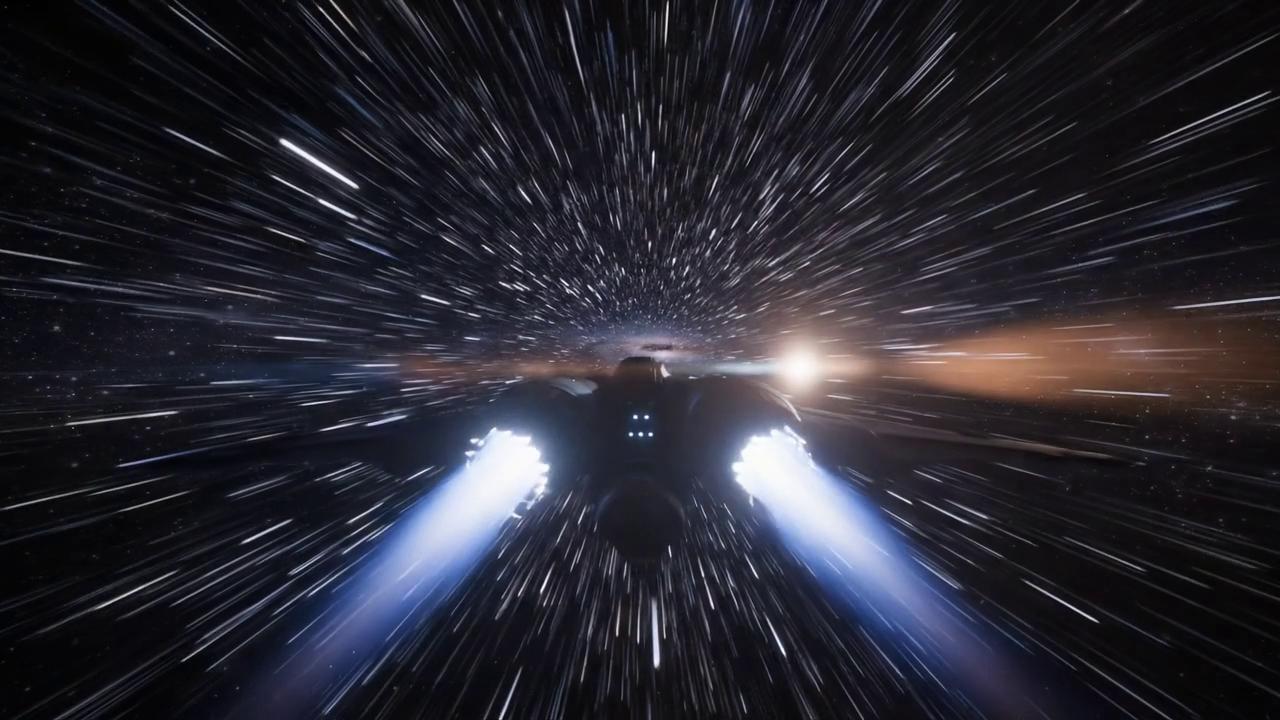}
\hspace{-2mm}
\includegraphics[width=0.2\textwidth]{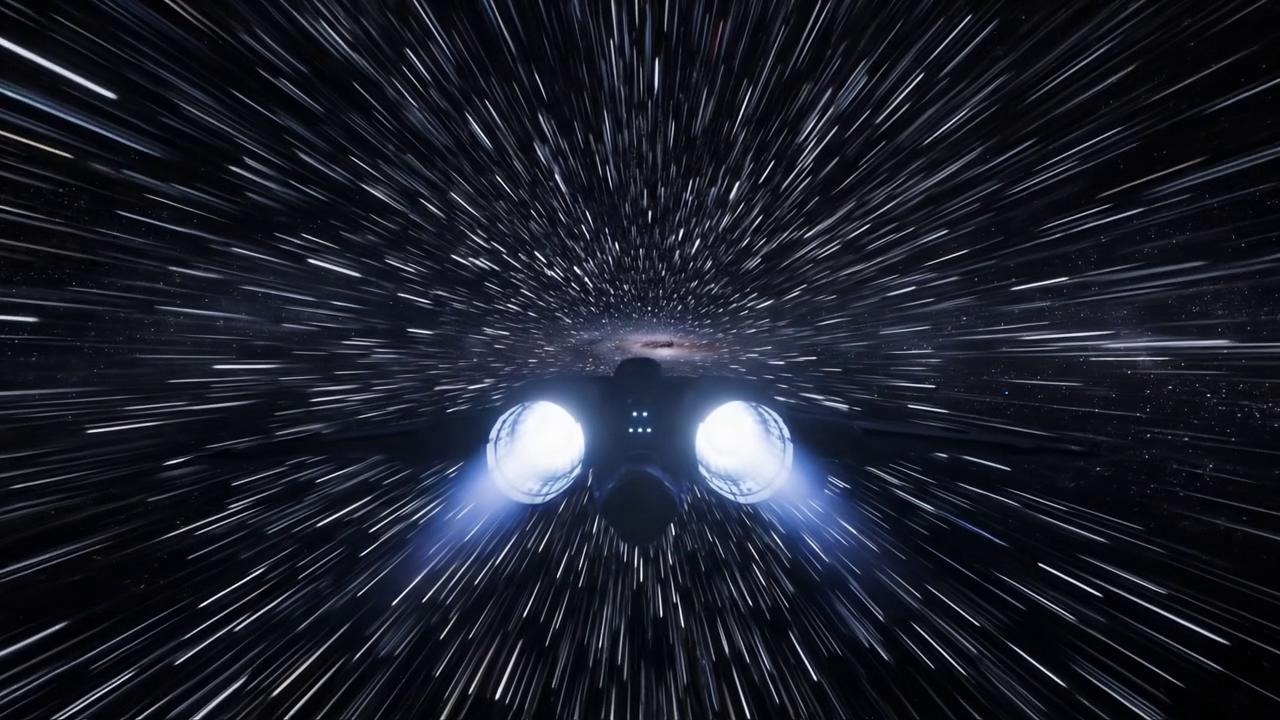}
\end{minipage}
\footnotesize{ \textbf{Prompt}: A spaceship entering hyperdrive, stars streaking past as it accelerates. \\
\textbf{Self-Refine}: The video is a strong visual representation of the prompt. The criticisms are minor and relate more to stylistic choices or common sci-fi tropes rather than a failure to meet the core request. It successfully conveys a spaceship accelerating into a hyperdrive state with stars streaking past.\\
\textbf{\model{} (Visual Fidelity)}: The Normal Judge praises the high technical quality, clarity, sharp details, and consistent sci-fi aesthetic, noting the well-executed lighting and vibrant engine glow. They suggest minor improvements like more dynamic lighting and nuanced atmospheric distortion. The Negative Judge, however, critically points out \textcolor{red}{uninspired lighting} on the spaceship, and \textcolor{red}{static background galaxy, and overly uniform, artificial-looking streaking stars} lacking variation or parallax. They also mention the abruptness of transitions affecting fidelity. Upon review, the Negative Judge's assessment carries more weight. While the video is technically clean, the lighting on the ship does appear quite uniform, lacking the dynamic interplay of light and shadow expected in space\dots
}\\

\begin{minipage}[t]{\textwidth}
\includegraphics[width=0.2\textwidth]{images/goodfeedback_2_6036516217665407039/goodfeedback_2_6036516217665407039_2.jpg}
\hspace{-2mm}
\includegraphics[width=0.2\textwidth]{images/goodfeedback_2_6036516217665407039/goodfeedback_2_6036516217665407039_32.jpg}
    \hspace{-2mm}
\includegraphics[width=0.2\textwidth]{images/goodfeedback_2_6036516217665407039/goodfeedback_2_6036516217665407039_57.jpg}
    \hspace{-2mm}
\includegraphics[width=0.2\textwidth]{images/goodfeedback_2_6036516217665407039/goodfeedback_2_6036516217665407039_77.jpg}
\hspace{-2mm}
\includegraphics[width=0.2\textwidth]{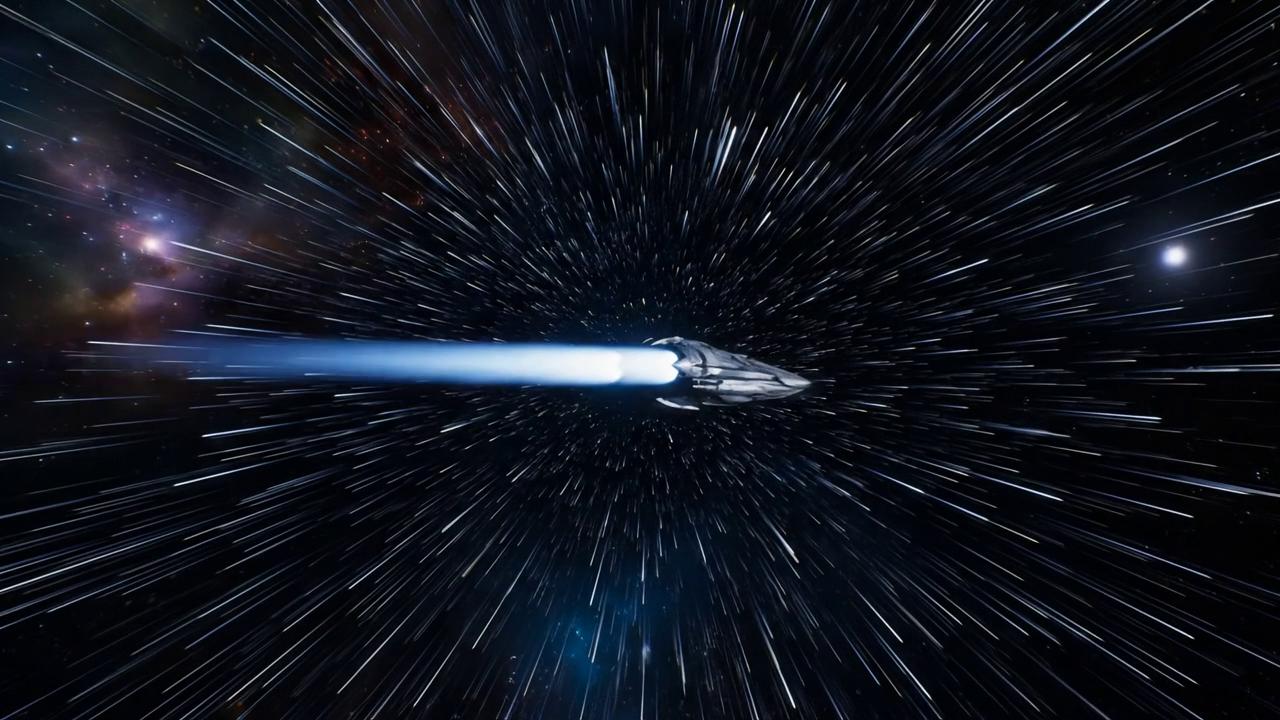}
\end{minipage}
\footnotesize{
\textbf{Self-Refine}: Overall, the generated video is **highly successful** in fulfilling the prompt. It accurately and visually compellingly depicts a spaceship entering hyperdrive with stars streaking past as it accelerates. There are no significant "wrongs" or misinterpretations of the prompt. The minor points above are more about stylistic choices or slight enhancements rather than fundamental errors.\\
\textbf{\model{} (Motions and Dynamics)}: While the Normal Judge praises the smoothness of the ship's motion and dynamic star streaking, the Negative Judge correctly identifies a major directional flaw: \textcolor{red}{the spaceship moves vertically, which conflicts with viewer expectations of horizontal acceleration}. Additionally, the Negative Judge points out \textcolor{red}{the lack of micro-dynamics (e.g., rotational drift, buildup phases) and unrealistic exhaust behavior, which diminish the believability of motion}. These omissions outweigh the surface-level smoothness. }   
\begin{minipage}[t]{\textwidth}
\includegraphics[width=0.2\textwidth]{images/goodfeedback_77_vpo_16374353314150394085/goodfeedback_77_vpo_16374353314150394085_1.jpg}
\hspace{-2mm}
\includegraphics[width=0.2\textwidth]{images/goodfeedback_77_vpo_16374353314150394085/goodfeedback_77_vpo_16374353314150394085_45.jpg}
\hspace{-2mm}
\includegraphics[width=0.2\textwidth]{images/goodfeedback_77_vpo_16374353314150394085/goodfeedback_77_vpo_16374353314150394085_77.jpg}
\hspace{-2mm}
\includegraphics[width=0.2\textwidth]{images/goodfeedback_77_vpo_16374353314150394085/goodfeedback_77_vpo_16374353314150394085_159.jpg}
\hspace{-2mm}\includegraphics[width=0.2\textwidth]{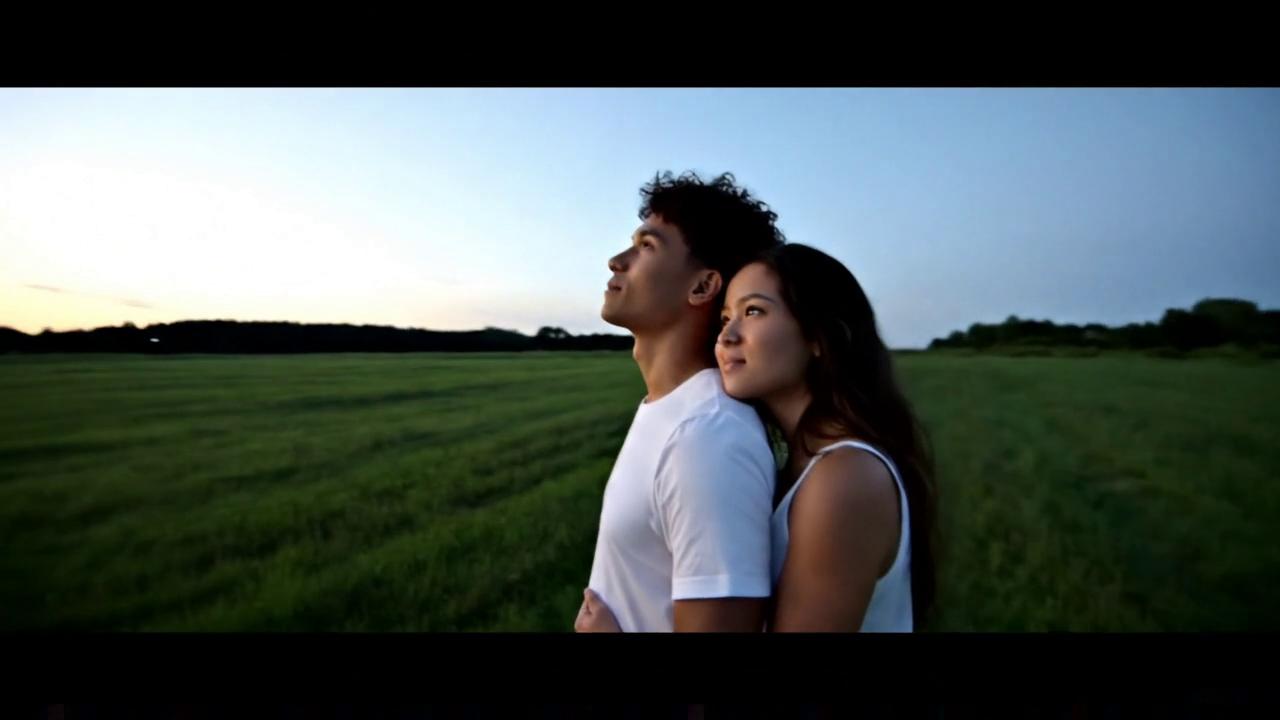}
\end{minipage}
        
\footnotesize{\textbf{Prompt}: The couple runs hand in hand to release a sky lantern, then watches it drift upward into the night sky, carried by the wind with the stars shining above.\\
\textbf{Self-Refine}: The video successfully portrays the actions of the couple running and releasing a lantern. However, it **fails significantly on the atmospheric detail of the "night sky with stars shining above."** The absence of stars is a major oversight that detracts from the prompt's intended mood and visual.\\
\textbf{\model{} (Contextual Suitability)}: The scene struggles significantly with internal logical consistency, particularly regarding its temporal and environmental progression. An abrupt and jarring shift from a bright, late-afternoon setting to a deep, artificial blue night sky without any visual cues for time passing severely undermines immersion. This featureless 'night sky' further detracts from plausibility, failing to represent a natural environment. While individual actions are suitable, their placement within such an inconsistent context creates a sense of unreality. \\
\textbf{\model{} (Visual Characters)}: While the two characters are central and necessary to the narrative, their portrayal lacks emotional depth and genuine interaction. Their expressions, particularly during the lantern release, appear somewhat static and posed, missing an opportunity to convey a deeper sense of wonder, intimacy, or shared experience. This underdevelopment makes them feel more like subjects performing actions rather than individuals experiencing a meaningful moment, limiting the scene's emotional resonance.\par

\caption{\model{} delivers high-level feedback on visual fidelity, contextual consistency, and emotional expression, surfacing nuanced flaws (e.g., static lighting, implausible scene transitions, and lack of character depth) that conventional LLM critiques often miss but are readily noticed by human judges.}
} \label{fig:if-high-level-feedback}
\end{figure}

\Cref{fig:if-high-level-feedback} shows more examples of critiques from \model{} compared to Visual Self-Refine.

\subsection{Deep-Thinking Prompting Agent}
\label{appdx:deep-thinking-prompting-example}

Recall the video from \Cref{fig:good-modifications}:

\begin{tcolorbox}[colback=white!95!gray, colframe=black!75, 
title=Deep-Thinking Procedure Answers,
fonttitle=\bfseries,
sharp corners,
enhanced,
breakable,
left=1mm,
right=1mm,
top=1mm,
bottom=1mm,
boxrule=0.5pt,
arc=0pt]

{\footnotesize
Deep-Thinking Procedure Answers:

\textbf{1.  Review the Issues}

    Upon a thorough review of the provided feedback, several areas for improvement have been identified, indicated by qualitative feedback suggesting less-than-optimal performance, even if a numerical score isn't explicitly given as "less than 8." The prompt asks to identify issues with scores less than 8, but since numerical scores aren't provided, I will interpret "major issues" as those with detailed negative qualitative feedback.

    The primary issues are:

    *   **Visual Fidelity:** The feedback highlights "visible reflections of the camera setup in the subject's sunglasses." This breaks immersion and indicates a lack of control over subtle visual details. Furthermore, the "prominent size" of the text overlays and their "abrupt appearance" are noted as detracting from a smooth visual experience. While legible, their visual impact is not ideal.
    *   **Motions and Dynamics:** This category specifically calls out the "instantaneous appearance and disappearance of the text overlays" as abrupt, reiterating the point from visual fidelity. This indicates a lack of fluid animation for on-screen graphics, which impacts the overall polish and dynamic flow of the video.
    *   **Audio Quality Cohesion:** A significant concern is the "noticeable wind noise throughout the outdoor segment," which compromises the professionalism and clarity of the audio. Additionally, the feedback points out a "lack of depth due to the absence of a subtle, natural ambient bed," making the scene feel "somewhat sterile" despite clear dialogue. This suggests the ambient sound generation was either insufficient or poorly integrated.
    *   **Timeline and Transition:** While the cut to the outro screen is described as "clean and intentional," the feedback explicitly states it is "not a 'fluid' transition (e.g., a dissolve or fade)." It suggests that "for absolute perfection in 'smoothness,' a subtle transition effect could have been employed." Although deemed "common and acceptable" for short-form content, it's highlighted as an area where "smoothness" could be improved, indicating it's not achieving the highest possible quality in this specific aspect.

    These issues collectively point to a need for more granular control over visual effects, on-screen text animation, environmental audio generation, and scene transitions to elevate the video's overall production quality and polish. Addressing these will enhance viewer immersion and the professional feel of the content.

\textbf{2.  **Define the Objectives**}

    The overarching objective of this video, as defined by the User Prompt, is to create a short-form piece of content centered around "Trivia and lighthearted knowledge sharing, with a focus on comedy." The video aims to be "Casual, informative, direct, and slightly humorous" in tone. The specific structure involves a man asking and answering a trivia question outdoors, followed by a branded outro screen with a call to action. The video has a strict length constraint of 8 seconds, with a precise breakdown of scene durations (5.5 seconds for the trivia segment and 2.5 seconds for the outro).

    The prompt implicitly sets high expectations for visual and audio fidelity, given the detailed descriptions of character appearance, actions, visual environment, camera work, and sound design. Success criteria include clear dialogue, appropriate visual framing, a natural outdoor setting, and a clean, effective call-to-action screen. The target audience appears to be general viewers interested in quick, engaging trivia content, likely on social media platforms where short, polished videos perform well. The key message is to share a piece of trivia and encourage viewer engagement through subscription. The prompt's specificity across various elements (e.g., "red baseball cap," "black sunglasses with blue reflective lenses," "purple hoodie," "black rectangular text box with white text") indicates a desire for precise execution and a high degree of alignment with the provided specifications. The feedback further reinforces this by praising "exceptional alignment" and "excellent temporal consistency," suggesting that the core objective of faithful reproduction was largely met, but refinement in specific areas is still needed for optimal quality.

\textbf{3.  **Identify Model Limitations, Given the Video Prompt**}

    Analyzing the identified major issues in conjunction with the Video Prompt, several points suggest potential limitations of the underlying video generation model when interpreting less explicit instructions or handling complex real-world phenomena.

    *   **Reflections in Sunglasses:** The presence of "visible reflections of the camera setup in the subject's sunglasses" points to a challenge in rendering complex optical phenomena accurately without explicit guidance. Simulating realistic reflections, especially on curved, reflective surfaces like sunglasses, requires sophisticated ray tracing or similar techniques. Without a specific instruction to "minimize reflections" or "ensure no camera reflections are visible," the model might default to a physically plausible but undesirable outcome. This is a common limitation in generative models that aim for realism but may not prioritize specific aesthetic refinements unless prompted.
    *   **Abrupt Text Overlays (Appearance/Disappearance Size):** The "instantaneous appearance and disappearance" of text overlays and their "prominent size" suggest the model's default behavior for on-screen graphics. When the prompt simply states "Text overlay appears" or "Text overlay changes," the model interprets this as a binary state change (on/off) rather than a smooth animation (e.g., fade, slide). Similarly, without explicit size or placement parameters, the model might choose a default that is visually impactful but not aesthetically subtle. This indicates a limitation in the model's ability to infer desired animation styles or optimal visual hierarchy for text without explicit instructions.
    *   **Wind Noise and Lack of Ambient Depth:** The "noticeable wind noise" and absence of a "subtle, natural ambient bed" highlight a limitation in generating nuanced and clean environmental audio. While the prompt asks for "Faint ambient street noise," generating specific, clean ambient sounds while simultaneously filtering out undesirable elements like wind noise (which is often present in outdoor recordings) is a complex audio engineering task. The model might struggle to differentiate between desired "outdoor sounds" and distracting "noise," or to layer sounds effectively to create depth without explicit instructions on sound mixing, noise reduction, or specific ambient sound profiles.
    *   **Non-Fluid Transition:** The "immediate cut" from the live-action scene to the outro, despite the feedback suggesting a "subtle transition effect" for "absolute perfection," indicates that the model defaults to a hard cut when no specific transition type is mentioned. This is a common behavior for generative models; they typically require explicit instructions for dissolves, fades, or other cinematic transitions, as a hard cut is the simplest and most direct way to move between scenes.

    These issues are not necessarily "failures" of the model but rather areas where its default interpretations or capabilities fall short of achieving a highly polished, professional output without more precise and detailed prompting.

\textbf{4.  **Identify Video Prompt Issues, Given the Model**}

    Considering the identified model limitations and the feedback, the Video Prompt exhibits several areas where its vagueness, lack of specificity, or missing information contribute directly to the observed issues.

    *   **Vague Text Overlay Instructions:** The prompt states "Text overlay appears" and "Text overlay changes." This is a vague instruction regarding the *manner* of appearance. It doesn't specify any animation (e.g., fade-in, slide-up, dissolve). This vagueness leads the model to default to an abrupt, instantaneous appearance and disappearance, as noted in the feedback for both "visual fidelity" and "motions and dynamics." Furthermore, the prompt does not specify the *size* or *prominence* of the text overlays, leading to them occupying a "significant portion of the screen," which was deemed too prominent.
    *   **Insufficient Audio Detail:** The instruction "Faint ambient street noise, possibly distant traffic or general outdoor sounds, is audible in the background" is too broad. It doesn't explicitly request noise reduction (e.g., for wind) or specify the *quality* or *composition* of the ambient soundscape (e.g., "subtle, consistent ambient bed"). This lack of detail allows the model to generate "noticeable wind noise" and a sterile soundscape lacking depth, as it interprets "outdoor sounds" without the necessary refinement or filtering instructions.
    *   **Missing Transition Specification:** The prompt defines the timestamps for the two scenes but provides no instruction whatsoever regarding the *type* of transition between the live-action segment and the outro screen. This missing information leads the model to implement a default hard cut, which, while functional, is not considered "fluid" or "perfectly smooth" according to the feedback. To achieve a more polished transition, a specific instruction (e.g., "subtle cross-dissolve") is required.
    *   **Lack of Aesthetic Refinement Instructions:** The prompt does not include any instructions related to minimizing undesirable visual artifacts like reflections in eyewear. While it describes the character's attire, it doesn't anticipate or address potential issues with reflective surfaces. This missing aesthetic constraint allows the model to generate "visible reflections of the camera setup," which detract from immersion.

    In essence, the prompt, while highly detailed in many aspects, lacks the necessary specificity in areas related to visual animation, audio refinement, and subtle aesthetic controls. These omissions force the model to make assumptions or default to basic implementations that do not meet the desired level of polish and fluidity.

\textbf{5.  **Propose Targeted Revisions to the Video Prompt**}

    Based on the comprehensive analysis of the issues, model limitations, and prompt shortcomings, the following targeted revisions are proposed to enhance the Video Prompt's effectiveness and guide the model toward generating a higher-quality, more polished video:

    1.  **For Visual Fidelity (Reflections in Sunglasses):**
        *   **Modification Action:** Add a specific instruction within the character description or visual environment section to minimize or avoid reflections of the camera setup in the subject's sunglasses. This could be phrased as a constraint or a desired aesthetic outcome.
        *   **Rationale:** This directly addresses the feedback regarding "visible reflections of the camera setup," which breaks immersion. By explicitly stating this, the prompt guides the model to prioritize a cleaner visual on reflective surfaces.

    2.  **For Visual Fidelity \& Motions/Dynamics (Text Overlays):**
        *   **Modification Action:** Modify the dialogue section for the text overlays to specify a smooth animation for their appearance and disappearance (e.g., "fade in" or "slide up from bottom"). Also, add a constraint on their size and placement to ensure they are legible but not overly prominent (e.g., "subtle, occupying minimal screen space at the bottom center").
        *   **Rationale:** This addresses the "abrupt appearance" and "instantaneous disappearance" of text, as well as their "prominent size." Specifying animation will create a smoother visual flow, and defining size/placement will ensure they serve their purpose without being distracting.

    3.  **For Audio Quality Cohesion (Wind Noise \& Ambient Sound):**
        *   **Modification Action:** Refine the "sounds" description for the first scene to explicitly request that dialogue be free of wind noise and to specify a subtle, consistent ambient sound bed. For example, "Clear male voice speaking, free of wind noise. A subtle, consistent ambient street soundscape, such as distant traffic or general outdoor sounds, is audible in the background, ensuring depth without distraction."
        *   **Rationale:** This directly tackles the "noticeable wind noise" and the "lack of depth" in the audio. By being more prescriptive about noise reduction and the quality of ambient sound, the prompt guides the model to generate a cleaner and richer audio experience.

    4.  **For Timeline and Transition (Scene Transition):**
        *   **Modification Action:** Add a specific instruction for the transition between the first scene (0-5.5s) and the second scene (5.5-8s). Specify a subtle, fluid transition type, such as a "subtle cross-dissolve" or "quick fade to black then fade in."
        *   **Rationale:** This addresses the feedback that the immediate cut, while acceptable, is not "fluid." By explicitly requesting a subtle transition, the prompt ensures a more polished and seamless flow between the live-action content and the call-to-action screen, enhancing the overall production quality.

    These revisions aim to provide the model with clearer, more specific instructions in areas where its default interpretations fell short of the desired polished outcome, without overly complicating the core content or structure of the video.

\textbf{6.  **Revise the Suggested Modification Actions**}

    Upon reviewing the proposed modifications against all the major issues identified, the suggested actions comprehensively address each point. The revisions are targeted, specific, and aim to provide the model with clearer guidance where the original prompt was vague or lacked detail.

    *   **Reflections in sunglasses:** The action to "Add a specific instruction within the character description or visual environment section to minimize or avoid reflections of the camera setup in the subject's sunglasses" directly addresses this visual fidelity issue.
    *   **Abrupt text overlays and prominent size:** The action to "Modify the dialogue section for the text overlays to specify a smooth animation for their appearance and disappearance (e.g., 'fade in' or 'slide up from bottom'). Also, add a constraint on their size and placement to ensure they are legible but not overly prominent (e.g., 'subtle, occupying minimal screen space at the bottom center')" directly addresses both the motion/dynamics and visual fidelity concerns related to text.
    *   **Wind noise and lack of ambient depth:** The action to "Refine the 'sounds' description for the first scene to explicitly request that dialogue be free of wind noise and to specify a subtle, consistent ambient sound bed. For example, 'Clear male voice speaking, free of wind noise. A subtle, consistent ambient street soundscape, such as distant traffic or general outdoor sounds, is audible in the background, ensuring depth and naturalism without distraction.'" directly addresses the audio quality cohesion issues.
    *   **Non-fluid transition:** The action to "Add a specific instruction for the transition between the first scene (0-5.5s) and the second scene (5.5-8s). Specify a subtle, fluid transition type, such as a 'subtle cross-dissolve' or 'quick fade to black then fade in.'" directly addresses the timeline and transition feedback.

    All major issues are covered. The modifications are actionable and aim to improve the video's polish and professional quality without introducing unnecessary complexity or deviating from the original intent. No further revisions to the suggested modifications are needed as they are precise and comprehensive.

\textbf{```python}

[
    ``Modify the 'characters' or 'visual\_environment' description for the first scene to include a constraint: 'Ensure reflections in the subject's sunglasses do not show camera equipment or distracting elements, maintaining visual immersion.'",
    "Update the 'dialogues' section for the first scene's text overlays: 'Text overlay appears: \"WHICH COMEDIAN IS KNOWN FOR THEIR DEADPAN DELIVERY?\" (Text overlay should smoothly fade in/slide up from the bottom, be legible, and occupy a subtle, non-prominent portion of the screen, ideally centered at the bottom.)' and 'Text overlay changes to: JEFF DYE (Text overlay should smoothly transition or fade out the previous text and fade in/slide up the new text, maintaining the subtle size and placement.)'",
    "Refine the 'sounds' description for the first scene: 'Clear male voice speaking, with dialogue free of noticeable wind noise. A subtle, consistent ambient street soundscape, such as distant traffic or general outdoor sounds, is audible in the background, ensuring depth and naturalism without distraction.'",
    "Add a specific instruction for the transition between the first scene (timestamp '0-5.5') and the second scene (timestamp '5.5-8'): 'The transition from the live-action scene to the outro screen should be a subtle cross-dissolve or a quick, smooth fade to black then fade in, rather than an immediate hard cut.'"
]\textbf{'''}}
\end{tcolorbox}

\end{document}

%% file: commands.tex
\definecolor{ourred}{HTML}{F19C99}
\definecolor{ourblue}{HTML}{7EA6E0}
\definecolor{plotred}{HTML}{f77189}
\definecolor{plotgreen}{HTML}{33b07a}
\definecolor{plotpurple}{HTML}{cc7af4}
\definecolor{plotmagenta}{HTML}{f565cc}
\definecolor{plotazure}{HTML}{38a9c5}

\definecolor{codegreen}{rgb}{0,0.6,0}
\definecolor{codegray}{rgb}{0.5,0.5,0.5}
\definecolor{codepurple}{rgb}{0.58,0,0.82}
\definecolor{backcolour}{rgb}{0.95,0.95,0.92}

\definecolor{Gray}{gray}{0.92}

\definecolor{stage1}{HTML}{34A853}
\definecolor{stage2}{HTML}{A680B8}
\definecolor{stage3}{HTML}{009999}

\usepackage{listings}

\lstdefinestyle{mystyle}{
    backgroundcolor=\color{backcolour},   
    commentstyle=\color{codegreen},
    keywordstyle=\color{magenta},
    numberstyle=\tiny\color{codegray},
    stringstyle=\color{codepurple},
    basicstyle=\ttfamily\scriptsize,
    breakatwhitespace=false,         
    breaklines=true,                 
    captionpos=b,                    
    keepspaces=true,                 
    numbersep=5pt,                  
    showspaces=false,                
    showstringspaces=false,
    showtabs=false,                  
    tabsize=2,
    frame=none,
    aboveskip=1pt,
    belowskip=1pt,
}
\usepackage[most]{tcolorbox}
\tcbset{enhanced,
boxrule=0pt,frame hidden,
borderline west={4pt}{0pt}{green!75!black},
top=0pt,bottom=0pt,
colback=green!10!white,
sharp corners}

\usepackage{colortbl}
\usepackage{float}

\usepackage{multirow}

%% file: math_commands.tex
\usepackage{amsmath,amsfonts,bm}

\def\eqref#1{equation~\ref{#1}}

\def\1{\bm{1}}

\DeclareMathAlphabet{\mathsfit}{\encodingdefault}{\sfdefault}{m}{sl}
\SetMathAlphabet{\mathsfit}{bold}{\encodingdefault}{\sfdefault}{bx}{n}

